\documentclass[a4paper,11pt]{article}
\usepackage[T1]{fontenc}
\usepackage{tikz}
\usetikzlibrary{arrows}
\usetikzlibrary{calc}
\usetikzlibrary{snakes}
\usepackage{cite}
\usepackage{graphicx}
\usepackage{array}
\usepackage{mdwmath}
\usepackage{mdwtab}
\usepackage{eqparbox}
\usepackage{fixltx2e}
\usepackage{url}
\usepackage{pgf}
\usepackage{lipsum}
\usepackage{multirow}
\usepackage{subcaption}
\usepackage[cmex10]{amsmath}
\usepackage{amsfonts,amssymb,amsthm}
\usepackage{xspace}
\usepackage{xcolor}
\usepackage{algorithmic}
\usepackage{bm}
\usepackage{bbm}
\usepackage{booktabs}
\usepackage{etoolbox}
\usepackage{makecell}
\usepackage{nicefrac}
\usepackage{textcomp}
\usepackage{stmaryrd}
\usepackage{mathrsfs}
\usepackage{paralist}
\usepackage{comment}

\usepackage[font=footnotesize,labelsep=space,labelfont=bf]{caption}

\usepackage{url}

\usepackage[pdftex,bookmarks=true,pdfstartview=FitH,colorlinks,linkcolor=blue,filecolor=blue,citecolor=blue,urlcolor=blue,pagebackref=false]{hyperref}
\makeatletter

\newcommand{\Rmnum}[1]{\expandafter\@slowromancap\romannumeral #1@}
\makeatother

\newtheorem{theorem}{Theorem}
\newtheorem*{theorem*}{Theorem}
\newtheorem{definition}{Definition}
\newtheorem{lemma}{Lemma}
\newtheorem*{lemma*}{Lemma}
\newtheorem{corollary}{Corollary}
\newtheorem*{cor*}{Corollary}
\newtheorem{remark}{Remark}
\newtheorem{fact}{Fact}

\newcommand{\namedref}[2]{\hyperref[#2]{#1~\ref*{#2}}}

\definecolor{darkred}{rgb}{0.5, 0, 0} 
\definecolor{darkblue}{rgb}{0,0,0.5} 
\hypersetup{
    colorlinks=true,
    linkcolor=darkred,
    citecolor=darkblue,
    urlcolor=darkblue   
}

\newcommand{\bzero}{\ensuremath{{\bf 0}}\xspace}

\newcommand{\bg}{\ensuremath{{\bf g}}\xspace}

\newcommand{\bx}{\ensuremath{{\bf x}}\xspace}

\newcommand{\by}{\ensuremath{{\bf y}}\xspace}

\newcommand{\bu}{\ensuremath{{\bf u}}\xspace}
\newcommand{\bv}{\ensuremath{{\bf v}}\xspace}

\newcommand{\bX}{\ensuremath{{\bf X}}\xspace}

\newcommand{\C}{\ensuremath{\mathcal{C}}\xspace}

\newcommand{\I}{\ensuremath{\mathcal{I}}\xspace}

\newcommand{\E}{\ensuremath{\mathcal{E}}\xspace}
\newcommand{\G}{\ensuremath{\mathcal{G}}\xspace}
\newcommand{\R}{\ensuremath{\mathbb{R}}\xspace}

\renewcommand{\paragraph}[1]{\smallskip\noindent{\bf #1}~}

\usepackage{algorithm}
\usepackage{algorithmic}

\usepackage{fullpage}
\usepackage{authblk}
\usepackage{amsmath}
\usepackage{amsfonts}

\usepackage{etoolbox}\AtBeginEnvironment{algorithmic}{\small}
\newtheorem{assumption}{Assumption}

\newcommand{\verts}[1]{\left\Vert #1 \right\Vert}
\newcommand{\lragnle}[1]{\left\langle #1 \right\rangle}

\begin{document}

\title{SPARQ-SGD: Event-Triggered and Compressed Communication in Decentralized Stochastic Optimization}

\author[1]{Navjot Singh}
\author[1]{Deepesh Data}
\author[2]{Jemin George}
\author[1]{Suhas Diggavi}

\affil[1]{University of California, Los Angeles, USA}
\affil[1] {\text{navjotsingh@ucla.com, deepesh.data@gmail.com, suhasdiggavi@ucla.edu}\vspace{0.25cm}}
\affil[2]{US Army Research Lab, Maryland, USA}
\affil[2] {\text{jemin.george.civ@mail.mil}}

\date{\vspace{-5ex}}
\maketitle

\begin{abstract}

In this paper, we propose and analyze SPARQ-SGD, which is an
event-triggered and compressed algorithm for decentralized training of
large-scale machine learning models over a graph. Each node can locally compute a
condition (event) which triggers a communication where quantized
and sparsified local model parameters are sent. In SPARQ-SGD each node
takes at least a fixed number ($H$) of local gradient steps and then
checks if the model parameters have significantly changed compared to
its last update; it communicates further compressed model parameters
only when there is a significant change, as specified by a (design)
criterion. We prove that the SPARQ-SGD converges as $O(\frac{1}{nT})$
and $O(\frac{1}{\sqrt{nT}})$ in the strongly-convex and non-convex
settings, respectively, demonstrating that such aggressive compression,
including event-triggered communication, model sparsification and
quantization does not affect the overall convergence rate as compared
to uncompressed decentralized training; 
thereby theoretically yielding communication efficiency for ``free''. We
evaluate SPARQ-SGD over real datasets to demonstrate significant
amount of savings in communication over the state-of-the-art. 
\end{abstract}

\section{Introduction}\label{intro}
There has been a recent interest in communication
efficient \emph{decentralized} training of large-scale machine
learning
models \emph{e.g.,} \cite{lian2017can,Tong_decentralized18,koloskova_decentralized_2019-1}.
In decentralized training, the nodes do not have a central
coordinator, and are not directly connected to all other nodes, but
are connected through a communication graph. This implies that the
communication is inherently more efficient, as the local connection
(degree) of such graphs could be a small constant, independent of the
network size.  In this paper, we propose SPARQ-SGD\footnote{Acronym
stands for SParsified Action Regulated Quantized SGD.} to improve
communication efficiency of decentralized training
through \emph{event-driven} exchange of quantized and sparsified model
parameters between the nodes.

Over the past few years, a number of different methods have been
developed to achieve communication efficiency in {\em distributed}
SGD, where there exists a central coordinator. These can be broadly
divided into 2 categories.  In the first category, to reduce
communication, workers send {\em compressed} updates either with
sparsification \cite{speech2,AjiHeafield17,DeepCompICLR18,stich_sparsified_2018,alistarh_convergence_2018}
or
quantization \cite{alistarh-qsgd-17,terngrad,teertha,stich-signsgd-19}
or a combination of
both \cite{basu_qsparse-local-sgd:_2019}.\footnote{\label{sparse-quan-footnote}In
sparsification, the vector sparsification is done by selecting either
its top $k$ entries (in terms of the absolute value) or random $k$
entries, where $k$ is less than the dimension of the
vector. Quantization consists of discretization of the vector by
rounding off its entries either randomly or deterministically (in the
extreme case, this can be just the sign operator).}  Another class of
algorithms that are based on the idea of {\em infrequent
communication}, workers do not communicate in each iteration; rather,
they send the updates after performing a {\em fixed} number of local
gradient
steps \cite{basu_qsparse-local-sgd:_2019,stich_local_2018,alibaba_local,Coppola15}.
The idea of compressed communication, using quantization or
sparsification, has been extended to the setting of {\em
decentralized}
optimization \cite{Tong_decentralized18,koloskova_decentralized_2019,koloskova_decentralized_2019-1}.


In this paper, we propose SPARQ-SGD with {\em event-triggered} communication, 
where a node initiates a (communication) action regulated by a locally computable 
triggering condition (event), thereby further reducing the communication among nodes. 
In particular, the proposed triggering condition is such that at least a fixed number of local 
gradient steps or iterations (say, $H$ local iterations) are first completed and after that the condition 
checks if there is a significant change (beyond a certain threshold) in its local model parameter vector 
since the last time communication occurred. Only if the change in model parameter exceeds the 
prescribed threshold, does a node trigger compressed communication. As far as we know, 
such an idea of event-triggered and compressed communication has not been proposed and 
analyzed in the context of decentralized (stochastic) training of large-scale machine learning models.

As mentioned earlier, in addition to event-triggered communication, we
also incorporate compression of the model parameters, when a
node communicates; \emph{i.e.,} when a node communicates its model
parameters, it sends a quantized and sparsified version of the model
parameters.  We therefore combine the recent ideas applied to
communication efficient training (quantization and sparsification)
with our event-triggered communication to propose
SPARQ-SGD\footnote{The idea of combining compression and \emph{fixed}
number of local iterations has been carried out in a {\em distributed}
setting (the master-worker architecture)
in \cite{basu_qsparse-local-sgd:_2019}.  In this work, in addition to
\emph{extending} this combination to the {\em decentralized} setting, we also
propose and analyze event-triggered communication.}; see
Algorithm~\ref{alg_dec_sgd_li}.  We analyze the performance of our
algorithm for both convex and (smooth) non-convex objective
functions, in terms of its convergence rate as a function of the
number of iterations $T$ (and also the number of communication rounds) and the amount of communication bits
exchanged to learn a model to a certain accuracy. We prove that the
SPARQ-SGD converges as $O(\frac{1}{nT})$ and $O(\frac{1}{\sqrt{nT}})$ 
in strongly-convex and non-convex settings, respectively, demonstrating that
such aggressive compression, including event-triggered communication
does not affect the overall convergence rate as compared to a
uncompressed decentralized training \cite{lian2017can}. Moreover, we
show that SPARQ-SGD yields significant amount of saving in
communication over the state-of-the-art; see Section~\ref{experiments} for more details.

\paragraph{Related work.}
In decentralized
setting, \cite{Tong_decentralized18,reisizadeh2018quantized}, propose
unbiased stochastic compression for gradient
exchange. \cite{assran2018stochastic,tatarenko2017non} analyze
Stochastic Gradient Push algorithm for non-convex objectives which
approximates distributed averaging instead of compressing the
gradients. Our work most closely relates
to \cite{koloskova_decentralized_2019-1} which proposed CHOCO-SGD, which uses compressed (sparsified {\em or}
quantized) updates; the distinction is that we propose
an \emph{event-triggered} communication where sparsified {\em and}
quantized model parameters are transmitted only when certain
conditions are met, further reducing communication.  The idea of
event-triggered communication has been explored previously in the
control
community \cite{Paulo-event-triggered12,distri-event-triggered12,event-triggered-consensus13,dynamic-event-triggered15}, \cite{liu2017asynchronous}
and in optimization
literature \cite{convex-optimize-event15,event-gradient-sum-consensus16,optimize-dynamic-event-triggered18}. These
papers focus on continuous-time, deterministic optimization algorithms
for convex problems; in contrast, we propose event-driven stochastic
gradient descent algorithms for both convex and non-convex problems.
\cite{chen2018lag} propose an adaptive scheme to skip gradient computations in a \emph{distributed} setting for \emph{deterministic} gradients; moreover, their focus is on saving communication rounds, and  do not have any compressed communication. As far as we know, our idea of event-triggered and compressed communication has not been studied for decentralized stochastic optimization.

\paragraph{Contributions.}
We study optimization in a decentralized setup, where $n$ different
workers, each having a different dataset $\mathcal{D}_i$ (the dataset
$\mathcal{D}_i$ has an associated objective function $f_i:\R^d\to\R$),
are linked through a connected graph $\G=([n],\E)$, where
$[n]:=\{1,2,\hdots,n\}$. Vertex $i$ in $\G$ is associated with the
$i$th worker who can only communicate with its neighbours
$\mathcal{N}_i=\{j\in[n]:\{i,j\}\in \E\}$.  We consider the empirical
risk minimization of the loss function:
\begin{align}\label{eq:obj-fn}
f(\bx) = \frac{1}{n} \sum_{i=1}^{n} f_i(\bx),
\end{align}
where $f_i(\bx)
= \mathbb{E}_{\xi_i\sim\mathcal{D}_i}[F_i(\bx,\xi_i)]$, where
$\xi_i\sim\mathcal{D}_i$ denotes a random data sample from
$\mathcal{D}_i$ and $F_i(\bx,\xi_i)$ denotes the risk associated with
the data sample $\xi_i$ w.r.t.~$\bx$ at the $i$th worker node.  We
solve the decentralized optimization in \eqref{eq:obj-fn} using
SPARQ-SGD.  Our theoretical results are the convergence analyses for
both strongly convex and non-convex objectives in the synchronous
setting; see Theorem~\ref{thm_cvx_li} and ~\ref{thm_noncvx_fix_li},
respectively.  In the strongly-convex setting, we show a convergence
rate of  $\mathcal{O}\left( \frac{1}{ nT}\right)+ 
\mathcal{O} \left( \frac{{c_0}}{ \delta^2  T^{(1+\epsilon)}}   \right) 
 + \mathcal{O} \left( \frac{{H^2}}{  \delta^4 \omega^2 T^2}   \right) + 
 \mathcal{O} \left( \frac{H^3}{ \omega^3 \delta^6T^3}  \right)$ for some $\epsilon \in (0,1)$,
the factors ($c_0$ for triggering threshold, $H$ for number of local
iterations, and $\omega$ for compression) for communication
efficiency, and $\delta$, the spectral-gap of the connectivity matrix
$W$, appear in the higher order terms.  Thus, for large enough $T$,
they do not affect the dominating term
$\mathcal{O}\left( \frac{1}{nT}\right)$, which, in fact, is the
convergence rate of centralized vanilla SGD with mini-batch size of
$n$.  Similar observation is also made in the non-convex setting,
where we get a convergence rate of $\mathcal{O}(\frac{1}{\sqrt{nT}})$;
see Corollary~\ref{cvx_cor}~and~\ref{noncvx_cor} and the following
remarks for more details.  Hence, for both the objectives, we get
essentially the same convergence rate as that of vanilla SGD, even
after applying SPARQ-SGD to gain communication efficiency; and hence,
we get communication efficiency essentially ``for free''.
 We compare our algorithm against CHOCO-SGD \cite{koloskova_decentralized_2019}, which is the state-of-the-art in compressed decentralized training and provide theoretical justification for communication efficiency of SPARQ-SGD over CHOCO-SGD to achieve the same target accuracy. We corroborate our theoretical understanding with numerical results in Section~\ref{experiments} where we demonstrate that SPARQ-SGD yields significant savings in communication bits. For a convex objective simulated on the MNIST dataset, SPARQ-SGD saves total communicated bits by a factor of $250 \times $ compared to CHOCO-SGD \cite{koloskova_decentralized_2019-1} and by $1000\times$ compared to vanilla SGD to converge to the same target accuracy. Similarly, for a non-convex objective simulated on the CIFAR-10 dataset \cite{cifar}, we save total bits by a factor of $250 \times $ compared to CHOCO-SGD \cite{koloskova_decentralized_2019} and around $15 \text{K} \times$ compared to vanilla SGD to reach the same target accuracy.

\paragraph{Paper organization.}
We describe SPARQ-SGD, our proposed algorithm, in Section~\ref{sparq-sgd}. 
In Section~\ref{main-results}, we state our main results for strongly-convex and non-convex objectives,
and give proof outlines of these theorems in Section~\ref{proof-outlines}.
We validate our theoretical findings with numerical experiments in Section~\ref{experiments}.


\section{Our Algorithm: SPARQ-SGD}\label{sparq-sgd}
In this section, we describe SPARQ-SGD, our decentralized SGD algorithm with compression and event-triggered communication.
First we need to define its main ingredients.

\begin{definition}[Compression, \cite{stich_sparsified_2018}]\label{definition:compression}
A (possibly randomized) function $\C:\R^d\to\R^d$ is called a {\em compression} operator, if there exists a positive constant $\omega<1$, such that the following holds for every $\bx\in\R^d$:
\begin{align}\label{eq:compression}
\mathbb{E}_{\C}[\|\bx-\C(\bx)\|_2^2]\leq (1-\omega)\|\bx\|_2^2,
\end{align}
where expectation is taken over the randomness of $\C$.
We assume $\C(\bzero)=\bzero$.
\end{definition}

It is known that some important sparsifiers as well as quantizers are examples of compression operators:
{\sf (i)} $Top_k$ and $Rand_k$ sparsifiers (in which we select $k$ entries; see Footnote~\ref{sparse-quan-footnote}) with $\omega=k/d$ \cite{stich_sparsified_2018},
{\sf (ii)} Stochastic quantizer $Q_s$ from \cite{alistarh-qsgd-17}\footnote{$Q_s:\R^d\to\R^d$ is a stochastic quantizer, if for every $\bx\in\R^d$, we have {\sf(i)} $\mathbb{E}[Q_s(\bx)]=\bx$ and {\sf (ii)} $\mathbb{E}[\|\bx-Q_s(\bx)\|_2^2]\leq\beta_{d,s}\|\bx\|_2^2$. $Q_s$ from \cite{alistarh-qsgd-17} satisfies this definition with $\beta_{d,s}=\min\left\{\frac{d}{s^2},\frac{\sqrt{d}}{s}\right\}$.} with $\omega=(1-\beta_{d,s})$ for $\beta_{d,s}<1$, and 
{\sf (iii)} Deterministic quantizer $\frac{\|\bx\|_1}{d}Sign(\bx)$ from \cite{stich-signsgd-19} with $\omega=\frac{\|\bx\|_1^2}{d\|\bx\|_2^2}$. 
It was shown in \cite{basu_qsparse-local-sgd:_2019} that if we compose these sparsifiers and quantizers, the resulting operator 
also gives compression and outperforms their individual components. For example, for any $Comp_k\in\{Top_k,Rand_k\}$, the following are compression operators: {\sf (iv)} $\frac{1}{(1+\beta_{k,s})}Q_s(Comp_k)$ with $\omega=\left(1-\frac{k}{d(1+\beta_{k,s})}\right)$ for any $\beta_{k,s}\geq0$, and {\sf (v)} $\frac{\|Comp_k(\bx)\|_1SignComp_k(\bx)}{k}$ with $\omega=\max\left\{\frac{1}{d},\frac{k}{d}\left(\frac{\|Comp_k(\bx)\|_1^2}{d\|Comp_k(\bx)\|_2^2}\right)\right\}$.


\paragraph{Event-triggered communication.}
As mentioned in Section~\ref{intro}, our proposed event-triggered
communication consists of two phases: in the first phase, nodes perform
a fixed number $H$ of local iterations, and in the second phase, they 
check for the communication-triggering condition (event), if satisfied, then they send the (compressed) updates.
%
%
%
Let $\I_T\subseteq[T]$ denote a set
of indices at which workers check for the triggering condition. Since
we are in the synchronous setting, we assume that $\I_T$ is same for
all workers.  Let $\I_T=\{I_{(1)},I_{(2)},\hdots,I_{(k)}\}$. The gap
of $\I_T$ is defined as
$gap(\I_T):=\max_{i\in[k-1]}\{(I_{(i+1)}-I_{(i)})\}$, \cite{stich_local_2018},
which is equal
to the maximum number of local iterations a worker performs before
checking for the triggering condition. Note
that $gap(\I_T)=1$ is equivalent to the case when workers check for
the communication triggering criterion in every iteration. 

\begin{algorithm}[hbt]
	\caption{SPARQ-SGD: SParsified Action Regulated Quantized SGD}
	\label{alg_dec_sgd_li}
	\begin{algorithmic}[1]
		\STATE Initial values $\bx_i^{(0)} \in \mathbb{R}^d$ on each node $i \in [n]$, consensus stepsize $\gamma$, SGD stepsizes $\{ \eta_t \}_{t \geq 0} $, threshold sequence $ \{c_t\}_{t \geq 0}$, compression operator \C having parameter $\omega$, communication graph $G = ([n],E)$ and mixing matrix $W$, set of synchronization indices $\mathcal{I}_T$, initialize $\hat{\bx}_i^{(0)} := 0$ for all $i$
		\FOR{$t=0$ {\bfseries to} $T-1$ in parallel for all workers $i \in [n]$ } 
		\STATE Sample $\xi_i^{(t)}$ and compute stochastic gradient $\bg_i^{(t)}:= \nabla F_i(\bx_i^{(t)}, \xi_i^{(t)})$
		\STATE $\bx_i^{(t+\frac{1}{2})} := \bx_i^{(t)} - \eta_t \bg_i^{(t)}$
		\IF{$(t+1) \in I_T $ }
		\FOR{neighbors $j \in \mathcal{N}_i \cup i $  }
		\IF{ $\Vert \bx_i^{(t+\frac{1}{2})} - \hat{\bx}_i^{(t)} \Vert_2^2 > {c_t \eta_t^2} $}
		\STATE Compute $\mathbf{q}_i^{(t)} := \C(\bx_i^{(t+\frac{1}{2})} - \hat{\bx}_i^{(t)} )$
		\STATE Send $\mathbf{q}_i^{(t)}$ and receive $\mathbf{q}_j^{(t)}$
		\ELSE 
		\STATE Send $\mathbf{0}$ and receive $\mathbf{q}_j^{(t)}$
		\ENDIF
		\STATE $\hat{\bx}_j^{(t+1)} := \mathbf{q}_j^{(t)} + \hat{\bx}_j^{(t)} $
		\ENDFOR
		\STATE $\bx_i^{(t+1)} = \bx_i^{(t+\frac{1}{2})} + \gamma \sum \limits_{j \in \mathcal{N}_i } w_{ij} (\hat{\bx}_j^{(t+1)} - \hat{\bx}_i^{(t+1)} ) $
		\ELSE
		\STATE $\hat{\bx}_i^{(t+1)} = \hat{\bx}_i^{(t)}$ ,  $\bx_i^{(t+1)} = \bx_i^{(t+\frac{1}{2})}$ for all $i \in [n]$
		\ENDIF
		\ENDFOR
	\end{algorithmic}
\end{algorithm}

Our algorithm, SPARQ-SGD, for optimizing \eqref{eq:obj-fn} in a decentralized setting is presented in Algorithm~\ref{alg_dec_sgd_li}.
For designing this, in addition to combining sparsification {\em and} quantization, we carefully incorporate local iterations and event-triggered communication into the CHOCO-SGD algorithm from \cite{koloskova_decentralized_2019-1}, which uses only sparsified {\em or} quantized updates. 
This poses several technical challenges in proving the convergence; see the proofs of Theorem~\ref{thm_cvx_li}, ~\ref{thm_noncvx_fix_li}, and in particular, the proof of Lemma~\ref{lem_dec_li_sgd}.

In SPARQ-SGD, each node $i \in [n]$ maintains a local parameter vector $\bx_i^{(t)}$, and their goal is to achieve consensus among themselves on the 
value of $\bx$ that minimizes \eqref{eq:obj-fn}, while allowing only for compressed and infrequent communication.
Node $i$ updates $\bx_i^{(t)}$ in each iteration $t$ by a stochastic gradient step (line 4). 
An estimate $\hat{\bx}_i^{(t)}$ of $\bx_i^{(t)}$ is also maintained at each neighbor $j \in \mathcal{N}_i$ and at $i$ itself. 
Thus, each node maintains an estimate of all its neighbors' local parameter vectors and of itself. 
In our algorithm, $\mathcal{I}_T$ is the set of indices for which the workers check for the triggering condition and take a consensus step. We also allow the triggering threshold ($c_t$) to vary with $t$ with the requirement that $c_t \sim o(t)$.
At time-step $t$, if $(t+1) \in\mathcal{I}_T$, the nodes check for the triggering condition (line 7),
if satisfied, then each node $i \in [n]$ sends to all its neighbors the compressed difference between its local parameter vector and its estimate that its neighbors have (line 8); and based on the messages received from its neighbors, the $i$th node updates $\hat{\bx}_j^{(t)}$ -- the estimate of the $j$th node's local parameter vector (line 13), and then every node performs the consensus step (line 15).



In SPARQ-SGD, observe that every worker node initializes its estimate $\hat{\bx}_i^{(0)}$ of the $i$th node's local parameter vector $\bx_i^{(0)}$ to be $\hat{\bx}_i^{(0)} := 0$, whereas, in principle, it should have been equal to $\bx_i^{(0)}$. 
To ensure this, in the first round of our algorithm, every worker sends its (compressed) local parameter vector to all its neighbours.

\section{Main Results}\label{main-results}

Our main results are under the following assumptions: 

\paragraph{Assumptions.}
{\sf (i)} {\bf $L$-Smoothness:} Each local function $f_i$ for $i \in [n]$ is $L$-smooth, i.e, $\forall \bx,\by \in \mathbb{R}^d$, we have 
$f_i(\by) \leq f_i(\bx) + \langle \nabla f_i(\bx), \by-\bx  \rangle + \frac{L}{2} \Vert \by-\bx \Vert^2$.
{\sf (ii)} {\bf Bounded variance:} For every $i\in[n]$, we have
$\mathbb{E}_{\xi_i} \Vert \nabla F_i (\bx,\xi_i) - \nabla f_i(\bx) \Vert^2 \leq \sigma^2$, for some finite $\sigma$,
where $\nabla F_i (\bx,\xi_i)$ is the unbiased gradient at worker $i$ such that $\mathbb{E}_{\xi_i} [\nabla F_i (\bx ,\xi_{i})] = \nabla f_i (\bx)$.
We define the average variance across all workers as $\bar{\sigma}^2 := \frac{1}{n} \sum_{i=1}^{n}  \sigma_i ^2$.
{\sf (iii)} {\bf Bounded second moment:} For every $i\in[n]$, we have $\mathbb{E}_{\xi_i} \Vert \nabla F_i (\bx,\xi{i}) \Vert^2 \leq G^2$, for some finite $G$.

Before stating the main results, we need some notations about the underlying communication graph $\G$ first.
Let $W \in \mathbb{R}^{n \times n}$ denote the weighted connectivity matrix of
$\G$, with $w_{ij}$ for every $i,j \in [n]$ being its $(i,j)$th entry,
which denotes the weight on the link between worker $i$ and $j$.  $W$
is assumed to be symmetric and doubly stochastic, which implies that
all its eigenvalues $\lambda_i(W), i=1,2,\hdots,n$, lie in $[-1,1]$.
Without loss of generality, assume that $|\lambda_1(W) |> |
\lambda_2(W) | \geq \hdots \geq | \lambda_n(W) |$.  Since $W$ is
doubly stochastic, we have $\lambda_1(W)=1$, and since $\G$ is
connected, we have $\lambda_2(W) < \lambda_1(W)$.  Let the spectral
gap of $W$ be defined as $\delta := 1- | \lambda_2(W) |$. Since
$|\lambda_2(W)|\in[0,1)$ we have that $\delta\in(0,1]$.  It is known
that simple matrices $W$ with $\delta >0$ exist for every connected
graph, \cite{koloskova_decentralized_2019-1}.

Now we state the main results of this paper both for strongly-convex and non-convex objectives.
As mentioned in Section~\ref{intro}, even after applying the techniques of compression and infrequent communication, we prove a convergence rate, matching with that of vanilla SGD in both strongly-convex and non-convex settings.

\begin{theorem}[Smooth and strongly-convex objective with decaying learning rate]\label{thm_cvx_li} 
Suppose $f_i$, for all $i\in[n]$ be $L$-smooth and $\mu$-strongly convex. Let $\C$ be a compression operator with parameter equal to $\omega \in (0,1]$. Let $gap(\I_T)\leq H$. 
If we run SPARQ-SGD with consensus step-size $\gamma = \frac{2\delta \omega}{64 \delta + \delta^2 + 16 \beta^2 + 8 \delta \beta^2 - 16\delta \omega}, (\text{where }\beta = \text{max}_i \{ 1- \lambda_i(W) \})$, an increasing threshold function 
$c_t  \leq c_0t^{(1-\epsilon)} $ for all $t$ where constant $c_0 \geq 0$ and $\epsilon \in (0,1)$ and decaying learning rate $\eta_t = \frac{8}{\mu(a+t)}$, where
$a \geq \max \{\frac{5H}{p}, \frac{32L}{\mu}\}$ for $p = \frac{\gamma\delta}{8}$, and let the algorithm generate $\{ \bx_i ^{(t)}  \}_{t=0}^{T-1}$ for $i\in[n]$, then the following holds:
	\begin{align*}
	\mathbb{E}f(\bx_{avg}^{(T)}) - f^* & \leq \frac{\mu a^3}{8S_T}\Vert \bx^{(0)} - \bx^* \Vert^2  + \frac{4T(T+2a)}{\mu S_T}\frac{ \Bar{\sigma}^2}{n} + \frac{512T}{\mu^2 S_T} \left(  2L + \mu  \right) \left( \frac{160}{p^2}\right)G^2H^2 \\
	& \qquad +\frac{6400c_0 \omega T ^{(2-\epsilon)}}{\mu^2 (2-\epsilon)S_T} \left(  \frac{2L + \mu}{p}  \right)
	\end{align*}
where $\Bar{\bx}^{(T)}_{avg} = \frac{1}{S_T} \sum_{t=0}^{T-1}w_t \bar{\bx}^{(t)}$, where $\bar{\bx}^{(t)} = \frac{1}{n} \sum_{i=1}^{n} \bx^{(t)}_{i}$, weights $w_t = (a+t)^2$, and $S_T = \sum_{t=0}^{T-1}w_t \geq \frac{1}{3}T^3$.
\end{theorem}
We provide a proof of Theorem \ref{thm_cvx_li} in Appendix \ref{proof_thm_cvx_li}. The analysis provided also works for any $c_t \sim o(t)$, however we provide it for $c_t \leq c_0 t^{(\epsilon-1)}$ to highlight the main idea.
Observe that the consensus step-size $\gamma$ does not appear explicitly in the above rate expression, 
but it does affect the convergence indirectly through $p=\gamma\delta/8$.
Note that $\delta\in(0,1]$, $\beta\leq 2$, and $\omega\geq0$. Substituting these in the expression of $\gamma$ and $p$ gives $\gamma \geq \frac{2\delta\omega}{161}$ and $p\geq\frac{\delta^2 \omega}{644}$; see also the proof of Lemma~\ref{lem_dec_li_sgd}.
Now we simplify the above expression to gain further insights as to how our techniques for reducing communication is affecting the convergence rate.

\begin{corollary}\label{cvx_cor}
Using $\mathbb{E} \verts{\bx^{(0)} - \bx^*}_2^2 \leq \frac{4G^2}{\mu^2}$ (from \cite[Lemma 2]{rakhlin2011making}) and $p\geq\frac{\delta^2 \omega}{644}$, and hiding constants (including $L$) in the $\mathcal{O}$ notation, we can simplify the rate expression in Theorem~\ref{thm_cvx_li} to the following:
\begin{align*}
	\mathbb{E}[f(\Bar{\bx}^{(T)}_{avg})] - f^* &\leq \mathcal{O}\left( \frac{\bar{\sigma}^2}{\mu nT}\right)+ \mathcal{O} \left( \frac{{c_0}}{ \mu^2 \delta^2  T^{(1+\epsilon)}}   \right)  + \mathcal{O} \left( \frac{{G^2H^2}}{ \mu^2 \delta^4 \omega^2 T^2}   \right) + \mathcal{O} \left( \frac{G^2H^3}{\mu \omega^3 \delta^6T^3}  \right)
\end{align*}
\end{corollary}
\begin{remark}\label{sgd-rate}
Observe that the dominating term $\mathcal{O}\left( \frac{\bar{\sigma}^2}{\mu nT}\right)$ is not affected by the compression factor $\omega$, the number of local iterations $H$, the factor $c_0$ in the triggering condition, and the topology of the underlying communication graph (which is controlled by the spectral gap $\delta$) -- they all appear in the higher order terms.
In order to ensure that they do not affect the dominating term while converging at a rate of $\mathcal{O}\left( \frac{\bar{\sigma}^2}{\mu nT}\right)$, we would require $T \geq T_0:=C\times\max\left\{\left(\frac{nc_0}{\mu\delta^2\bar{\sigma}^2}\right)^{\frac{1}{\epsilon}},\left(\frac{nH^2G^2}{\mu\bar{\sigma}^2\delta^4\omega^2}\right)\right\}$ for sufficiently large constant $C$. 
This implies that for large enough $T$, we get benefits of all these techniques in saving communication bits, without affecting the convergence rate significantly.

Now we analyze the effect of $\omega,H,c_0,\delta$ on the threshold $T_0$: 
{\sf(i)} if we compress the communication more, i.e., smaller $\omega$, then $T_0$ increases, as expected;
{\sf(ii)} if we take more number of local iterations $H$, $T_0$ would again increase, as expected, because increasing $H$ means communicating less frequently;
{\sf(iii)} if we increase $c_0$, which means that the triggering threshold has become bigger, we expect less frequent communication, thus $T_0$ increases, as expected;
{\sf(iv)} if the spectral gap $\delta\in(0,1]$ is closer to 1, which implies that the graph is well-connected, then the threshold $T_0$ decreases,
which is also expected, as good connectivity means faster spreading of information, resulting in faster consensus.\footnote{If we are to design the underlying communication graph, one possible choice is to consider the {\em expander graphs}, \cite{Wotao-expander-decentralized16}, that will simultaneously give low communication and faster convergence, as they have constant degree and large spectral gap, \cite{expander-survey}.}
\end{remark}
\begin{remark}\label{sgd-rate2}
Observe that after a large enough $T\geq T_0$, we get the same rate as that of distributed vanilla SGD and also a distributed gain of $n$ with the number of nodes.
Thus, we essentially converge at the same rate as that of vanilla SGD, while significantly saving in terms of communication bits among all the workers; this can be seen in our numerical results in Section~\ref{experiments}.
\end{remark}
Now we state our convergence result for the non-convex objective.
\begin{theorem}[Smooth and non-convex objective with fixed learning rate]\label{thm_noncvx_fix_li} 
Suppose $f_i$, for all $i\in[n]$ be $L$-smooth. Let $\C$ be a compression operator with parameter equal to $\omega \in (0,1]$. Let $gap(\I_T)\leq H$. 
If we run SPARQ-SGD for $T \geq 64nL^2$ iterations with fixed learning rate $\eta = \sqrt{\frac{n}{T}}$, an increasing threshold function $c_t$ such that $c_t < \frac{1}{\eta}$ for all $t$ and consensus step-size $\gamma = \frac{2\delta \omega}{64 \delta + \delta^2 + 16 \beta^2 + 8 \delta \beta^2 - 16\delta \omega}$, (where $\beta = \max_i \{ 1- \lambda_i(W) \}$), and let the algorithm generate $\{ \bx_i ^{(t)}  \}_{t=0}^{T-1}$ for $i\in[n]$, then the averaged iterates $\bar{\bx}^{(t)} := \frac{1}{n} \sum_{i=0}^n \bx_i^{(t)}$ satisfy:
	\begin{align*}
	\frac{\sum_{t=0}^{T-1}  \mathbb{E} \Vert \nabla f(\bar{\bx}^{(t)}) \Vert_2^2}{T}  &
\leq \frac{4 \left( f(\bar{\bx}_{0}) - f^* + L\bar{\sigma}^2 \right) }{\sqrt{nT}} +  \frac{64G^2H^2L^2n}{T p^2} \left( 1+ \frac{2p}{\omega}  \right)  + \frac{20L^2 \omega  \sqrt{n^{(1+\epsilon)}}  } { p \sqrt{T^{(1+\epsilon)}} } \\
& \qquad  +  \frac{256G^2H^2L^3  n^{\nicefrac{3}{2}} }{T^{\nicefrac{3}{2}} p^2} \left( 1+ \frac{2p}{\omega}  \right)  +  \frac{80L^3 \omega  \sqrt{n^{(2+\epsilon)}}  } { p \sqrt{T^{(2+\epsilon)}} }
	\end{align*}
Here $p=\frac{\gamma\delta}{8}$ and we assume $c_t \leq \frac{1}{\eta^{(1-\epsilon)}}$ for all $t$ where $\epsilon \in (0,1)$.
\end{theorem}
We prove Theorem~\ref{thm_noncvx_fix_li} in Appendix \ref{proof_thm_noncvx_fix_li}.
As mentioned after Theorem~\ref{thm_cvx_li}, though the consensus step-size $\gamma$ does not appear in the rate expression, 
it does affect the convergence through the parameter $p$. As argued after Theorem~\ref{thm_cvx_li}, we can show similarly show that $p\geq\frac{\delta^2\omega}{644}$.
Now we simplify the above expression in the following corollary.
\begin{corollary}\label{noncvx_cor}
	Let $f(\bar{\bx}^{(0)}) - f^* \leq J^2 $, where $J^2 < \infty$ is a constant. 
	Using $p\geq\frac{\delta^2\omega}{644}$, substituting the value of $A$, and hiding constants (including $L$) in the $\mathcal{O}$ notation, we can simplify the rate expression in Theorem~\ref{thm_noncvx_fix_li} to the following:
	\begin{align*}
		\frac{\sum_{t=0}^{T-1}  \mathbb{E} \Vert \nabla f(\bar{\bx}^{(t)}) \Vert_2^2}{T}  & \leq \mathcal{O} \left(  \frac{J^2 + \bar{\sigma}^2 }{\sqrt{nT}}\right) +\mathcal{O}\left( \frac{n}{T}\left(1 + \sqrt{\frac{n}{T}}\right)\left[\frac{(1+\delta^2)G^2H^2}{\omega^2\delta^4}  \right] \right) \\
&		\qquad + \mathcal{O}\left( \left(\frac{n}{T}\right)^{\frac{1+\epsilon}{2}}\left(1 + \sqrt{\frac{n}{T}}\right)\left[\frac{1}{\delta^2}  \right] \right)
	\end{align*}
\end{corollary}
\begin{remark}\label{sgd-rate-non-cvx}
Observe that $\omega,H,\delta$ do not affect the dominating term $\mathcal{O} \left(\frac{J^2 + \bar{\sigma}^2 }{\sqrt{nT}}\right)$. 
Since Theorem~\ref{thm_noncvx_fix_li} provides non-asymptotic guarantee, we need to decide the horizon $T$ before running the algorithm;
so, to ensure that the dominating term does not get affected by these different factors, while converging at a rate of $\mathcal{O} \left(\frac{J^2 + \bar{\sigma}^2 }{\sqrt{nT}}\right)$, we would be required to fix $T \geq T_1:=C_1\times\max\left\{\left(\frac{n^{(2+\epsilon)}}{(J^2+\bar{\sigma}^2)^{2}  \delta^4 }\right)^{\nicefrac{1}{\epsilon}}, \frac{n^3G^4H^4}{(J^2+\bar{\sigma}^2)^2\omega^4\delta^4}\right\}$ for sufficiently large constant $C_1$. 
This implies that for large enough $T$, we get the benefits of all these techniques in saving on the communication bits, essentially for ``free'', without affecting the convergence rate by too much.
The rest of Remark~\ref{sgd-rate} and Remark~\ref{sgd-rate2} are also applicable here.
\end{remark}

Note that the result of Theorem~\ref{thm_noncvx_fix_li} is for fixed learning rate and gives non-asymptotic convergence;
the corresponding result with decaying learning rate, which gives an asymptotic convergence rate 
of $\mathcal{O}\left(\frac{1}{\log T}\right)$ is provided in Appendix~\ref{proof_thm_noncvx_var_li}.

\begin{remark} (Theoretical justification for communication gain)
	The convergence result for SPARQ-SGD highlights savings in communication compared to CHOCO-SGD \cite{koloskova_decentralized_2019-1}. For the sake of argument, consider the case when SPARQ-SGD only performs local iterations and no threshold based triggering ($c_t =0, \,  \forall t$) . For the same compression operator $\omega$ used for both SPARQ and CHOCO, to transmit the same number of bits (i.e., having same number of communication rounds), $T$ iterations of CHOCO would correspond to $T \times H$ iterations of SPARQ (due to H local SGD steps). Thus for the same number of bits transmitted, the bound on sub-optimality for convex objective for CHOCO is $\sim \mathcal{O}( \nicefrac{1}{ \mu nT}) + \mathcal{O}( \nicefrac{G^2}{\omega^2 \delta^4 \mu^2 T^2}  )$ while for SPARQ it is :
	$\sim \mathcal{O}(\nicefrac{1}{\mu nH T}) + \mathcal{O}( \nicefrac{G^2}{\omega^2 \delta^4 \mu^2 T^2})$. Thus for the same amount of communication (same number of communication rounds), SPARQ-SGD has a better performance compared to CHOCO-SGD (the first dominant term is affected by H).
	Similarly, for the same number of communication rounds, the bound on sub-optimality for CHOCO-SGD for non-convex objectives is $ \sim \mathcal{O}(\nicefrac{1}{\sqrt{T}}  ) + \mathcal{O}(\nicefrac{1}{T}) $ while for SPARQ-SGD it is $ \sim \mathcal{O}(\nicefrac{1}{\sqrt{HT}}) + \mathcal{O}(\nicefrac{H}{T}) $. Thus, it can be seen that for large values of T, the performance of SPARQ-SGD is better than that of CHOCO-SGD for the number of communicated bits. 
	Thus there is theoretical justification for our algorithm to have a better performance while using less bits for communication and this claim is also supported through our experiments.
\end{remark}

\section{Proof Outlines}\label{proof-outlines}
In this section, we give proof outlines of Theorem \ref{thm_cvx_li} and \ref{thm_noncvx_fix_li}.
Our proof outlines have been adapted from 
\cite{koloskova_decentralized_2019,koloskova_decentralized_2019-1}, with significant changes in the proof details 
arising due to event-triggered communication.
We provide complete proofs of both these theorems in Appendix \ref{proof_thm_cvx_li} and \ref{proof_thm_noncvx_fix_li}, respectively.

\subsection{Proof Outline of Theorem \ref{thm_cvx_li}}
Consider the collection of iterates $\{\bx_i ^{(t)}\}_{t=0}^{T-1}$, $i \in [n]$ generated by Algorithm~\ref{alg_dec_sgd_li} at time $t$. 
For any time $t\geq0$, we have from line 15 of Algorithm~\ref{alg_dec_sgd_li} that 
\begin{align*}
\bx_i^{(t+1)} = \bx_i^{(t+\frac{1}{2})} + \mathbbm{1}_{(t+1) \in \mathcal{I}_T}\gamma \sum_{j=1}^n w_{ij} (\hat{\bx}_j^{(t+1)} - \hat{\bx}_i^{(t+1)}),
\end{align*}
where $\bx_i^{(t+\frac{1}{2})} = \bx_i ^{(t)} - \eta_t \nabla F_i(\bx_i^{(t)}, \xi_i^{(t)})$ (line 4).
Note that we changed the summation from $j\in\mathcal{N}_i$ to $j=1$ to $n$; this is because $w_{ij}=0$ whenever $j\notin\mathcal{N}_i$.

Let $\bar{\bx}^{(t)} = \frac{1}{n} \sum_{i=1}^{n} \bx_{i}^{(t)}$ denote the average of the local iterates at time $t$.
Now we argue that $\bar{\bx}^{(t+1)}=\bar{\bx}^{(t+\frac{1}{2})}$.
This trivially holds when $(t+1)\notin \I_T$. For the other case, i.e., $(t+1)\in \I_T$, this follows because
$\sum_{i=1}^n\sum_{j=1}^n w_{ij} (\hat{\bx}_j^{(t+1)} - \hat{\bx}_i^{(t+1)})=0$, which uses the fact that $W$ is a doubly stochastic matrix. 
Thus, we have 
\begin{equation}\label{eq:thm_cvx_interim0}
\bar{\bx}^{(t+1)} =  \bar{\bx}^{(t)} - \frac{\eta_t}{n}\sum_{j=1}^n \nabla F_j(\bx_j^{(t)},\xi_j^{(t)}).
\end{equation}
Subtracting $\bx^*$ (the minimizer of \eqref{eq:obj-fn}) from both sides gives
\begin{align}\label{eq:thm_cvx_interim1}
	  \bar{\bx}^{(t+1)} - \bx^* =  \Bar{\bx}^{(t)} - \frac{\eta_t}{n}\sum_{j=1}^n \nabla F_j(\bx_j^{(t)},\xi_j^{(t)}) -\bx^*
\end{align}
Using $\eta_t \leq \frac{1}{4L}$ (which follows from substituting $a \geq \frac{32L}{\mu}$ in $\eta_t=\frac{8}{\mu(a+t)}$), 
together with some algebraic manipulations provided in Appendix~\ref{proof_thm_cvx_li}, we have the following sequence relation for $\{\bar{x}^{(t)}\}$:
\begin{align}\label{eq:thm_cvx_interim3}
&\mathbb{E}\Vert \Bar{\bx}^{(t+1)} - \bx^* \Vert^2 \leq \left( 1-\frac{\eta_t \mu}{2} \right) \mathbb{E} \Vert \Bar{\bx}^{(t)} - \bx^* \Vert^2 + \frac{\eta_t^2 \Bar{\sigma}^2}{n}  - \eta_te_t + \eta_t \left(  \frac{2L + \mu}{n}  \right) \sum_{j=1}^n \mathbb{E}\Vert \Bar{\bx}^{(t)} - \bx_j^{(t)} \Vert^2 
\end{align}
where $e_t := \mathbb{E}f(\Bar{\bx}^{(t)}) - f^*$ and expectation is taken w.r.t.~the entire process. 
We need to bound the last term of \eqref{eq:thm_cvx_interim3}. For this, 
let $I_{(t_0)}$ denote the last synchronization index in $\mathcal{I}_T$ before time $t$.
This, together with the assumption that $gap(\I_T)\leq H$, implies $t-I_{(t_0)}\leq H$.  
Using this and the bounded gradient assumption, we can easily bound the last term in the RHS of \eqref{eq:thm_cvx_interim3} (calculations are done in the appendix in a more general matrix form):
\begin{align}\label{eq:thm_cvx_interim4}
\sum_{j=1}^n \mathbb{E} \left\Vert \bar{\bx}^{{(t)}} - \bx^{{(t)}}_{j} \right\Vert^2 
\leq 2\mathbb{E}\sum_{j=1}^n \left\Vert \bar{\bx}^{I_{(t_0)}} - \bx^{I_{(t_0)}}_{j} \right\Vert^2 + 2n \eta_{I_{(t_0)}}^2H^2G^2 
\end{align}
In the following lemma, we show that the local iterates $\bx_j^{(t)}, j\in[n]$ asymptotically approach to the average iterate $\bar{\bx}^{(t)}$, thereby proving the contraction of the first term on the RHS of \eqref{eq:thm_cvx_interim4}.

\begin{lemma}[Contracting deviation of local iterates and the averaged iterates]\label{lem_dec_li_sgd}
Under the assumptions of Theorem~\ref{thm_cvx_li}, for any $I_{(t)}$ such that $I_{(t)}\in\I_T$, we have
\begin{align*}
\sum_{j=1}^n \mathbb{E} \left\Vert \bar{\bx}^{I_{(t)}} - \bx^{I_{(t)}}_{j} \right\Vert^2 \leq \frac{20A_{I_{(t)}}\eta_{I_{(t)}}^2}{p^2},
\end{align*}
where $A_{I_{(t)}}=2nG^2H^2 + \frac{p }{2} \left(\frac{8nG^2H^2}{\omega} + \frac{5\omega n c_{I_{(t)}}}{4} \right)$ with $c_{I_{(t)}}$ denoting the threshold function evaluated at timestep $I_{(t)}$.
\end{lemma}
We give a proof sketch of the above lemma at the end of this proof; see Appendix~\ref{proof_lem_dec_li_sgd} for a complete proof.

Note that $\eta_{I_{(t_0)}}\leq 2\eta_t$, which follows from the following set of inequalities:
$\frac{\eta_{I_{(t_0)}}}{\eta_{t}} = \frac{a+t}{a+ I_{(t_0)}} \leq \frac{a+I_{(t_0)}+H}{a+ I_{(t_0)}} 
\stackrel{\text{(a)}}{\leq} \frac{2(a+I_{(t_0)})}{a+ I_{(t_0)}} = 2$, where (a) follows from our assumption that $a\geq H$.
Now, substituting the bound from Lemma~\ref{lem_dec_li_sgd} in \eqref{eq:thm_cvx_interim4} and using 
$\eta_{I_{(t_0)}}\leq 2\eta_t$ gives
$\sum_{j=1}^n \mathbb{E} \left\Vert \bar{\bx}^{{(t)}} - \bx^{{(t)}}_{j} \right\Vert^2 \leq  4\eta_t^2\left(\frac{40A_t}{p^2} + 2nH^2G^2\right)$.
Putting this back in \eqref{eq:thm_cvx_interim3} yields
\begin{align*}
	\mathbb{E} \Vert \Bar{\bx}^{(t+1)} - \bx^* \Vert^2  \leq \left( 1-\frac{\eta_t \mu}{2} \right) \mathbb{E} \Vert \Bar{\bx}^{(t)} - \bx^* \Vert^2  + \frac{\eta_t^2 \Bar{\sigma}^2}{n} - \eta_t e_t + 4 \eta_t^3 \left(  \frac{2L + \mu}{n}  \right) \left( \frac{40A_t}{p^2} + 2nH^2 G^2 \right)
\end{align*}
Substituting the value of $A_t = 2nG^2H^2 + \frac{p }{2} \left(\frac{8nG^2H^2}{\omega} + \frac{5\omega n c_{{t}}}{4} \right) $ and defining $a_t = \mathbb{E} \Vert \Bar{\bx}^{(t)} - \bx^* \Vert^2 $, $Q= \frac{ \Bar{\sigma}^2}{n} $, $R =  8 \left(  {2L + \mu}  \right) \left( \frac{40}{p^2}+ \frac{80}{p\omega} + 1 \right)G^2H^2$, $U =  100 \left(  \frac{2L + \mu}{p}  \right)\omega  $   and $U_t =U c_t $, 
we get the recursion:
\begin{align*}
	a_{t+1} \leq \left(1- \frac{\mu \eta_t}{2} \right) a_{t} - \eta_t e_t + \eta_t^2Q + \eta_t^3R + \eta_t^3U_t 
\end{align*}
Employing a modified version of \cite[Lemma 3.3]{stich_sparsified_2018}, which is provided in the appendix, gives
\begin{align*}
	\frac{1}{S_T} \sum_{t=0}^{T-1}w_t e_t \leq \frac{\mu a^3}{8S_T}a_0 + \frac{4T(T+2a)}{\mu S_T}Q + \frac{64T}{\mu^2 S_T}R + \frac{64c_0T^{(2-\epsilon)}}{\mu^2 (2-\epsilon) S_T }U ,
\end{align*}
where we've used that $c_t \leq c_0 t^{(1-\epsilon)}$ for $c_0 \geq 0$ and some $\epsilon \in (0,1)$, $w_t = (a+t)^2$ and $S_T = \sum_{t=0}^{T-1}w_t \geq \frac{T^3}{3}$.
Using convexity of the global objective $f$ in the above inequality gives
\begin{align*}
\mathbb{E}f(\bar{\bx}_{avg}^{(T)}) - f^* \leq \frac{\mu a^3}{8S_T}a_0 + \frac{4T(T+2a)}{\mu S_T}Q + \frac{64T}{\mu^2 S_T}R + \frac{64c_0T^{(2-\epsilon)}}{\mu^2 (2-\epsilon) S_T }U,
\end{align*}
where $\bar{\bx}_{avg}^{(T)} = \frac{1}{S_T} \sum_{t=0}^{T-1} w_t \bar{\bx}^{(t)} $. 
Substituting the values of $a_0,Q,R,U$ in the above inequality gives the result of Theorem \ref{thm_cvx_li}.

Now we give a proof sketch of Lemma \ref{lem_dec_li_sgd}, which states that 
$e_{I_{(t)}}^{(1)}:=\sum_{j=1}^n \mathbb{E} \left\Vert \bar{\bx}^{I_{(t)}} - \bx^{I_{(t)}}_{j} \right\Vert^2$ 
-- the difference between local and the average iterates at the synchronization indices -- 
decays asymptotically to zero for decaying learning rate $\eta_t$.
We show this by setting up a contracting recursion for $e_{I_{(t)}}^{(1)}$. First we prove that 
\begin{align}
e_{I_{(t+1)}}^{(1)} \leq (1-\alpha_1)e_{I_{(t)}}^{(1)} + (1-\alpha_1)e_{I_{(t)}}^{(2)} + c_1\eta_{I_{(t)}}^2, \label{eq:dec_avg_local-interim1}
\end{align}
where $e_{I_{(t)}}^{(2)}:=\sum_{j=1}^n \mathbb{E} \left\Vert \hat{\bx}^{I_{(t+1)}} - \bx^{I_{(t)}}_{j} \right\Vert^2$, $\alpha_1\in(0,1)$, and $c_1$ is a constant that depends on $n,\delta,H,G$. 
Note that \eqref{eq:dec_avg_local-interim1} gives a contracting recursion in $e_{I_{(t)}}^{(1)}$, but it also gives the other term $e_{I_{(t)}}^{(2)}$, which we have to bound. 
It turns out that we can prove a similar inequality for $e_{I_{(t)}}^{(2)}$ as well:
\begin{align}
e_{I_{(t+1)}}^{(2)} \leq (1-\alpha_2)e_{I_{(t)}}^{(1)} + (1-\alpha_2)e_{I_{(t)}}^{(2)} + c_2(t)\eta_{I_{(t)}}^2, \label{eq:dec_avg_local-interim2}
\end{align}
where $\alpha_2\in(0,1)$; furthermore, we can choose $\alpha_1,\alpha_2$ such that $\alpha_1+\alpha_2 > 1$.
In \eqref{eq:dec_avg_local-interim2}, $c_2(t)$, in addition to $n,\delta,H,G$, also depends on the compression factor $\omega$ and  $c_t$ which is the triggering threshold at timestep $t$.

\begin{remark}
Note that \cite{koloskova_decentralized_2019-1} also proved analogous inequalities \eqref{eq:dec_avg_local-interim1} and \eqref{eq:dec_avg_local-interim2} with constants $c_1=c_2=0$. Here $c_1,c_2(t)$ are non-zero (with $c_2(t)$ possibly varying with t) and arise due to the use of local iterations and event-triggered communication,
which make the proof of these inequalities (in particular, the inequality \eqref{eq:dec_avg_local-interim2}) significantly more involved than the corresponding inequalities in \cite{koloskova_decentralized_2019-1}.
\end{remark}

Define $e_{I_{(t)}}:=e_{I_{(t)}}^{(1)} + e_{I_{(t)}}^{(2)}$. Adding \eqref{eq:dec_avg_local-interim1} and \eqref{eq:dec_avg_local-interim2} gives the following recursion with $\alpha\in(0,1)$:
\begin{align}
e_{I_{(t+1)}} \leq (1-\alpha)e_{I_{(t)}} + c_3(t)\eta_{I_{(t)}}^2. \label{eq:dec_avg_local-interim3}
\end{align}
From \eqref{eq:dec_avg_local-interim3}, we can show that $e_{I_{(t)}}\leq c(t)\eta_{I_{(t)}}^2$ for some $c(t)$ that depends on $n,\delta,H,G,\omega,c_t$. Lemma \ref{lem_dec_li_sgd} follows from this because $\sum_{j=1}^n \mathbb{E} \left\Vert \bar{\bx}^{I_{(t)}} - \bx^{I_{(t)}}_{j} \right\Vert^2 = e_{I_{(t)}}^{(1)} \leq e_{I_{(t)}}$. See Appendix~\ref{proof_lem_dec_li_sgd} for a complete proof of Lemma \ref{lem_dec_li_sgd}.
\qed

\subsection{Proof Outline of Theorem~\ref{thm_noncvx_fix_li}}
Note that \eqref{eq:thm_cvx_interim0} holds irrespective to the learning rate schedule. So, by substituting $\eta_t$ with $\eta$ in \eqref{eq:thm_cvx_interim0}, we get
\begin{align*}
\bar{\bx}^{(t+1)} =  \bar{\bx}^{(t)} - \frac{\eta}{n}\sum_{j=1}^n \nabla F_j(\bx_j^{(t)},\xi_j^{(t)}).
\end{align*}
With some algebraic manipulations given in Appendix \ref{proof_thm_noncvx_fix_li}, we have the following sequence relation for $\{f(\bar{\bx}^{(t)})\}$:
\begin{align} \label{eq:thm_noncvx_interim1}
\mathbb{E}[f(\bar{\bx}^{(t+1)})] 
 \leq  \mathbb{E}f(\bar{\bx}^{(t)}) - \frac{\eta}{4} \mathbb{E}\Vert \nabla f(\bar{\bx}^{(t)})  \Vert_2^2 + \frac{L\eta^2 \bar{\sigma}^2}{n}  +  \left[ \frac{\eta L^2}{2n} + \frac{2L^3\eta^2}{n} \right] \sum_{j=1}^n \mathbb{E} \Vert \bar{\bx}^{(t)} - \bx_j^{(t)} \Vert^2 
\end{align}
where expectation is taken over the entire process.
Let $I_{(t_0)}$ be the last synchronization index in $\mathcal{I}_T$ before time $t$. 
Note that $t - I_{(t_0)} \leq H$.
Similar to \eqref{eq:thm_cvx_interim4}, we can also bound the last term on the RHS of \eqref{eq:thm_noncvx_interim1} as 
(by replacing $\eta_{I_{t_{(0)}}} $ in \eqref{eq:thm_cvx_interim4} by $\eta$)
\begin{align}  \label{eq:thm_noncvx_interim2}
\sum_{j=1}^n \mathbb{E} \Vert \bar{\bx}^{{(t)}} - \bx^{{(t)}}_{j} \Vert^2 &\leq  2\mathbb{E}\sum_{j=1}^n \Vert \bar{\bx}^{I_{(t_0)}} - \bx^{I_{(t_0)}}_{j} \Vert^2 + 2n \eta^2H^2G^2
\end{align}
We can use the following lemma to bound the first term in the RHS of \eqref{eq:thm_noncvx_interim2}.
This lemma is analogous to Lemma~\ref{lem_dec_li_sgd} in the fixed learning rate. Observe that if we simply replace $\eta_{I_{(t_0)}}$ with $\eta$ in 
the bound of Lemma~\ref{lem_dec_li_sgd}, we would get a slightly weaker bound than what we obtain in the following lemma, which we prove in Appendix~\ref{proof_lemm_dec_li_sgd_fix}
\begin{lemma}[Bounded deviation of local iterates and the averaged iterates] \label{lemm_dec_li_sgd_fix}
Under the assumptions of Theorem~\ref{thm_noncvx_fix_li}, for any $I_{(t)}$ such that $I_{(t)}\in\I_T$, we have
\begin{align*}
	\sum_{j=1}^n \mathbb{E}\Vert \bar{\bx}^{I_{(t)}} - \bx^{I_{(t)}}_{j} \Vert^2 \leq \frac{4A\eta^2}{p^2},
\end{align*}
where $A=2nG^2H^2 + \frac{p }{2} \left(\frac{8nG^2H^2}{\omega} + \frac{5\omega n }{4 \eta^{(1-\epsilon)} } \right)$. 
\end{lemma}
Using the bound from Lemma~\ref{lemm_dec_li_sgd_fix} in \eqref{eq:thm_noncvx_interim2} gives $\sum_{j=1}^n \mathbb{E} \left\Vert \bar{\bx}^{{(t)}} - \bx^{{(t)}}_{j} \right\Vert^2 \leq C:= \frac{8A}{p^2} \eta^2 + 2n \eta^2H^2G^2$.
Note that for the case of fixed learning rate $\eta$, we have to fix the time horizon (the number of iterations) $T$ before the algorithm begins.
By setting $\eta=\sqrt{\frac{n}{T}}$ and $T \geq 64nL^2$, we get $\eta \leq \frac{1}{8L}$.
Now, substituting the bound on $\sum_{j=1}^n \mathbb{E} \left\Vert \bar{\bx}^{{(t)}} - \bx^{{(t)}}_{j} \right\Vert^2$ and $\eta \leq \frac{1}{8L}$ in \eqref{eq:thm_noncvx_interim1}, rearranging terms, and then summing from $t=0$ to $T-1$ gives:
\begin{align*}
\sum_{t=0}^{T-1} \eta  \mathbb{E} \Vert \nabla f(\bar{\bx}^{(t)}) \Vert_2^2\ \leq\ 4 \left( f(\bar{\bx}^{(0)}) - \mathbb{E} f(\bar{\bx}^{(t)}) \right)  + \frac{2L^2C}{n} { \sum_{t=0}^{T-1}\eta^3} + \frac{8L^3C}{n}{ \sum_{t=0}^{T-1}\eta^4} + \frac{4L \bar{\sigma}^2}{n} {\sum_{t=0}^{T-1}\eta^2}
\end{align*}
Dividing both sides by $\eta T$, setting $\eta = \sqrt{\frac{n}{T}} $ and substituting the value of $A$ proves Theorem~\ref{thm_noncvx_fix_li}.
\qed

\section{Experiments}\label{experiments}

In this section, we compare SPARQ-SGD with CHOCO-SGD (\cite{koloskova_decentralized_2019-1,koloskova_decentralized_2019}), which only employs compression (sparsification {\em or} quantization) and is state-of-the-art in communication efficient decentralized training.

\begin{figure*}[htb]
	\begin{subfigure}{0.5\linewidth}
		\centerline{\includegraphics[scale=0.5]{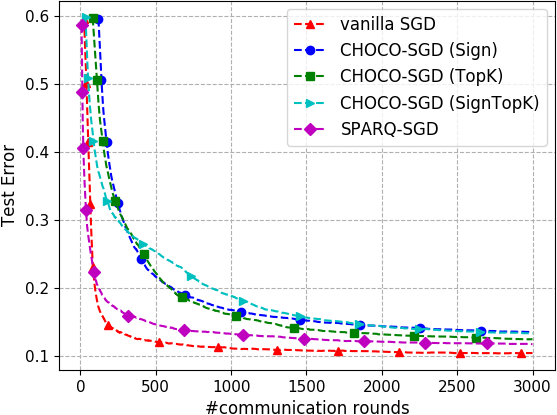}}
		\caption{}
		\label{fig:lo-cvx}
	\end{subfigure}
	\begin{subfigure}{0.5\linewidth}
		\centerline{\includegraphics[scale=0.5]{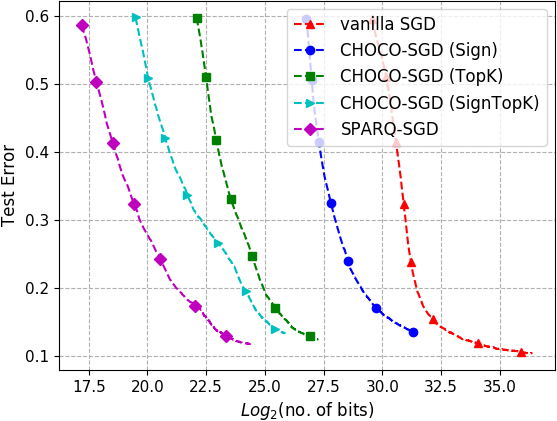}}
		\caption{}
		\label{fig:te-logbits-cvx} \end{subfigure}
	\begin{subfigure}{0.5\linewidth}
		\centerline{\includegraphics[scale=0.5]{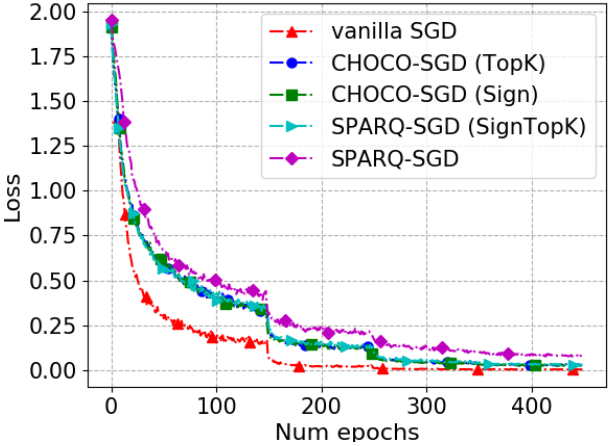}}
		\caption{}
		\label{fig:lo-noncvx}
	\end{subfigure}\hfill
	\begin{subfigure}{0.5\linewidth}
		\centerline{\includegraphics[scale=0.5]{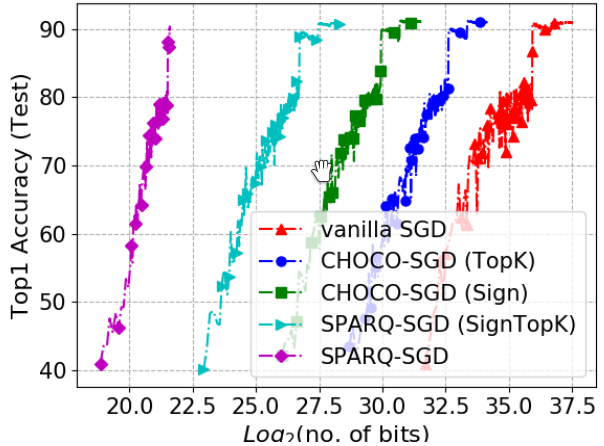}}
		\caption{}
		\label{fig:te-logbits-noncvx}
	\end{subfigure}
	~\label{Fig4_2}
	\caption{Figure~\ref{fig:lo-cvx} and \ref{fig:te-logbits-cvx} are for convex objective and we plot test error vs number of communication rounds and test error vs total number of bits communicated, respectively, for different algorithms. Figure~\ref{fig:lo-noncvx} and \ref{fig:te-logbits-noncvx} are for non-convex objective where we plot training loss vs epcohs and Top-1 accuracy vs total number of bits communicated, respectively.}
\end{figure*}

\subsection{Convex objectives}
We run SPARQ-SGD on MNIST dataset and use multi-class cross-entropy loss to model the local objectives $f_i, i\in[n]$. We consider $n=60$ nodes connected in a ring topology, each processing a mini-batch size of 5 per iteration and having heterogeneous distribution of data across classes.

The learning rate is $\eta_t = b/(t+a)$, where the hyper-parameter $b$ is tuned via grid search. We take $a=5H/p$, as in Theorem $\ref{thm_cvx_li}$, where $p= \delta\gamma/8$ 
and $H$ denotes the synchronization period. Specifically, we work with $\eta_t = 1/(t+100)$ and $H=5$. For compression, we use the composed operator $SignTopK$ \cite{basu_qsparse-local-sgd:_2019} with $k=10$ (out of 7840 length vector for MNIST dataset)
For our experiments, we initially set the triggering constant equal to 5000 in SPARQ-SGD (line 7) and keep it unchanged until a certain number of iterations and then increase it periodically under assumptions of Theorem \ref{thm_cvx_li}; this is to prevent all the workers satisfying the triggering criterion $c_0\eta_t^2$ in later iterations, as $\eta_t$ eventually becomes very small.

$\bullet$ {\bf Results.}
We use $SignTopK$ compression in SPARQ-SGD and compare its performance against CHOCO-SGD.
In Figure \ref{fig:lo-cvx}, we observe 
 SPARQ-SGD can reach a target test error in fewer communication rounds while converging at a rate similar to that of vanilla SGD.
The advantage to SPARQ-SGD comes from the significant savings in the number of bits communicated to achieve a desired test error, as seen in Figure \ref{fig:te-logbits-cvx}: to achieve a test error of around 0.12, SPARQ-SGD gets 250$\times$ savings as compared to CHOCO-SGD with $Sign$ quantizer, around 10-15$\times$ savings than CHOCO-SGD with $TopK$ sparsifier, and around 1000$\times$ savings than vanilla decentralized SGD. 
We also implement the composed operator $SignTopK$ in the CHOCO-SGD framework for comparison, though it was not done in that paper. 


\subsection{Non-convex objectives}
We match the setting in CHOCO-SGD 
and perform our experiments on the CIFAR-10 \cite{cifar} dataset and train a Resnet-20 \cite{wen2016learning} model with $n=8$ nodes connected in a ring topology. We use a learning rate schedule consisting of a warmup period of 5 epochs followed by a piecewise decay of 5 at epoch 150 and 250 and stop training at epoch 450. The SGD algorithm is implemented with momentum with a factor of 0.9 and mini-batch size of 128. 
SPARQ-SGD consists of $H=5$ local iterations followed by checking for a triggering condition, and then communicating with the composed $SignTopK$ operator, 
where we take top 10\% elements of each tensor and only transmit the sign and norm of the result. The triggering threshold follows a schedule piecewise constant: initialized to 2.0 and increases by 1.0 after every 10 epochs till 60 epochs are complete. We compare performance of SPARQ-SGD against CHOCO-SGD with $Sign$, $TopK$ compression 
(taking top 10\% of elements of the tensor) and decentralized vanilla SGD \cite{lian2017can}. 
We also provide a plot for using the composed $SignTopK$ operator without event-triggering titled `SPARQ-SGD (Sign-TopK)' for comparison.

%
$\bullet$ {\bf Results.}
We plot global loss function evaluated at average parameter across nodes in Figure \ref{fig:lo-noncvx}, where we observe SPARQ-SGD converging at a similar rate to CHOCO-SGD and vanilla decentralized SGD. Figure \ref{fig:te-logbits-noncvx} shows the performance for a given bit-budget, where we show the Top-1 test accuracy
as a function of the total bits communicated. For Top-1 test accuracy of around 90\%, SPARQ-SGD requires about 250$\times$ less bits than CHOCO-SGD with $Sign$ compression, about 1000$\times$ less bits than CHOCO-SGD with $TopK$ compression, and around 15K$\times$ less bits than vanilla decentralized SGD to achieve the same Top-1 accuracy.

\section{Conclusion}\label{conclusion}
We propose SPARQ-SGD, a communication efficient algorithm for decentralized learning. 
The efficiency stems from employing compression to the exchanged updates and initiating communication only when a locally computable triggering condition at a node is satisfied; 
specifically, a node triggers communication when it observes a significant change in its local model parameter vector (since the last time communication occurred) after completing a fixed number of local gradient steps. 
We develop our convergence analyses for strongly convex and non-convex objectives, and show that the proposed algorithm achieves the same rate as vanilla decentralized SGD in each of these settings. 
Our experiments demonstrate that SPARQ-SGD saves significant bits in communication over the state-of-the-art without compromising much in accuracy.
We leave incorporating momentum in our algorithm and removing the bounded gradient assumption in our analyses as future extensions to this work.

\section*{Acknowledgments}
This work was partially supported by NSF grant \#1514531, by UC-NL grant LFR-18-548554 and by Army Research Laboratory under Cooperative Agreement W911NF-17-2-0196. 
The views and conclusions contained in this document are those of the authors and should not be interpreted as representing the official policies, either expressed or implied, of the Army Research Laboratory or the U.S. Government. The U.S. Government is authorized to reproduce and distribute reprints for Government purposes notwithstanding any copyright notation here on.
\bibliography{ref} 
\bibliographystyle{alpha}

\newpage
\onecolumn
\appendix
\section{Some Helpful Facts}
\subsection{Vector and Matrix inequalities}
\begin{fact}
	Let $\mathbf{M} \in \mathbb{R}^{p \times q} $ be a matrix with entries $[m_{ij}] $, $i \in [p], j \in [q]$. The Frobenius norm of $\mathbf{M}$ is given by : $$ \verts{\mathbf{M}}_F  =  \sqrt{ \sum\limits_{i=1}^{p} \sum\limits_{j=1}^{q} \vert m_{ij} \vert^2  } $$ 
	Consider any two matrices $ \mathbf{A} \in \mathbb{R}^{d \times n}$, $\mathbf{B} \in \mathbb{R}^{n \times n}$. Then the following holds:
	\begin{align} \label{bound_frob_mult}
		\Vert \mathbf{AB} \Vert_F \leq \Vert \mathbf{A} \Vert_F \Vert \mathbf{B} \Vert_2
	\end{align}
\end{fact}
\begin{fact}
	For any set of $n$ vectors $ \{ \mathbf{a_1} ,\hdots , \mathbf{a_n}    \}$ where $\mathbf{a_i} \in \mathbb{R}^d$, we have:
	\begin{align} \label{bound_seq_sum}
		\verts{\sum_{i=1}^{n} \mathbf{a_i}}^2 \leq n \sum_{i=1}^{n} \verts{\mathbf{a_i}}^2
	\end{align}
\end{fact}
\begin{fact}
	For any two vectors $\mathbf{a},\mathbf{b} \in \mathbb{R}^d$, for all $\gamma >0$, we have:
	\begin{align} \label{bound_inner_prod}
		2 \lragnle{\mathbf{a},\mathbf{b}} \leq \gamma \verts{\mathbf{a}}^2 + \gamma^{-1} \verts{\mathbf{b}}^2
	\end{align}
\end{fact}
\begin{fact} \label{bound_l2_sum}
	For any two vectors $\mathbf{a},\mathbf{b} \in \mathbb{R}^d$, for all $\alpha >0$, we have:
	\begin{align} 
	\verts{\mathbf{a}+\mathbf{b}}^2 \leq (1+\alpha) \verts{\mathbf{a}}^2 + {(1 + \alpha^{-1})} \verts{\mathbf{b}}^2
	\end{align}
	Similar inequality holds for matrices in Frobenius norm, i.e., for any two matrices $\mathbf{A},\mathbf{B} \in \mathbb{R}^{p \times q} $ and for any $\alpha >0$ , we have
	\begin{align*}
	\verts{\mathbf{A} + \mathbf{B} }_F^2 \leq (1 + \alpha) \verts{\mathbf{A}}_F^2 + (1 + \alpha^{-1})\verts{\mathbf{B}}_F^2		
	\end{align*}
\end{fact}
\subsection{Properties of functions}
\begin{definition}[Smoothness]
	A differentiable function $f : \mathbb{R}^d \rightarrow \mathbb{R}$ is L-smooth with parameter $L \geq 0$ if
	\begin{align} \label{l_smooth}
	f(\by) \leq f(\bx) + \langle \nabla f(\bx), \by-\bx  \rangle + \frac{L}{2} \Vert \by-\bx \Vert^2, \hspace{2cm} \forall \bx,\by \in \mathbb{R}^d
	\end{align}
\end{definition}
\begin{definition}[Strong convexity]
	A differentiable function $f : \mathbb{R}^d \rightarrow \mathbb{R}$ is $\mu$-strongly convex with parameter $\mu \geq 0$ if
	\begin{align} \label{mu_strong_cvx}
	f(\by) \geq f(\bx) + \langle \nabla f(\bx), \by-\bx  \rangle + \frac{\mu}{2} \Vert \by-\bx \Vert^2, \hspace{2cm} \forall \bx,\by \in \mathbb{R}^d
	\end{align}
\end{definition}
\begin{lemma}
	Let $f$ be an $L$-smooth function with global minimizer $\bx^*$. We have
	\begin{align} \label{l_smooth_prop}
	\Vert \nabla f(\bx) \Vert^2 \leq 2L( f(\bx) - f(\bx^*) ).
	\end{align}
	
\end{lemma}
\begin{proof}
	By definition of $L$-smoothness, we have
	\begin{align*}
	f(\by) & \leq f(\bx) + \langle \nabla f(\bx), \by-\bx \rangle + \frac{L}{2}\Vert \by-\bx \Vert^2. \\
	\intertext{Taking infimum over y yields:}
	\inf_{\by} f(\by) & \leq \inf_{\by} \left( f(\bx) + \langle \nabla f(\bx), \by-\bx \rangle + \frac{L}{2}\Vert \by-\bx \Vert^2 \right) \\
	& \stackrel{\text{(a)}}{=}  \inf_{\bv: \Vert \bv \Vert = 1} \inf_t \left( f(\bx) + t \langle \nabla f(\bx), \bv \rangle + \frac{L t^2}{2} \right) \\
	& \stackrel{\text{(b)}}{=}  \inf_{\bv: \Vert \bv \Vert = 1} \left( f(\bx) - \frac{1}{2L} \langle \nabla f(\bx),\bv \rangle^2  \right) \\
	& \stackrel{\text{(c)}}{=}  \left( f(\bx) - \frac{1}{2L} \Vert  \nabla f(\bx) \Vert^2  \right) \\
	\end{align*}  
The value of $t$ that minimizes the RHS of (a) is $t=-\frac{1}{L}\langle \nabla f(\bx), \bv \rangle$, this implies (b);
(c) follows from the Cauchy-Schwartz inequality: $\langle \bu, \bv \rangle \leq \| \bu \| \| \bv \|$, where equality is achieved whenever $u=v$.
Now, substituting $\inf \limits_{\by} f(\by) = f(\bx^*)$ in the RHS of (c) yields the result.
\end{proof}
\subsection{Matrix form notation} \label{mat_not_sec}
Consider the set of parameters $\{ \bx_i^{(t)} \}_{i=1}^{n} $  at the nodes at timestep $t$ and  estimates of the parameter $\{ \hat{\bx}_i^{(t)} \}_{i=1}^{n} $. The matrix notation is given by : 
	$$ \bX^{(t)} := [\bx_1^{(t)}, \hdots, \bx_n^{(t)} ] \in \mathbb{R}^{d \times n }, \hspace{0.5cm} \hat{\bX}^{(t)} := [\hat{\bx}_1^{(t)}, \hdots, \hat{\bx}_n^{(t)} ] \in \mathbb{R}^{d \times n }, \hspace{0.5cm}  \Bar{\bX}^{(t)} := [\Bar{\bx}^{(t)}, \hdots, \Bar{\bx}^{(t)} ] \in \mathbb{R}^{d \times n} $$
	$$  {\partial F}(\bX^{(t)}, \boldsymbol{\xi}^{(t)}) := [ {\nabla F_1}(\bx_1^{(t)}, \xi_1^{(t)}), \hdots, {\nabla F_n}(\bx_n^{(t)}, \xi_n^{(t)}) ]  \in \mathbb{R}^{d \times n} $$
	Where $\nabla F_i(\bx_i^{(t)}, \xi_i^{(t)})$ denotes the stochastic gradient at node $i$ at timestep $t$ and the vector $  \Bar{\bx}^{(t)} $  denotes the average of node parameters at time $t$, specifically : $ \Bar{\bx}^{(t)} := \frac{1}{n}\sum_{i=1}^{n} \bx_i^{(t)}  $.\\
	 Let $\Gamma^{(t)} \subseteq [n] $ be the set of nodes that do not communicate at time $t$. We define $\mathbf{P}^{(t)} \in \mathbb{R}^{n \times n} $,  a diagonal matrix with $\mathbf{P}_{ii}^{(t)} = 0$ for $i \in \Gamma^{(t)} $ and $\mathbf{P}_{ii}^{(t)} = 1$ otherwise.\\ \\
\paragraph{SPARQ-SGD  in Matrix notation}  \label{mat_not_li} \mbox{}\\
{	Consider  Algorithm \ref{alg_dec_sgd_li}  with synchronization indices given by the set $\{ I_{(1)},I_{(2)}, \hdots, I_{(t)}, \hdots  \}$. Using the above notation, the sequence of parameters updates from synchronization index $I_{(t)}$ to $I_{(t+1)}$ is given by:
	\begin{align*}
	\bX^{I_{(t+\frac{1}{2})}} & = \bX^{I_{(t)}} - \sum_{t' = I_{(t)}}^{I_{(t+1)}-1} \eta_{t'} {\partial F}(\bX^{(t')},\boldsymbol{\xi}^{(t')}) \\
	\hat{\bX}^{I_{(t+1)}} & = \hat{\bX}^{I_{(t)}} + \C((\bX^{I_{(t+\frac{1}{2})}} - \hat{\bX}^{I_{(t)}})\mathbf{P}^{(I_{(t+1)}-1)} ) \\
	\bX^{I_{(t+1)}} & = \bX^{I_{(t+\frac{1}{2})}} + \gamma\hat{\bX}^{I_{(t+1)}}(\mathbf{W}-\mathbf{I}) 
	\end{align*}
	where $\C( . )$ denotes the contraction operator applied column-wise to the argument matrix and $\mathbf{I}$ is the identity matrix. } \mbox{}\\ \\
We now note some useful properties of the iterates in matrix notation which would be used throughout the paper:
\begin{enumerate}
	\item 	If $\mathbf{W}$ is a doubly stochastic matrix :  $\mathbf{W} \in [0,1]^{n \times n}, \, \mathbf{W}=\mathbf{W}^T \, , \mathbf{W} \mathbf{\mathbbm{1}} = \mathbf{\mathbbm{1}} \, , \mathbf{\mathbbm{1}}^T\mathbf{W} = \mathbf{\mathbbm{1}}^T. $ 
	\begin{align} \label{mean_prop}
	\Bar{\bX}^{(t)} = \bX^{(t)} \frac{1}{n} \mathbf{\mathbbm{1}}\mathbf{\mathbbm{1}}^T, \hspace{1cm} \Bar{\bX}^{(t)}\mathbf{W} = \Bar{\bX}^{(t)}
	\end{align}
	Where the first expression follows from the definition of $\bar{\bX}^{(t)}$ and the second expression follows from $\mathbf{W} \frac{\mathbf{\mathbbm{1}} \mathbf{\mathbbm{1}}^T }{n} =  \frac{\mathbf{\mathbbm{1}} \mathbf{\mathbbm{1}}^T}{n}  $ as $\mathbf{W}$ is a doubly stochastic matrix and the fact that $\mathbf{W} \frac{\mathbf{\mathbbm{1}} \mathbf{\mathbbm{1}}^T }{n} = \frac{\mathbf{\mathbbm{1}} \mathbf{\mathbbm{1}}^T }{n} \mathbf{W} $. \\
	\item The average of the iterates in Algorithm \ref{alg_dec_sgd_li}  follows  : 
	\begin{align} \label{mean_seq_iter}
	\Bar{\bX}^{(t+1)} & = \Bar{\bX}^{(t+ \frac{1}{2})} + \mathbf{\mathbbm{1}}_{(t+1) \in \mathcal{I}_T} \left[\gamma\hat{\bX}^{{(t+1)}}( \mathbf{W}- \mathbf{I})\frac{1}{n} \mathbf{\mathbbm{1}}\mathbf{\mathbbm{1}}^T \right] = \,  \Bar{\bX}^{(t+ \frac{1}{2})} 
	\end{align}
	where $\mathcal{I}_T $ denotes the set of synchronization indices of Algorithm \ref{alg_dec_sgd_li}. The above follows from the observation that $\mathbf{W} \frac{\mathbf{\mathbbm{1}} \mathbf{\mathbbm{1}}^T }{n} =\frac{\mathbf{\mathbbm{1}} \mathbf{\mathbbm{1}}^T }{n} $ as $\mathbf{W}$ is a doubly stochastic matrix.	
\end{enumerate}
\subsection{Assumptions and useful facts}
\begin{assumption}  
	(Bounded Gradient Assumption)
	We assume that the expected stochastic gradient for any worker has a bounded second moment; specifically, for all $i \in [n]$ with stochastic sample $\xi_i$ and any $\bx \in \mathbb{R}^d $,  we have:
	\begin{align*}
		\mathbb{E}_{\xi_i} \verts{\nabla F_i (\bx, \xi_i)}^2 \leq G^2 
	\end{align*}
	Using the matrix notation established above, for all \bX, the second moment of $\partial F(\bX, \mathbf{\xi})$ is bounded as:
	\begin{align} \label{bound_grad}
	\mathbb{E}_{\xi} \verts{\partial F(\bX, \xi)}^2_F \leq nG^2 
	\end{align}		
\end{assumption}
\begin{assumption} \label{bound_var}
	(Variance bound for workers) 
	Consider the variance bound on the stochastic gradient for nodes $i \in [n]$ : $\mathbb{E}_{\xi_i} \verts{\nabla F_i (\bx, \xi_i) - \nabla f_i (\bx) }^2 \leq \sigma_i^2 $ where $\mathbb{E}_{\xi_i} [\nabla F_i(\bx, \xi_i)] = \nabla f_i(\bx) $, then:
	\begin{align*}
		\mathbb{E}_{\boldsymbol{\xi}^{(t)}} \left\Vert \frac{1}{n}\sum_{j=1}^n \left(\nabla f_j(\bx_j^{(t)}) - \nabla F_j(\bx_j^{(t)},\xi_j^{(t)}) \right)\right\Vert^2 \leq \frac{\Bar{\sigma}^2}{n}
	\end{align*}
	where ${\boldsymbol{\xi}^{(t)}} = \{ \xi_1^{(t)}, \xi_2^{(t)}, \hdots, \xi_n^{(t)} \}$ denotes the stochastic sample for the nodes at any timestep $t$ and $\frac{ \sum_{j=1}^n \sigma_j^2 }{n}= {\Bar{\sigma}^2}$
	\begin{proof}
		\begin{align*}
		 &\mathbb{E}_{\xi^{(t)}}\left\Vert \frac{1}{n}\sum_{j=1}^n \nabla f_j(\bx_j^{(t)}) - \frac{1}{n}\sum_{j=1}^n \nabla F_j(\bx_j^{(t)},\xi_j^{(t)}) \right\Vert^2  = 
		\frac{1}{n^2} \sum_{j=1}^n \mathbb{E}_{\xi^{(t)}} \Vert \nabla f_j(\bx_j^{(t)}) - \nabla F_j(\bx_j^{(t)},\xi_j^{(t)})\Vert^2 \\
		& \hspace{4.1cm} + \frac{1}{n^2}\sum_{i \neq j} \mathbb{E}_{\xi^{(t)}} \left\langle \nabla f_i(\bx_i^{(t)}) - \nabla F_i(\bx_i^{(t)},\xi_j^{(t)}), \nabla f_j(\bx_j^{(t)}) - \nabla F_j(\bx_j^{(t)},\xi_j^{(t)}) \right\rangle 
		\end{align*}
	Since $\xi_i$ is independent of $\xi_j$, the second term is zero in expectation, thus the above reduces to:
		\begin{align*}
		\mathbb{E}_{\xi^{(t)}}\left\Vert \frac{1}{n}\sum_{j=1}^n \nabla f_j(\bx_j^{(t)}) - \frac{1}{n}\sum_{j=1}^n \nabla F_j(\bx_j^{(t)},\xi_j^{(t)}) \right\Vert^2 & = \frac{1}{n^2} \sum_{j=1}^n \mathbb{E}_{\xi^{(t)}} \Vert \nabla f_j(\bx_j^{(t)}) - \nabla F_j(\bx_j^{(t)},\xi_j^{(t)})\Vert^2 \\
		& \leq \frac{1}{n^2} \sum_{j=1}^n \sigma_j^2 = \frac{\Bar{\sigma}^2}{n}
		\end{align*}
	\end{proof}
\end{assumption}
\begin{definition} (Compression Operator \cite{stich_sparsified_2018} ) \C : $\mathbb{R}^d \rightarrow \mathbb{R}$ is called a compression operator if it satisfies :
	\begin{align} \label{bound_comp}
	\mathbb{E}_\C \verts{\C(\bx) - \bx}^2 \leq (1- \omega ) \verts{\bx}^2, \hspace{1cm} \forall \bx \in \mathbb{R}^d
	\end{align}
	for a parameter $\omega > 0 $. $\mathbb{E}_\C$ denotes the expectation over internal randomness of operator $\C$. 
\end{definition}
\begin{fact} (Triggering rule)
	Consider the set of nodes $ {\Gamma}^{(t)}  $ which do not communicate at time $t$. The triggering rule dictates : $$\Vert \bx_i^{(t+\frac{1}{2})} - \hat{\bx}_i^{(t)} \Vert_2^2 \leq  {c_t \eta_t^2}  \hspace{1cm} \forall i \in \Gamma^{(t)}$$ for Algorithm \ref{alg_dec_sgd_li} and threshold sequence $\{c_t\}_{t=0}^{T-1}$. Using notation from section \ref{mat_not_sec}, this is translated as:
	\begin{align} \label{bound_trig}
	\verts{ (\bX^{(t+\frac{1}{2})}  - \hat{\bX}^{(t)}) (\mathbf{I} - \mathbf{P}^{(t)}) }_F^2 \leq nc_t \eta_t^2
	\end{align}
\end{fact}
\begin{fact} 
	For doubly stochastic matrix $\mathbf{W}$ with second largest eigenvalue $1- \delta = |\lambda_2(\mathbf{W}) | <1 $:
	\begin{align} \label{bound_W_mat}
	\left\Vert \mathbf{W}^k - \frac{1}{n} \mathbf{\mathbbm{1}}\mathbf{\mathbbm{1}}^T  \right\Vert_2 = (1-\delta)^k
	\end{align}
	for any non-negative integer $k$. The proof follows from Lemma 16 in \cite{koloskova_decentralized_2019-1}
\end{fact}
\begin{fact} 
	Consider the set of synchronization indices in Algorithm \ref{alg_dec_sgd_li} : $\{ I_{(1)},I_{(2)}, \hdots, I_{(k)}, \hdots  \}  \in \mathcal{I}_T $. We assume that the maximum gap between any two consecutive elements in  $\mathcal{I}_T$ is bounded by $H$. Let ${\xi^{(t)}} = \{ \xi_1^{(t)}, \xi_2^{(t)}, \hdots, \xi_n^{(t)} \}$ denote the stochastic samples for the nodes at any timestep $t$. Consider any two consecutive synchronization indices $I_{(k)}$ and $I_{(k+1)}$ and define $\xi :=  \{  \xi^{(t')} :  I_{(k)} \leq t' \leq I_{(k+1)}  \} $. Then using (\ref{bound_grad}), we have:
	\begin{align} \label{bound_gap_grad}
	\mathbb{E}_{\xi} \left[\left\Vert \sum_{t' = I_{(k)}}^{I_{(k+1)}-1}\eta_{t'} \partial F (\bX^{t'}, \mathbf{\xi}^{t'}) \right\Vert_F^2\right] \leq \eta_{I_{(k)}}^2 H^2 n G^2
	\end{align}
\end{fact}

\section{Omitted details from Section \ref{main-results}}
We restate the sequence of updates for Algorithm \ref{alg_dec_sgd_li} in matrix form for reference (see Section \ref{mat_not_sec}):
\begin{align*}
\bX^{I_{(t+\frac{1}{2})}} & = \bX^{I_{(t)}} - \sum_{t' = I_{(t)}}^{I_{(t+1)}-1} \eta_{t'} \partial F(\bX^{(t')},\xi^{(t')}) \\
\hat{\bX}^{I_{(t+1)}} & = \hat{\bX}^{I_{(t)}} + \C((\bX^{I_{(t+\frac{1}{2})}} - \hat{\bX}^{I_{(t)}})\mathbf{P}^{(I_{(t+1)}-1)} ) \\
\bX^{I_{(t+1)}} & = \bX^{I_{(t+\frac{1}{2})}} + \gamma\hat{\bX}^{I_{(t+1)}}(\mathbf{W}-\mathbf{I}) 
\end{align*}
where $\{ I_{(1)},I_{(2)}, \hdots, I_{(t)}, \hdots  \}$ denote the synchronization indices and $\C( . )$ denotes the contraction operator applied elementwise to the argument matrix. Let $\Gamma^{(t)} \subseteq [n] $ be the set of nodes that do not communicate at time $t$. We define $\mathbf{P}^{(t)} \in \mathbb{R}^{n \times n} $,  a diagonal matrix with $\mathbf{P}_{ii}^{(t)} = 0$ for $i \in \Gamma^{(t)} $ and $\mathbf{P}_{ii}^{(t)} = 1$ otherwise. \\
The above equalities are used throughout the proofs in this section.
\subsection{Proof of Lemma \ref{lem_dec_li_sgd}} \label{proof_lem_dec_li_sgd}
\begin{lemma*} (Restating Lemma \ref{lem_dec_li_sgd})
		Let $ \{ \bx_t ^{(i)}  \}_{t=0}^{T-1} $ be generated according to Algorithm \ref{alg_dec_sgd_li} under assumptions of Theorem \ref{thm_cvx_li} with stepsize $\eta_t := \frac{b}{(a+t)}$ (where $a \geq \frac{5H}{p}, \, b > 0$), an increasing threshold fucntion $c_t \sim {o}(t)$ and define $\bar{\bx}_t = \frac{1}{n} \sum_{i=1}^{n} x_t^{(i)} $.
	Consider the set of synchronization indices $\mathcal{I}_T$ =  $\{ I_{(1)},I_{(2)}, \hdots, I_{(t)}, \hdots  \}$. Then for any $I_{(t)} \in \mathcal{I}_T$, we have:
	\begin{align*}
	\mathbb{E}\sum_{j=1}^n \left\Vert \bar{\bx}^{I_{(t)}} - \bx^{I_{(t)}}_{j} \right\Vert^2 = \mathbb{E} \Vert \bX^{I_{(t)}} - \Bar{\bX}^{I_{(t)}} \Vert_F^2  \leq \frac{20A_{I_{(t)}}  \eta_{I_{(t)}}^2}{p^2}
	\end{align*}
	for $p = \frac{\delta \gamma}{8} $, $\delta := 1 - | \lambda_2(\mathbf{W})|$, $\omega$ is compression parameter for operator $\C$, $A_{I_{(t)}}=2nG^2H^2 + \frac{p}{2} \left( \frac{8nG^2H^2}{\omega} + \frac{5\omega n c_{I_{(t)}}}{4} \right)$ and $c_{I_{(t)}}$ and $\eta_{I_{(t)}}$ are respectively the triggering threshold and learning rate evaluated at timestep $I_{(t)}$.
\end{lemma*}

Our proof for Lemma \ref{lem_dec_li_sgd} involves analyzing the following expression:
\begin{align*}
	e_{I_{(t+1)}} := \underbrace{\mathbb{E} \Vert \bX^{I_{(t+1)}} - \Bar{\bX}^{I_{(t+1)}} \Vert_F^2}_{e^{(1)}_{I_{(t+1)}}} + \underbrace{ \mathbb{E} \Vert \bX^{I_{(t+1)}} - \hat{\bX}^{I_{(t+2)}} \Vert_F^2}_{e^{(2)}_{I_{(t+1)}}}
\end{align*}
The first term $e^{(1)}_{I_{(t+1)}}$ in RHS above captures the deviation of local parameters of the nodes from the global average parameter. The second term $e^{(2)}_{I_{(t+1)}}$ captures the deviation of the node parameters from their copies. In the section for Proof Outlines \ref{proof-outlines}, we proposed the idea of writing a bound for 
$e^{(1)}_{I_{(t+1)}}$ (and $e^{(2)}_{I_{(t+1)}}$) in terms of $e^{(1)}_{I_{(t)}}$ (and $e^{(2)}_{I_{(t)}}$), enabling us to write a recursive expression which could then be translated to a recursive expression for the sum $e_{I_{(t+1)}}$ in terms of $e_{I_{(t)}}$. \\
We follow a different approach here, where we bound $e^{(1)}_{I_{(t+1)}}$ and $e^{(2)}_{I_{(t+1)}}$ individually in terms of parameters evaluated at the auxiliary index $I_{(t+\frac{1}{2})}$. These bounds are provided in Lemma \ref{lem_cvx_e1} and Lemma \ref{lem_cvx_e2} below. These individual bounds are then added to yield a bound for the sum $e_{I_{(t+1)}}$ in terms of parameters evaluated at the auxiliary index $I_{(t+\frac{1}{2})}$. 
Further simplification allows us to bound $e_{I_{(t+1)}}$ in terms of $e_{I_{(t)}}$, thus giving a recursive form.
 This recursion enables us to bound $e_{I_{(t+1)}}$ as $e_{I_{(t+1)}} \leq \frac{20A_{I_{(t+1)}}}{p^2} \eta_{I_{(t+1)}}^2$ where $A_{I_{(t+1)}}$ is defined in the statement of Lemma \ref{lem_dec_li_sgd}. Thus, the quantity of interest $e^{(1)}_{I_{(t+1)}} = \mathbb{E} \Vert \bX^{I_{(t+1)}} - \Bar{\bX}^{I_{(t+1)}} \Vert_F^2 $ is also bounded by $\frac{20A_{I_{(t+1)}}}{p^2} \eta_{I_{(t+1)}}^2$, proving Lemma \ref{lem_dec_li_sgd}. \\
 We first state and prove bounds on $e^{(1)}_{I_{(t+1)}}$ and $e^{(2)}_{I_{(t+1)}}$ in terms of parameters evaluated at the auxiliary index $I_{(t+\frac{1}{2})}$ in Lemma \ref{lem_cvx_e1} and \ref{lem_cvx_e2} respectively. Using these bounds, we then proceed to prove Lemma \ref{lem_dec_li_sgd}. \\
 
\begin{lemma} \label{lem_cvx_e1}
	Consider the sequence of updates in Algorithm \ref{alg_dec_sgd_li} in matrix form (refer \ref{mat_not_li}). The expected deviation between the local node parameter $\bX^{I_{(t+1)}}$ and the global average parameter $\Bar{\bX}^{I_{(t+1)}}$ evaluated at some $I_{(t+1)} \in \mathcal{I}_T$ satisfies:
	\begin{align*}
 	e^{(1)}_{I_{(t+1)}} = \mathbb{E} \Vert \bX^{I_{(t+1)}} - \Bar{\bX}^{I_{(t+1)}} \Vert_F^2 \leq (1+\alpha_1)(1-\gamma \delta)^2 \mathbb{E} \Vert \bX^{I_{(t+\frac{1}{2})}} - \Bar{\bX}^{I_{(t+\frac{1}{2})}} \Vert_F^2 \\
 	\qquad + (1+\alpha_1^{-1}) \gamma^2 \beta^2 \mathbb{E} \Vert  {\bX}^{I_{(t+\frac{1}{2})}}-\hat{\bX}^{I_{(t+1)}}\Vert_F^2
	\end{align*}
	where $\alpha_1 >0$ is a constant, $\delta$ is the spectral gap, $\gamma$ is the consensus stepsize and $\beta = \verts{\mathbf{W}-\mathbf{I}}_2$ where $\mathbf{W}$ a doubly stochastic mixing matrix.
\end{lemma} 
 \begin{proof}
(The proof uses techniques similar to that of \cite[Lemma 17]{koloskova_decentralized_2019-1}).\\
 	Using the definition of $\bX^{I_{(t+1)}}$ from matrix notation in Section \ref{mat_not_sec}, we have:
 	\begin{align}
 	\Vert \bX^{I_{(t+1)}} - \Bar{\bX}^{I_{(t+1)}} \Vert_F^2 & = \Vert \bX^{I_{(t+\frac{1}{2})}} - \Bar{\bX}^{I_{(t+1)}} + \gamma \hat{\bX}^{I_{(t+1)}} (\mathbf{W}-\mathbf{I}) \Vert_F^2 \notag 
 	\intertext{Noting that $\Bar{\bX}^{I_{(t+1)}} = \Bar{\bX}^{I_{(t+\frac{1}{2})}} $ from (\ref{mean_seq_iter}) and $\Bar{\bX}^{I_{(t+\frac{1}{2})}} (\mathbf{W}-\mathbf{I})= 0$ from (\ref{mean_prop}), we get: }
 	\Vert \bX^{I_{(t+1)}} - \Bar{\bX}^{I_{(t+1)}} \Vert_F^2 & = 
 	 \Vert (\bX^{I_{(t+\frac{1}{2})}} - \Bar{\bX}^{I_{(t+\frac{1}{2})}})((1-\gamma)\mathbf{I} + \gamma \mathbf{W})+ \gamma (\hat{\bX}^{I_{(t+1)}}-\bX^{I_{(t+\frac{1}{2})}}) (\mathbf{W}-\mathbf{I}) \Vert_F^2 \notag
 	\intertext{Using the fact $\Vert \mathbf{A} + \mathbf{B} \Vert_F^2 \leq (1+\alpha_1)\Vert \mathbf{A}\Vert_F^2 + (1+\alpha_1^{-1})\Vert \mathbf{B} \Vert_F^2$ for any $\alpha_1 >0$, }
 	\Vert \bX^{I_{(t+1)}} - \Bar{\bX}^{I_{(t+1)}} \Vert_F^2 & \leq (1+\alpha_1)\Vert (\bX^{I_{(t+\frac{1}{2})}} - \Bar{\bX}^{I_{(t+\frac{1}{2})}})((1-\gamma)\mathbf{I} + \gamma \mathbf{W})\Vert_F^2 \notag \\
 	& \qquad + (1+\alpha_1^{-1}) \Vert \gamma (\hat{\bX}^{I_{(t+1)}}-\bX^{I_{(t+\frac{1}{2})}}) (\mathbf{W}-\mathbf{I}) \Vert_F^2 \notag \\
 	\intertext{Using $\Vert \mathbf{A} \mathbf{B} \Vert_F \leq \Vert \mathbf{A} \Vert_F \Vert \mathbf{B} \Vert_2 $ as per (\ref{bound_frob_mult}), we have: }
 	\Vert \bX^{I_{(t+1)}} - \Bar{\bX}^{I_{(t+1)}} \Vert_F^2 &  \leq (1+\alpha_1)\Vert (\bX^{I_{(t+\frac{1}{2})}} - \Bar{\bX}^{I_{(t+\frac{1}{2})}})((1-\gamma)\mathbf{I} + \gamma \mathbf{W})\Vert_F^2 \notag \\
 	& \qquad + (1+\alpha_1^{-1}) \gamma^2 \Vert  (\hat{\bX}^{I_{(t+1)}}-\bX^{I_{(t+\frac{1}{2})}})\Vert_F^2 .\Vert(\mathbf{W}-\mathbf{I})\Vert_2^2 \label{suppl_cvx_li_temp_eqn}
 	\end{align}
 	To bound the first term in (\ref{suppl_cvx_li_temp_eqn}), we use the triangle inequality for Frobenius norm, giving us:
 	\begin{align*}
 	\Vert (\bX^{I_{(t+\frac{1}{2})}} - \Bar{\bX}^{I_{(t+\frac{1}{2})}})((1-\gamma)\mathbf{I} + \gamma \mathbf{W})\Vert_F & \leq (1-\gamma)\Vert \bX^{I_{(t+\frac{1}{2})}} - \Bar{\bX}^{I_{(t+\frac{1}{2})}} \Vert_F \\
 	& \qquad + \gamma \Vert (\bX^{I_{(t+\frac{1}{2})}} - \Bar{\bX}^{I_{(t+\frac{1}{2})}} )W\Vert_F
 	\end{align*}
 	{From (\ref{mean_prop}), using $\Bar{\bX}^{I_{(t+\frac{1}{2})}} = \bX^{I_{(t+\frac{1}{2})}}\frac{\mathbbm{1}\mathbbm{1}^T}{n} $ and noting that $ \Bar{\bX}^{I_{(t+\frac{1}{2})}} \frac{\mathbbm{1}\mathbbm{1}^T}{n} = \Bar{\bX}^{I_{(t+\frac{1}{2})}} $, we get: }
 	\begin{align*}
 	\Vert (\bX^{I_{(t+\frac{1}{2})}} - \Bar{\bX}^{I_{(t+\frac{1}{2})}})((1-\gamma)\mathbf{I} + \gamma \mathbf{W})\Vert_F & \leq (1-\gamma)\Vert \bX^{I_{(t+\frac{1}{2})}} - \Bar{\bX}^{I_{(t+\frac{1}{2})}} \Vert_F \\
 	& \qquad + \gamma \left\Vert (\bX^{I_{(t+\frac{1}{2})}} - \Bar{\bX}^{I_{(t+\frac{1}{2})}})\left (\mathbf{W} - \frac{\mathbbm{1}\mathbbm{1}^T}{n} \right) \right\Vert_F 
 	\end{align*}
 	{Using $\Vert \mathbf{A} \mathbf{B} \Vert_F \leq \Vert \mathbf{A} \Vert_F \Vert \mathbf{B} \Vert_2 $ as per (\ref{bound_frob_mult}) and using (\ref{bound_W_mat}) for $k=0$, we can simplify the above to: }
 	\begin{align*}
 	\Vert (\bX^{I_{(t+\frac{1}{2})}} - \Bar{\bX}^{I_{(t+\frac{1}{2})}})((1-\gamma)\mathbf{I} + \gamma \mathbf{W})\Vert_F \leq (1-\gamma \delta) \Vert \bX^{I_{(t+\frac{1}{2})}} - \Bar{\bX}^{I_{(t+\frac{1}{2})}} \Vert_F
 	\end{align*}
 	Substituting the above in (\ref{suppl_cvx_li_temp_eqn}) and using $\beta = \text{max}_i \{ 1- \lambda_i(\mathbf{W}) \} \Rightarrow \Vert \mathbf{W}-\mathbf{I} \Vert_2^2 \leq \beta^2 $, we get: 
 	\begin{align*}
 	\Vert \bX^{I_{(t+1)}} - \Bar{\bX}^{I_{(t+1)}} \Vert_F^2 \leq (1+\alpha_1)(1-\gamma \delta)^2  \Vert \bX^{I_{(t+\frac{1}{2})}} - \Bar{\bX}^{I_{(t+\frac{1}{2})}} \Vert_F^2 + (1+\alpha_1^{-1}) \gamma^2 \beta^2 \Vert  {\bX}^{I_{(t+\frac{1}{2})}}-\hat{\bX}^{I_{(t+1)}}\Vert_F^2
 	\end{align*}
 	Taking expectation w.r.t the entire process, we have:
 	\begin{align*}
 	\mathbb{E} \Vert \bX^{I_{(t+1)}} - \Bar{\bX}^{I_{(t+1)}} \Vert_F^2 & \leq (1+\alpha_1)(1-\gamma \delta)^2 \mathbb{E} \Vert \bX^{I_{(t+\frac{1}{2})}} - \Bar{\bX}^{I_{(t+\frac{1}{2})}} \Vert_F^2 \notag \\
 	& \qquad + (1+\alpha_1^{-1}) \gamma^2 \beta^2 \mathbb{E} \Vert  {\bX}^{I_{(t+\frac{1}{2})}}-\hat{\bX}^{I_{(t+1)}}\Vert_F^2
 	\end{align*}
 \end{proof}

\begin{lemma}\label{lem_cvx_e2}
		Consider the sequence of updates in Algorithm \ref{alg_dec_sgd_li} in matrix form (refer \ref{mat_not_li}) with the threshold sequence $ \{  c_t \}_{t=0}^{T-1} $ . The expected deviation between the local node parameters $\bX^{I_{(t+1)}}$ and their copies $\hat{\bX}^{I_{(t+2)}}$ evaluate at timestep $I_{(t+1)}$ satisfy:
		\begin{align*}
			e^{(2)}_{I_{(t+1)}} =	\mathbb{E} \Vert \bX^{I_{(t+1)}} - \hat{\bX}^{I_{(t+2)}}\Vert_F^2	& \leq r_1 \mathbb{E} \Vert \bX^{I_{(t+\frac{1}{2})}} - \hat{\bX}^{I_{(t+1)}}\Vert_F^2 
			 + r_2 \mathbb{E} \Vert \bX^{I_{(t+\frac{1}{2})}}- \Bar{\bX}^{I_{(t+\frac{1}{2})}} \Vert_F^2 \\
			 & \qquad  +  r_{I_{(t+1)}} \eta_{I_{(t+1)}}^2
		\end{align*}
		for $r_1 = (1+\gamma \beta )^2 (1+\alpha_4) (1+\alpha_3) (1+\alpha_2) (1-\omega)$, $r_2 = \gamma^2 \beta^2 (1+\alpha_4^{-1}) (1+\alpha_3) (1+\alpha_2) (1-\omega) $ and $r_t = (1+\alpha_3^{-1}) (1+\alpha_2) (1-\omega)  nH^2G^2 + (1+\alpha_2)\omega n c_t + (1+\alpha_2^{-1}) nH^2G^2 $ with $r_{I_{(t+1)}}$ denoting its evaluation at timestep ${I_{(t+1)}}$.
		Here $\alpha_2, \alpha_3, \alpha_4$ are positive constants, $\omega$ is the compression coefficient for operator \C, $\gamma$ is the consensus stepsize, $\beta = \verts{\mathbf{W} - \mathbf{I}}_2$ with $\mathbf{W}$ being the doubly stochastic mixing matrix and $H$ denotes the synchronization period.
\end{lemma} 
\begin{proof}
	Using definition of $\hat{\bX}^{I_{(t+2)}}$ from matrix notation in Section \ref{mat_not_sec} and considering expectation w.r.t entire process, we have:
	\begin{align*}
	& \mathbb{E} \Vert \bX^{I_{(t+1)}} - \hat{\bX}^{I_{(t+2)}}\Vert_F^2  = \mathbb{E} \Vert \bX^{I_{(t+1)}} - \hat{\bX}^{I_{(t+1)}} - \C ( (\bX^{I_{(t+\frac{3}{2})}} - \hat{\bX}^{I_{(t+1)}}) \mathbf{P}^{(I_{(t+2)}-1)} )\Vert_F^2 \\
	&  \hspace{2.4cm} = \mathbb{E} \Vert \bX^{I_{(t+\frac{3}{2})}} - \hat{\bX}^{I_{(t+1)}}+ \bX^{I_{(t+1)}} - \bX^{I_{(t+\frac{3}{2})}} - \C ( (\bX^{I_{(t+\frac{3}{2})}} - \hat{\bX}^{I_{(t+1)}}) \mathbf{P}^{(I_{(t+2)}-1)} )\Vert_F^2 
	\end{align*}
	Using $\Vert \mathbf{A} + \mathbf{B} \Vert_F^2 \leq (1+\alpha_2)\Vert \mathbf{A}\Vert_F^2 + (1+\alpha_2^{-1})\Vert \mathbf{B} \Vert_F^2$ for any $\alpha_2 >0$,
	\begin{align}
	\mathbb{E} \Vert \bX^{I_{(t+1)}} - \hat{\bX}^{I_{(t+2)}}\Vert_F^2 & \leq (1+\alpha_2) \mathbb{E} \Vert \bX^{I_{(t+\frac{3}{2})}} - \hat{\bX}^{I_{(t+1)}} - \C ( (\bX^{I_{(t+\frac{3}{2})}}- \hat{\bX}^{I_{(t+1)}}) \mathbf{P}^{(I_{(t+2)}-1)} )\Vert_F^2 \notag \\
	&\hspace{2cm} + (1+\alpha_2^{-1})\Vert \bX^{I_{(t+1)}} - \bX^{I_{(t+\frac{3}{2})}} \Vert_F^2 \notag \\
	& = (1+\alpha_2) \mathbb{E} \Vert \bX^{I_{(t+\frac{3}{2})}} - \hat{\bX}^{I_{(t+1)}} - \C ( (\bX^{I_{(t+\frac{3}{2})}} - \hat{\bX}^{I_{(t+1)}}) \mathbf{P}^{(I_{(t+2)}-1)} )\Vert_F^2 \notag \\
	& \hspace{2cm} + (1+\alpha_2^{-1})\left\Vert \sum_{t' = I_{(t+1)}}^{I_{(t+2)}-1}\eta_{t'} \partial F (\bX^{t'}, \xi^{t'}) \right\Vert_F^2 \label{cvx_lem1_interim1}
	\end{align}
	Bounding the last term in \eqref{cvx_lem1_interim1} using (\ref{bound_gap_grad}), we get:
	\begin{align*}
	\mathbb{E} \Vert \bX^{I_{(t+1)}} - \hat{\bX}^{I_{(t+2)}}\Vert_F^2 & \leq (1+\alpha_2) \mathbb{E} \Vert \bX^{I_{(t+\frac{3}{2})}} - \hat{\bX}^{I_{(t+1)}} - \C ( (\bX^{I_{(t+\frac{3}{2})}} - \hat{\bX}^{I_{(t+1)}}) \mathbf{P}^{(I_{(t+2)}-1)} )\Vert_F^2 \\
	& \qquad + (1+\alpha_2^{-1}) \eta_{I_{(t+1)}}^2nH^2G^2 
	\end{align*}
	Noting that the entries of $\mathbf{P}^{(I_{(t+2)}-1)} $ and $\mathbf{I} - \mathbf{P}^{(I_{(t+2)}-1)} $ are disjoint, we can separate them in the squared Frobenius norm:
	\begin{align*}
	\mathbb{E} \Vert \bX^{I_{(t+1)}} - \hat{\bX}^{I_{(t+2)}}\Vert_F^2  & \leq  (1+\alpha_2)  \mathbb{E} \Vert (\bX^{I_{(t+\frac{3}{2})}} - \hat{\bX}^{I_{(t+1)}})\mathbf{P}^{(I_{(t+2)}-1)} - \C ( (\bX^{I_{(t+\frac{3}{2})}} - \hat{\bX}^{I_{(t+1)}}) \mathbf{P}^{(I_{(t+2)}-1)} )\Vert_F^2  \\
	&  \hspace{0.5cm} + (1+\alpha_2) \Vert (\bX^{I_{(t+\frac{3}{2})}} - \hat{\bX}^{I_{(t+1)}})(\mathbf{I} - \mathbf{P}^{(I_{(t+2)}-1)})\Vert_F^2  + (1+\alpha_2^{-1}) \eta_{I_{(t+1)}}^2nH^2G^2 
	\end{align*}
	Using the compression property of operator $\C$ as per (\ref{bound_comp}), we have:
	\begin{align*}
	& \mathbb{E} \Vert \bX^{I_{(t+1)}} - \hat{\bX}^{I_{(t+2)}}\Vert_F^2 \leq (1+\alpha_2) (1-\omega) \mathbb{E}  \Vert (\bX^{I_{(t+\frac{3}{2})}} - \hat{\bX}^{I_{(t+1)}})\mathbf{P}^{(I_{(t+2)}-1)} \Vert_F^2 \\
	& \hspace{2.5cm} +  (1+\alpha_2) \mathbb{E} \Vert (\bX^{I_{(t+\frac{3}{2})}} - \hat{\bX}^{I_{(t+1)}})(\mathbf{I} - \mathbf{P}^{(I_{(t+2)}-1)})\Vert_F^2  + (1+\alpha_2^{-1}) \eta_{I_{(t+1)}}^2nH^2G^2 
	\end{align*}
	Adding and subtracting $ (1+\alpha_2)(1-\omega)  \mathbb{E}\Vert (\bX^{I_{(t+\frac{3}{2})}} - \hat{\bX}^{I_{(t+1)}})(\mathbf{I} - \mathbf{P}^{(I_{(t+2)}-1)})\Vert_F^2$, we get:
	\begin{align*}
	\mathbb{E} \Vert \bX^{I_{(t+1)}} - \hat{\bX}^{I_{(t+2)}}\Vert_F^2 & \leq (1+\alpha_2) (1-\omega)  \mathbb{E} \Vert \bX^{I_{(t+\frac{3}{2})}} - \hat{\bX}^{I_{(t+1)}} \Vert_F^2 + (1+\alpha_2^{-1}) \eta_{I_{(t+1)}}^2nH^2G^2  \\
	& \qquad +(1+\alpha_2) \omega  \mathbb{E} \Vert (\bX^{I_{(t+\frac{3}{2})}} - \hat{\bX}^{I_{(t+1)}})(\mathbf{I} - \mathbf{P}^{(I_{(t+2)}-1)})\Vert_F^2 
	\end{align*}
	The third term in the RHS above denotes the norm of nodes which did not communicate and thus should be bounded by the triggering condition using (\ref{bound_trig}),
	\begin{align}
	\mathbb{E}& \Vert \bX^{I_{(t+1)}} - \hat{\bX}^{I_{(t+2)}}\Vert_F^2 \notag \\
	& \leq (1+\alpha_2) (1-\omega) \mathbb{E} \Vert \bX^{I_{(t+\frac{3}{2})}} - \hat{\bX}^{I_{(t+1)}} \Vert_F^2 +(1+\alpha_2)\omega n c_{I_{(t+1)}} \eta_{I_{(t+1)}}^2  + (1+\alpha_2^{-1}) \eta_{I_{(t+1)}}^2nH^2G^2 \notag \\
	& = (1+\alpha_2) (1-\omega) \mathbb{E} \left\Vert \bX^{I_{(t+1)}}  - \sum_{t' = I_{(t+1)}}^{I_{(t+2)}-1}\eta_{t'} \partial F (\bX^{t'}, \xi^{t'}) -  \hat{\bX}^{I_{(t+1)}} \right\Vert_F^2 + (1+\alpha_2)\omega n c_{I_{(t+1)}} \eta_{I_{(t+1)}}^2 \notag \\
	& \hspace{2cm} + (1+\alpha_2^{-1}) \eta_{I_{(t+1)}}^2nH^2G^2  \notag \\
	& \leq (1+\alpha_3) (1+\alpha_2) (1-\omega) \mathbb{E} \Vert \bX^{I_{(t+1)}}  -  \hat{\bX}^{I_{(t+1)}} \Vert_F^2 + (1+\alpha_2)\omega n c_{I_{(t+1)}} \eta_{I_{(t+1)}}^2 \notag \\
	& \hspace{1cm} + (1+\alpha_2^{-1}) \eta_{I_{(t+1)}}^2nH^2G^2 + (1+\alpha_3^{-1}) (1+\alpha_2) (1-\omega) \mathbb{E} \left\Vert \sum_{t' = I_{(t+1)}}^{I_{(t+2)}-1}\eta_{t'} \partial F (\bX^{t'}, \xi^{t'}) \right\Vert_F^2 \label{cvx_lemm_interim2}
	\end{align}
	where in the last inequality, we have used (\ref{bound_l2_sum}) for some constant $\alpha_3 > 0$. Using (\ref{bound_gap_grad}) to bound the last term in \eqref{cvx_lemm_interim2} , we have:
	\begin{align*}
	\mathbb{E} & \Vert \bX^{I_{(t+1)}} - \hat{\bX}^{I_{(t+2)}}\Vert_F^2 \\
	& \leq (1+\alpha_3) (1+\alpha_2) (1-\omega) \mathbb{E} \Vert \bX^{I_{(t+1)}}  -  \hat{\bX}^{I_{(t+1)}} \Vert_F^2 +  (1+\alpha_3^{-1}) (1+\alpha_2) (1-\omega) \eta_{I_{(t+1)}}^2 nH^2G^2 \\
	& \hspace{1cm}+ (1+\alpha_2)\omega n c_{I_{(t+1)}} \eta_{I_{(t+1)}}^2 + (1+\alpha_2^{-1}) \eta_{I_{(t+1)}}^2nH^2G^2 \\
	& = (1+\alpha_3) (1+\alpha_2) (1-\omega) \mathbb{E} \Vert \bX^{I_{(t+\frac{1}{2})}} + \gamma \hat{\bX}^{I_{(t+1)}} (\mathbf{W}-\mathbf{I}) -  \hat{\bX}^{I_{(t+1)}} \Vert_F^2 \\
	& \hspace{1cm} +  (1+\alpha_3^{-1}) (1+\alpha_2) (1-\omega) \eta_{I_{(t+1)}}^2 nH^2G^2  + (1+\alpha_2)\omega n c_{I_{(t+1)}}\eta_{I_{(t+1)}}^2 + (1+\alpha_2^{-1}) \eta_{I_{(t+1)}}^2nH^2G^2 \\
	& \leq (1+\alpha_3) (1+\alpha_2) (1-\omega)\mathbb{E} \Vert (\bX^{I_{(t+\frac{1}{2})}} - \hat{\bX}^{I_{(t+1)}})((1+\gamma)\mathbf{I} - \gamma \mathbf{W})+ \gamma (\bX^{I_{(t+\frac{1}{2})}}- \Bar{\bX}^{I_{(t+\frac{1}{2})}})(\mathbf{W}-\mathbf{I})\Vert_F^2 \\
	& \hspace{1cm} +  (1+\alpha_3^{-1}) (1+\alpha_2) (1-\omega) \eta_{I_{(t+1)}}^2 nH^2G^2 + (1+\alpha_2)\omega n c_{I_{(t+1)}} \eta_{I_{(t+1)}}^2 + (1+\alpha_2^{-1}) \eta_{I_{(t+1)}}^2nH^2G^2
	\end{align*}
	where in the last inequality we've used $ \Bar{\bX}^{I_{(t+\frac{1}{2})}} (\mathbf{W}-\mathbf{I})= 0 $. For $\alpha_4 > 0 $, using (\ref{bound_l2_sum}) gives us:
	\begin{align*}
	\mathbb{E} \Vert \bX^{I_{(t+1)}} - \hat{\bX}^{I_{(t+2)}}\Vert_F^2	& \leq (1+\alpha_4) (1+\alpha_3) (1+\alpha_2) (1-\omega) \mathbb{E} \Vert (\bX^{I_{(t+\frac{1}{2})}} - \hat{\bX}^{I_{(t+1)}})((1+\gamma)\mathbf{I} - \gamma \mathbf{W})\Vert_F^2 \\
	& \hspace{0.5cm} + (1+\alpha_4^{-1}) (1+\alpha_3) (1+\alpha_2) (1-\omega) \mathbb{E} \Vert \gamma (\bX^{I_{(t+\frac{1}{2})}}- \Bar{\bX}^{I_{(t+\frac{1}{2})}})(\mathbf{W}-\mathbf{I})\Vert_F^2 \\
	& \hspace{1.0cm} +  (1+\alpha_3^{-1}) (1+\alpha_2) (1-\omega) \eta_{I_{(t+1)}}^2 nH^2G^2 + (1+\alpha_2)\omega n c_{I_{(t+1)}} \eta_{I_{(t+1)}}^2 \\
	& \hspace{1.5cm} + (1+\alpha_2^{-1}) \eta_{I_{(t+1)}}^2nH^2G^2 
	\end{align*}
	Using $\Vert (1+\gamma)\mathbf{I} - \gamma W \Vert_2 = \Vert I + \gamma (\mathbf{I} - \mathbf{W})\Vert_2 = 1 +  \gamma \Vert \mathbf{I} - \mathbf{W} \Vert_2  = 1 + \gamma \beta $ (by definition of $\beta = \text{max}_i \{ 1- \lambda_i(\mathbf{W}) \}$) and $\verts{\mathbf{I} - \mathbf{W}}_2 = \beta$  along with $\verts{\mathbf{AB}}_F \leq \verts{\mathbf{A}}_F\verts{\mathbf{B}}_2$ from (\ref{bound_frob_mult}):
	\begin{align*}
	\mathbb{E} \Vert \bX^{I_{(t+1)}} - \hat{\bX}^{I_{(t+2)}}\Vert_F^2	& \leq (1+\gamma \beta )^2 (1+\alpha_4) (1+\alpha_3) (1+\alpha_2) (1-\omega)\mathbb{E} \Vert \bX^{I_{(t+\frac{1}{2})}} - \hat{\bX}^{I_{(t+1)}}\Vert_F^2 \notag \\
	& \hspace{0.5cm} + \gamma^2 \beta^2 (1+\alpha_4^{-1}) (1+\alpha_3) (1+\alpha_2) (1-\omega) \mathbb{E} \Vert \bX^{I_{(t+\frac{1}{2})}}- \Bar{\bX}^{I_{(t+\frac{1}{2})}} \Vert_F^2 \notag \\
	& \hspace{1cm} +  (1+\alpha_3^{-1}) (1+\alpha_2) (1-\omega) \eta_{I_{(t+1)}}^2 nH^2G^2 + (1+\alpha_2)\omega n c_{I_{(t+1)}} \eta_{I_{(t+1)}}^2 \\
	& \hspace{1.5cm} + (1+\alpha_2^{-1}) \eta_{I_{(t+1)}}^2nH^2G^2
	\end{align*}
\end{proof}

\begin{proof} [Proof of Lemma \ref{lem_dec_li_sgd}]
	We now proceed to the main proof of the lemma. Consider the following expression :
	\begin{align} \label{suppl_et_eqn_li}
	e_{I_{(t+1)}} := \underbrace{\mathbb{E} \Vert \bX^{I_{(t+1)}} - \Bar{\bX}^{I_{(t+1)}} \Vert_F^2}_{e^{(1)}_{I_{(t+1)}}} + \underbrace{ \mathbb{E} \Vert \bX^{I_{(t+1)}} - \hat{\bX}^{I_{(t+2)}} \Vert_F^2}_{e^{(2)}_{I_{(t+1)}}}
	\end{align}
	We note that Lemma \ref{lem_cvx_e1} and Lemma \ref{lem_cvx_e2} provide bounds for the first and second term of in RHS of (\ref{suppl_et_eqn_li}). Substituting them in (\ref{suppl_et_eqn_li}) gives us:
	\begin{align} \label{suppl_et_eqn_total_li}
	e_{I_{(t+1)}} & = \mathbb{E} \Vert \bX^{I_{(t+1)}} - \Bar{\bX}^{I_{(t+1)}} \Vert_F^2 + \mathbb{E} \Vert \bX^{I_{(t+1)}} - \hat{\bX}^{I_{(t+2)}} \Vert_F^2 \notag \\ 
	& \leq (1+\alpha_1)(1-\gamma \delta)^2 \mathbb{E}\Vert \bX^{I_{(t+\frac{1}{2})}} - \Bar{\bX}^{I_{(t+\frac{1}{2})}} \Vert_F^2 + (1+\alpha_1^{-1}) \gamma^2 \beta^2 \mathbb{E} \Vert  \bX^{I_{(t+\frac{1}{2})}}-\hat{\bX}^{I_{(t+1)}}\Vert_F^2 \notag \\ 
	& \hspace{0.5cm} + (1+\gamma \beta )^2 (1+\alpha_4) (1+\alpha_3) (1+\alpha_2) (1-\omega) \mathbb{E} \Vert \bX^{I_{(t+\frac{1}{2})}} - \hat{\bX}^{I_{(t+1)}}\Vert_F^2 \notag \\
	& \hspace{1.0cm} + \gamma^2 \beta^2 (1+\alpha_4^{-1}) (1+\alpha_3) (1+\alpha_2) (1-\omega) \mathbb{E} \Vert \bX^{I_{(t+\frac{1}{2})}}- \Bar{\bX}^{I_{(t+\frac{1}{2})}} \Vert_F^2 \notag \\ 
	&  \hspace{1.5cm} +  (1+\alpha_3^{-1}) (1+\alpha_2) (1-\omega) \eta_{I_{(t+1)}}^2 nH^2G^2+ (1+\alpha_2)\omega n c_{I_{(t+1)}} \eta_{I_{(t+1)}}^2 \notag \\
	&  \hspace{2.0cm} + (1+\alpha_2^{-1}) \eta_{I_{(t+1)}}^2nH^2G^2
	\end{align}
	Define the following:
	\begin{align*}
	\pi_1 (\gamma) & := \gamma^2 \beta^2 (1+\alpha_1^{-1}) + (1+\gamma \beta )^2 (1+\alpha_4) (1+\alpha_3) (1+\alpha_2) (1-\omega) \\
	\pi_2 (\gamma) & := (1-\delta \gamma)^2(1+\alpha_1) + \gamma^2 \beta^2 (1+\alpha_4^{-1}) (1+\alpha_3) (1+\alpha_2) (1-\omega) \\
	\pi_t & := (1+\alpha_3^{-1}) (1+\alpha_2) (1-\omega) nH^2G^2 + (1+\alpha_2)\omega n c_t + (1+\alpha_2^{-1})nH^2G^2 
	\end{align*}
	The bound on $e_{I_{(t+1)}}$ in (\ref{suppl_et_eqn_total_li}) can be rewritten as:
	\begin{align*}
	e_{I_{(t+1)}} & \leq \pi_1 (\gamma) \mathbb{E} \Vert \bX^{I_{(t+\frac{1}{2})}} - \hat{\bX}^{I_{(t+1)}} \Vert_F^2 + \pi_2 (\gamma) \mathbb{E} \Vert \bX^{I_{(t+\frac{1}{2})}} - \Bar{\bX}^{I_{(t+\frac{1}{2})}} \Vert_F^2 + \pi_{I_{(t+1)}} \eta_{I_{(t+1)}}^2 \\
	& \leq max \{ \pi_1 (\gamma) , \pi_2 (\gamma) \} \, \mathbb{E} \left[ \Vert {\bX}^{I_{(t+\frac{1}{2})}} - \hat{\bX}^{I_{(t+1)}} \Vert_F^2 + \Vert \bX^{I_{(t+\frac{1}{2})}} - \Bar{\bX}^{I_{(t+\frac{1}{2})}} \Vert_F^2 \right] + \pi_{I_{(t+1)}}  \eta_{I_{(t+1)}}^2
	\end{align*}
	Calculation of $ max \{ \pi_1 (\gamma) , \pi_2 (\gamma) \}$ and $\pi_t$ is given in Lemma \ref{suppl_lem_coeff_calc}, where we show that:\\
	 $ max \{ \pi_1 (\gamma) , \pi_2 (\gamma) \} \leq   \left(  1 - \frac{\gamma^* \delta}{8}  \right)   \leq \left( 1 - \frac{\delta^2 \omega}{644} \right)  $ and $\pi_t \leq \left( \frac{8nG^2H^2}{\omega} + \frac{5\omega n c_t}{4} \right) $. This yields: 
	\begin{align*}
	e_{I_{(t+1)}} \leq \left( 1 - \frac{\delta \gamma^*}{8} \right)\mathbb{E}  \left[ \Vert \bX^{I_{(t+\frac{1}{2})}} - \hat{\bX}^{I_{(t+1)}} \Vert_F^2 + \Vert \bX^{I_{(t+\frac{1}{2})}} - \Bar{\bX}^{I_{(t+\frac{1}{2})}} \Vert_F^2 \right] + \left( \frac{8nG^2H^2}{\omega} + \frac{5\omega n c_{I_{(t)}}}{4} \right) \eta_{I_{(t)}}^2
	\end{align*}
	where $\gamma^* = \frac{2 \delta \omega}{64 \delta + \delta^2 + 16 \beta^2 + 8 \delta \beta^2 - 16\delta \omega}$ from Lemma \ref{suppl_lem_coeff_calc} and we've used the fact that  $\pi_{I_{(t)}} \eta_{I_{(t)}}^2 \geq  \pi_{I_{(t+1)}} \eta_{I_{(t+1)}}^2 $ which holds because $c_t \sim o(t)$. \\
	From definition of $e_{I_{(t+1)}} = \mathbb{E} \left[ \Vert \bX^{I_{(t+1)}} - \Bar{\bX}^{I_{(t+1)}} \Vert_F^2 + \Vert \bX^{I_{(t+1)}} - \hat{\bX}^{I_{(t+2)}} \Vert_F^2 \right] $ and defining $p = \frac{\gamma^* \delta }{8}$ and $z_t :=\left( \frac{8nG^2H^2}{\omega} + \frac{5\omega n c_t}{4} \right) $, we have: 
	\begin{align*}
	\mathbb{E}& \left[ \Vert \bX^{I_{(t+1)}} - \Bar{\bX}^{I_{(t+1)}} \Vert_F^2 + \Vert \bX^{I_{(t+1)}} - \hat{\bX}^{I_{(t+2)}} \Vert_F^2 \right] \\
	& \hspace{2cm} \leq (1-p) \mathbb{E} \left[\Vert \bX^{I_{(t+\frac{1}{2})}} - \Bar{\bX}^{I_{(t+\frac{1}{2})}} \Vert_F^2 + \mathbb{E}\Vert \bX^{I_{(t+\frac{1}{2})}} - \hat{\bX}^{I_{(t+1)}} \Vert_F^2\right]  + z_{I_{(t)}} \eta_{I_{(t)}}^2
	\end{align*}
	Noting the fact that $\Bar{\bX}^{I_{(t+\frac{1}{2})}} = \Bar{\bX}^{I_{(t)}} - \sum_{t' = I_{(t)}}^{(I_{(t+1)}-1)} \eta_{t'}\partial F(\bX^{(t')}, \boldsymbol{\xi}^{(t')} )   \frac{\mathbbm{1}\mathbbm{1}^T}{n}  $: 
	\begin{align*}
	\mathbb{E} & \left[ \Vert \bX^{I_{(t+1)}} - \Bar{\bX}^{I_{(t+1)}} \Vert_F^2 + \Vert \bX^{I_{(t+1)}} - \hat{\bX}^{I_{(t+2)}} \Vert_F^2 \right]\\ & \hspace{2cm} \leq (1-p)\mathbb{E}\left\Vert \Bar{\bX}^{I_{(t)}} - \bX^{I_{(t)}} -  \sum_{t' = I_{(t)}}^{I_{(t+1)}-1} \eta_{t'}\partial F(\bX^{(t')}, \xi^{(t')} )\left(   \frac{\mathbbm{1}\mathbbm{1}^T}{n} - I \right) \right\Vert_F^2  \\
	&\hspace{2.5cm} + (1-p)\mathbb{E}\left\Vert \hat{\bX}^{I_{(t+1)}} - \bX^{I_{(t)}} + \sum_{t' = I_{(t)}}^{I_{(t+1)}-1} \eta_{t'}\partial F(\bX^{(t')}, \xi^{(t')} ) \right\Vert_F^2 + z_{I_{(t)}} \eta_{I_{(t)}}^2
	\end{align*}
	Using $\Vert \mathbf{A} + \mathbf{B} \Vert_F^2 \leq (1+\alpha_5)\Vert \mathbf{A}\Vert_F^2 + (1+\alpha_5^{-1})\Vert \mathbf{B} \Vert_F^2$ for any $\alpha_5 >0$,
	\begin{align}
	\mathbb{E} & \left[ \Vert \bX^{I_{(t+1)}} - \Bar{\bX}^{I_{(t+1)}} \Vert_F^2 + \Vert \bX^{I_{(t+1)}} - \hat{\bX}^{I_{(t+2)}} \Vert_F^2 \right] \notag \\
	& \hspace{2.0cm} \leq (1-p)(1+\alpha_5^{-1})\mathbb{E}\left[ \Vert\Bar{\bX}^{I_{(t)}} - \bX^{I_{(t)}}\Vert_F^2  + \Vert \hat{\bX}^{I_{(t+1)}} - \bX^{I_{(t)}} \Vert_F^2 \right] \notag \\
	& \hspace{2.5cm} + (1-p)(1+\alpha_5) \mathbb{E}\left\Vert \sum_{t' = I_{(t)}}^{I_{(t+1)}-1} \eta_{t'}\partial F(\bX^{(t')}, \xi^{(t')} )\left(   \frac{\mathbbm{1}\mathbbm{1}^T}{n} - I \right) \right\Vert_F^2 \notag  \\ 
	& \hspace{3cm}  + (1-p)(1+\alpha_5) \mathbb{E}\left\Vert \sum_{t' = I_{(t)}}^{I_{(t+1)}-1} \eta_{t'}\partial F(\bX^{(t')}, \xi^{(t')} )\right\Vert_F^2 + z_{I_{(t)}}  \eta_{I_{(t)}}^2 \label{cvx_lemm_interim3}
	\end{align}
	Using (\ref{bound_frob_mult}) to bound the second term in \eqref{cvx_lemm_interim3} and noting that $\mathbb{E} \left\Vert \sum_{t' = I_{(t)}}^{I_{(t+1)}-1} \eta_{t'}\partial F(\bX^{(t')}, \xi^{(t')} )\right\Vert_F^2  \leq \eta_{I_{(t)}} nH^2G^2 $ from (\ref{bound_gap_grad}) and $\Vert \frac{\mathbbm{1}\mathbbm{1}^T}{n} - I \Vert_2^2 = 1 $  from (\ref{bound_W_mat}) (with $k=0$) respectively, we get: 
	\begin{align*}
	\mathbb{E} & \left[ \Vert \bX^{I_{(t+1)}} - \Bar{\bX}^{I_{(t+1)}} \Vert_F^2 + \Vert \bX^{I_{(t+1)}} - \hat{\bX}^{I_{(t+2)}} \Vert_F^2 \right] \\ &  \hspace{2cm} \leq (1-p) (1+\alpha_5^{-1})\mathbb{E}\left[ \Vert\Bar{\bX}^{I_{(t)}} - \bX^{I_{(t)}}\Vert_F^2  + \Vert \hat{\bX}^{I_{(t+1)}} - \bX^{I_{(t)}} \Vert_F^2 \right]  \\
	& \hspace{3cm} + 2(1-p)(1+\alpha_5) H^2G^2n\eta_{I_{(t)}}^2+ z_{I_{(t)}}  \eta_{I_{(t)}}^2 \\
	&  \hspace{2cm} \stackrel{(\alpha_5 = \frac{2}{p})}{\leq} \left( 1-\frac{p}{2} \right)\mathbb{E}\left[ \Vert\Bar{\bX}^{I_{(t)}} - \bX^{I_{(t)}}\Vert_F^2  + \Vert \hat{\bX}^{I_{(t+1)}} - \bX^{I_{(t)}} \Vert_F^2 \right] \\
	& \hspace{3cm} + \frac{4n}{p}\eta_{I_{(t)}}^2G^2H^2 + z_{I_{(t)}}  \eta_{I_{(t)}}^2
	\end{align*}
	Define $A_t := 2nG^2H^2 + \frac{p z_t}{2}$ (where $z_t= \left( \frac{8nH^2G^2}{\omega} + \frac{5\omega n c_t}{4} \right) $ as above), thus we have the following relation:
	\begin{align*}
	\mathbb{E}\left[\Vert \bX^{I_{(t+1)}} - \Bar{\bX}^{I_{(t+1)}} \Vert_F^2 + \Vert \bX^{I_{(t+1)}} - \hat{\bX}^{I_{(t+2)}} \Vert_F^2 \right] & \leq \left( 1-\frac{p}{2} \right)\mathbb{E}\left[ \Vert\Bar{\bX}^{I_{(t)}} - \bX^{I_{(t)}}\Vert_F^2  + \Vert \hat{\bX}^{I_{(t+1)}} - \bX^{I_{(t)}} \Vert_F^2 \right] \\
	& \qquad + \frac{2A_{I_{(t)}}}{p}\eta_{I_{(t)}}^2
	\end{align*}
	Using $e_{I_{(t)}} : = \mathbb{E} \left[ \Vert\Bar{\bX}^{I_{(t)}} - \bX^{I_{(t)}}\Vert_F^2  + \Vert \hat{\bX}^{I_{(t+1)}} - \bX^{I_{(t)}} \Vert_F^2 \right] $, above can be written as:
	\begin{align} \label{suppl_lemm_rec_rel_li}
	e_{I_{(t+1)}} \leq \left( 1- \frac{p}{2} \right) e_{I_{(t)}} + \frac{2A_{I_{(t)}}}{p} \eta_{I_{(t)}}^2
	\end{align}
	Thus, employing Lemma \ref{suppl_lem_e_seq_li}, the sequence $e_{I_{(t)}}$ follows the bound for all $t$: 
	\begin{align*}
	e_{I_{(t)}} \leq \frac{20A_{I_{(t)}} \eta_{I_{(t)}}^2}{p^2}
	\end{align*}
	Note that we also have: $
	\mathbb{E}\Vert\Bar{\bX}^{I_{(t)}} - \bX^{I_{(t)}}\Vert_F^2 \leq \mathbb{E} \left[ \Vert\Bar{\bX}^{I_{(t)}} - \bX^{I_{(t)}}\Vert_F^2  + \Vert \hat{\bX}^{I_{(t+1)}} - \bX^{I_{(t)}} \Vert_F^2 \right] := e_{I_{(t)}}$. Thus, we get:
	\begin{align*}
	\mathbb{E}\Vert\Bar{\bX}^{I_{(t)}} - \bX^{I_{(t)}}\Vert_F^2 \leq \frac{20A_{I_{(t)}}\eta_{I_{(t)}}^2}{p^2}
	\end{align*}
	where $A_{I_{(t)}} := 2nG^2H^2 + \frac{p}{2} \left( \frac{8nH^2G^2}{\omega} + \frac{5\omega n c_{I_{(t)}}}{4}  \right) $ and $p = \frac{\delta \gamma^*}{8}$ with $\gamma^* =  \frac{2 \delta \omega}{64 \delta + \delta^2 + 16 \beta^2 + 8 \delta \beta^2 - 16\delta \omega} $ 
\end{proof}
\begin{lemma} \label{suppl_lem_coeff_calc}
	(Variant of \cite[Lemma 18]{koloskova_decentralized_2019-1})
	Consider the following variables:
	\begin{align*}
	\pi_1 (\gamma) & := \gamma^2 \beta^2 (1+\alpha_1^{-1}) + (1+\gamma \beta )^2 (1+\alpha_4) (1+\alpha_3) (1+\alpha_2) (1-\omega) \\
	\pi_2 (\gamma) & := (1-\delta \gamma)^2(1+\alpha_1) + \gamma^2 \beta^2 (1+\alpha_4^{-1}) (1+\alpha_3) (1+\alpha_2) (1-\omega) \\
	\pi_t & := (1+\alpha_3^{-1}) (1+\alpha_2) (1-\omega) nG^2H^2 + (1+\alpha_2)\omega n c_t + (1+\alpha_2^{-1})nG^2 H^2
	\end{align*}
	and the following choice of variables:
	\begin{align*}
	\alpha_1 := \frac{\gamma \delta}{2}, \, 
	\alpha_2 := \frac{\omega}{4}, \,
	\alpha_3 := \frac{\omega}{4}, \,
	\alpha_4 := \frac{\omega}{4} \\
	\gamma^* := \frac{2 \delta \omega}{64 \delta + \delta^2 + 16 \beta^2 + 8 \delta \beta^2 - 16\delta \omega} \\
	\end{align*}
	Then, it can be shown that:
	\begin{align*}
	max \{ \pi_1 (\gamma^*) , \pi_2 (\gamma^*) \}  \leq 1 - \frac{\delta^2 \omega}{644} \hspace{0.5cm}, \hspace{0.5cm}
	\pi_t \leq \frac{8nG^2H^2}{\omega} + \frac{5\omega nc_t}{4}
	\end{align*}
\end{lemma}
\begin{proof}
	Consider:
	\begin{align*}
	(1+\alpha_4) (1+\alpha_3) (1+\alpha_2) (1-\omega) & = (1+\frac{\omega}{4})^3 (1-\omega) \\
	& = \left( 1 - \frac{\omega^4}{64} - \frac{11 \omega^3}{64} - \frac{9 \omega^2}{16} - \frac{\omega}{4} \right) \\
	& \leq \left( 1 - \frac{\omega}{4} \right)
	\end{align*}
	This gives us:
	\begin{align*}
	\pi_1(\gamma) \leq \gamma^2\beta^2\left(1+\frac{2}{\gamma \delta} \right) + (1+\gamma\beta)^2 \left( 1 - \frac{\omega}{4} \right)
	\end{align*}
	Noting that $\gamma^2 \leq \gamma $ (for $\gamma \leq 1 $ which is true for $\gamma^*$ ) and $\beta \leq 2$, we have:
	\begin{align*}
	\pi_1(\gamma) \leq \beta^2\left(\gamma +\frac{2\gamma}{\delta} \right) + (1+8\gamma) \left( 1 - \frac{\omega}{4} \right)
	\end{align*}
	Substituting value of $\gamma^*$ in above, it can be shown that:
	\begin{align*}
	\pi_1(\gamma^*) \leq 1 - \frac{\delta^2\omega}{4(64 \delta + \delta^2 + 16 \beta^2 + 8 \delta \beta^2 - 16\delta \omega)}
	\end{align*}
	Now we note that:
	\begin{align*}
	\pi_2 (\gamma) & =  (1-\delta \gamma)^2 \left(1+\frac{\delta \gamma}{2} \right) + \gamma^2 \beta^2 \left( 1+\frac{4}{\omega} \right) \left(1+\frac{\omega}{4} \right)^2 (1-\omega) 
	\intertext{Noting the fact that for $x = \delta \gamma \leq 1 $ : $\left( 1 - x  \right) \left(1 + \frac{x}{2}\right) \leq \left( 1 - \frac{x}{2}  \right) $ and $ \left( 1 -x \right) \left( 1 - \frac{x}{2} \right) \leq \left( 1 - \frac{x}{2}\right)^2 $, }
	\pi_2 (\gamma) & \leq \left( 1-\frac{\gamma \delta}{2} \right)^2 + \gamma^2 \beta^2 \left( 1+\frac{4}{\omega} \right) \left(1+\frac{\omega}{4} \right)^2 (1-\omega)\\
	& = \left( 1-\frac{\gamma \delta}{2} \right)^2 + \gamma^2 \beta^2 \left(3+\frac{3\omega}{4}+\frac{\omega^2}{16} + \frac{4}{\omega} \right) (1-\omega) \\
	& \leq \left( 1-\frac{\gamma \delta}{2} \right)^2 + \gamma^2 \beta^2 \frac{4}{\omega} \, : = \zeta(\gamma)
	\end{align*}
	Note that $\zeta(\gamma)$ is convex and quadratic in $\gamma$, and attains minima at $\gamma' = \frac{2 \delta \omega}{16 \beta^2 + \delta^2 \omega}$ with value $\zeta(\gamma') = \frac{16\beta^2}{16\beta^2 + \omega \delta^2}$ \\
	By Jensen's inequality, we note that for any $\lambda \in [0,1]$
	\begin{align*}
	\zeta(\lambda \gamma') \leq (1-\lambda) \zeta(0) + \lambda \zeta(\gamma') = 1 - \lambda \frac{\delta^2 \omega}{16 \beta^2 + \delta^2 \omega}
	\end{align*}
	For the choice $\lambda' = \frac{16\beta^2 + \omega \delta^2}{64 \delta + \delta^2 + 16 \beta^2 + 8 \delta \beta^2 - 16\delta \omega}$, it can be seen that $\lambda' \gamma' = \gamma^*$. Thus we get:
	\begin{align*}
	\pi_2(\gamma^*) \leq  \zeta(\lambda' \gamma')  & \leq 1 - \frac{\delta^2\omega}{(64 \delta + \delta^2 + 16 \beta^2 + 8 \delta \beta^2 - 16\delta \omega)} \\
	& \leq 1 - \frac{\delta^2\omega}{4(64 \delta + \delta^2 + 16 \beta^2 + 8 \delta \beta^2 - 16\delta \omega)}
	\end{align*}
	Now we note the value of $\pi_t$ (here $\omega \in (0,1)$):
	\begin{align*}
	\pi_t & = \left( 1+\frac{4}{\omega} \right)nG^2H^2 + \left( 1 + \frac{\omega}{4} \right) (1 - \omega)\left( 1+\frac{4}{\omega} \right) nG^2H^2 + \left( 1+\frac{\omega}{4} \right)\omega n c_t\\
	& = \left( 1+\frac{4}{\omega} \right) nG^2H^2 \left[ 2 - \frac{3\omega}{4} - \frac{\omega^2}{4} \right] + \left( 1+\frac{\omega}{4} \right)\omega n c_t\\
	& \leq 2nG^2H^2\left( 1+\frac{4}{\omega} \right)\left( 1-\frac{3\omega}{8} \right)+ \frac{5\omega n c_t}{4} = 2nG^2H^2\left( 1-\frac{3\omega}{8} + \frac{4}{\omega} - \frac{3}{2} \right) + \frac{5\omega n c_t}{4} \\
	& \leq \frac{8nG^2H^2}{\omega} + \frac{5\omega n c_t}{4}
	\end{align*}
	Thus we have:
	\begin{align*}
	max \{ \pi_1 (\gamma^*) , \pi_2 (\gamma^*) \} & \leq 1 - \frac{\delta^2\omega}{4(64 \delta + \delta^2 + 16 \beta^2 + 8 \delta \beta^2 - 16\delta \omega)}
	\intertext{from the value of $\gamma^*$ calculated above, we note that $ \frac{\delta^2\omega}{4(64 \delta + \delta^2 + 16 \beta^2 + 8 \delta \beta^2 - 16\delta \omega)} = \frac{\delta \gamma^*}{8}$. Using crude estimates $\delta \leq 1, \omega \geq 0, \beta \leq2$, we thus have: }
		max \{ \pi_1 (\gamma^*) , \pi_2 (\gamma^*) \} & \leq 1 - \frac{\gamma^* \delta}{8}   \leq 1 - \frac{\delta^2 \omega}{644}
	\end{align*}
\end{proof}
\begin{lemma}  \label{suppl_lem_e_seq_li} (Variant of \cite[Lemma 22]{koloskova_decentralized_2019-1})
	Consider the sequence \{$e_{I_{(t)}}$\} given by
	\begin{align*}
	e_{I_{(t+1)}} \leq \left( 1-\frac{p}{2} \right)e_{I_{(t)}} + \frac{2}{p}\eta_{I_{(t)}}^2A_{I_{(t)}}
	\end{align*}
	where $ \mathcal{I}_T = \{ I_{(1)}, I_{(2)} , \hdots , I_{(t)}, \hdots  \} \in [T] $ denotes the set of synchronization indices. For a parameter $p >0$, an increasing positive sequence $\{A_t\}_{t=0}^{T-1}$, stepsize $\eta_t = \frac{b}{t+a}$ with parameter $a \geq \frac{5H}{p}$ and arbitrary $b >0$, we have:
	\begin{align*}
	e_{I_{(t)}} \leq \frac{20}{p^2}A_{I_{(t)}}\eta_{I_{(t)}}^2
	\end{align*}
\end{lemma}
\begin{proof}
	We will proceed the proof by induction. Note that for t=0, $e_{I_{(0)}}:=0$ (we assume first synchronization index is 0), thus statement is true. Assume the statement holds for index $I_{(t)}$, then for index $I_{(t+1)}$:  
	\begin{align} \label{cvx_lemm_rec_interim1}
	e_{I_{(t+1)}} \leq \left( 1- \frac{p}{2} \right)e_{I_{(t)}} + \frac{2}{p}A_{(t)}\eta_{I_{(t)}}^2 & \leq \left( 1- \frac{p}{2} \right)\frac{20}{p^2}A_{(t)}\eta_{I_{(t)}}^2 + \frac{2}{p}A_{(t)}\eta_{I_{(t)}}^2 \notag \\
	& = \frac{A_{(t)}\eta_{I_{(t)}}^2}{p^2}(20-8p)  \stackrel{(p \geq \frac{5H}{a})}{\leq} \frac{20A_{(t)}\eta_{I_{(t)}}^2}{p^2}\left( 1- \frac{2H}{a} \right)
	\end{align}
	Now, we note the following:
	\begin{align*}
	(a+ I_{(t+1)})^2 \left( 1- \frac{2H}{a} \right) & \leq (a+ I_{(t)} + H )^2 \left( 1- \frac{2H}{a} \right) \\
	& = (a+ I_{(t)})^2 + 2H(a+ I_{(t)}) + H^2 \\
	& \qquad - \left[ \frac{2H(a+ I_{(t)})^2}{a} + \frac{4H^2(a+ I_{(t)})}{a} + \frac{2H^3}{a} \right] \\
	& \leq (a+ I_{(t)})^2 + 2H(a+ I_{(t)}) + H^2 - \left[ 2H(a+ I_{(t)}) + 4H^2 \right] \\
	& \leq (a+ I_{(t)})^2
	\end{align*}
	Thus, for $\eta_{I_{(t+1)}} = \frac{b}{a + I_{(t+1)} }$, we get:
	\begin{align*}
	\eta_{I_{(t)}}^2 \left( 1- \frac{2H}{a} \right) \leq \eta_{I_{(t+1)}}^2
	\end{align*}
	Substituting the above bound in the bound for $e_{I_{(t+1)}}$ in \eqref{cvx_lemm_rec_interim1} and using the fact that $A_t$ is an increasing function:
	\begin{align*}
	e_{I_{(t+1)}} \leq \frac{20A_{I_{(t)}}\eta_{I_{(t+1)}}^2}{p^2} \leq \frac{20A_{I_{(t+1)}}\eta_{I_{(t+1)}}^2}{p^2}
	\end{align*}
	Thus, by induction : $e_{I_{(t)}} \leq \frac{20A_{I_{(t)}}\eta_{I_{(t)}}^2}{p^2}$ for all $I_{(t)} \in \mathcal{I}_T$.	
\end{proof}
\subsection{Proof of Theorem  \ref{thm_cvx_li} (Strongly convex objective)} \label{proof_thm_cvx_li}
To proceed with the proof for Theorem, we first note the following lemma from \cite[Lemma 20]{koloskova_decentralized_2019-1}.
\begin{lemma} \label{suppl_cvx_lemm}
	Let $ \{ \bx_t ^{(i)}  \}_{t=0}^{T-1} $ be generated according to Algorithm \ref{alg_dec_sgd_li} with stepsize $\eta_t $ and define $\bar{\bx}_t = \frac{1}{n} \sum_{i=1}^{n} \bx_t^{(i)} $. Then we have the following result for $\bar{\bx}^{(t)}$ :
	\begin{align*}
	\mathbb{E}_{\boldsymbol{\xi}^{(t)}} \Vert \Bar{\bx}^{(t+1)} - \bx^* \Vert^2 \leq & \left( 1-\frac{\eta_t \mu}{2} \right) \Vert \Bar{\bx}^{(t)} - \bx^* \Vert^2 + \frac{\eta_t^2 \Bar{\sigma}^2}{n} - 2\eta_t (1-2L\eta_t) (f(\Bar{\bx}^{(t)}) - f^*)  \\
	&\hspace{2cm} + \eta_t \left(  \frac{2\eta_t L^2 + L + \mu}{n}  \right) \sum_{j=1}^n \Vert \Bar{\bx}^{(t)} - \bx_j^{(t)} \Vert^2
	\end{align*}
	where $\boldsymbol{\xi}^{(t)}$:=$ \{ \xi_1^{(t)},\xi_2^{(t)}, \hdots, \xi_n^{(t)}  \}$ is the set of random samples at each worker at time step $t$ and $\bar{\sigma}^2 = \frac{1}{n} \sum_{i=1}^{n} \sigma_i^2$
\end{lemma}
\begin{proof} Consider expectation taken over sampling at time instant $t$: ${\boldsymbol{\xi}^{(t)}} = \{ \xi_1^{(t)}, \xi_2^{(t)}, \hdots, \xi_n^{(t)} \}$ and using $\Bar{\bX}^{(t)}  = \Bar{\bX}^{(t+\frac{1}{2})} $ (from (\ref{mean_seq_iter})) which gives: $ \Bar{\bx}^{(t+1)} =  \frac{1}{n}\sum_{j=1}^n \nabla F_j(\bx_j^{(t)},\xi_j^{(t)}) $ ,  we have:
	\begin{align}
	\mathbb{E}_{\boldsymbol{\xi}^{(t)}}& \Vert \Bar{\bx}^{(t+1)} - \bx^* \Vert^ 2 \notag \\
	& =  \mathbb{E}_{\boldsymbol{\xi}^{(t)}} \left\Vert \Bar{\bx}^{(t)} - \frac{\eta_t}{n}\sum_{j=1}^n \nabla F_j(\bx_j^{(t)},\xi_j^{(t)}) - \bx^* \right\Vert^2  \notag\\
	& = \mathbb{E}_{\boldsymbol{\xi}^{(t)}} \left\Vert \Bar{\bx}^{(t)} - \bx^* - \frac{\eta_t}{n}\sum_{j=1}^n \nabla f_j(\bx_j^{(t)}) + \frac{\eta_t}{n}\sum_{j=1}^n \nabla f_j(\bx_j^{(t)}) - \frac{\eta_t}{n}\sum_{j=1}^n \nabla F_j(\bx_j^{(t)},\xi_j^{(t)})  \right\Vert^2  \notag\\
	& =  \left\Vert \Bar{\bx}^{(t)} - \bx^* - \frac{\eta_t}{n}\sum_{j=1}^n \nabla f_j(\bx_j^{(t)}) \right\Vert^2 + \eta_t^2 \mathbb{E}_{\boldsymbol{\xi}^{(t)}} \left\Vert \frac{1}{n}\sum_{j=1}^n \nabla f_j(\bx_j^{(t)}) - \frac{1}{n}\sum_{j=1}^n \nabla F_j(\bx_j^{(t)},\xi_j^{(t)}) \right\Vert^2 \notag\\
	& \hspace{1cm} + \frac{2\eta_t}{n}\mathbb{E}_{\boldsymbol{\xi}^{(t)}} \left\langle \Bar{\bx}^{(t)} - \bx^* - \frac{\eta_t}{n}\sum_{j=1}^n \nabla f_j(\bx_j^{(t)}) , \sum_{j=1}^n \nabla f_j(\bx_j^{(t)}) - \sum_{j=1}^n \nabla F_j(\bx_j^{(t)},\xi_j^{(t)}) \right\rangle \label{suppl_cvx_lemm_total}
	\end{align}
	The last term in (\ref{suppl_cvx_lemm_total}) is zero as $\mathbb{E}_{\xi_i^{(t)}} \nabla F_i(\bx_i^{(t)}, \xi_i^{(t)}) = \nabla f_i(\bx_i^{(t)}) $ for all $i \in [n]$. The second term in (\ref{suppl_cvx_lemm_total}) can be bounded via the variance bound (\ref{bound_var}) by $ \frac{\eta_t^2 \bar{\sigma}^2}{n}$.\\
	We thus consider the first term in the (\ref{suppl_cvx_lemm_total}) :
	\begin{align} \label{suppl_cvx_lemm_firstterm}
	\left\Vert \Bar{\bx}^{(t)} - \bx^* - \frac{\eta_t}{n}\sum_{j=1}^n \nabla f_j(\bx_j^{(t)}) \right\Vert^2 = \Vert \Bar{\bx}^{(t)} - \bx^* \Vert^2 + \eta_t^2  \underbrace{\left\Vert \frac{1}{n}\sum_{j=1}^n \nabla f_j(\bx_j^{(t)}) \right\Vert^2}_{T_1} \notag \\
	- \underbrace{2\eta_t  \left\langle \Bar{\bx}^{(t)} - \bx^*,\frac{1}{n}\sum_{j=1}^n \nabla f_j(\bx_j^{(t)}) \right\rangle}_{T_2}
	\end{align}
	To bound $T_1$ in (\ref{suppl_cvx_lemm_firstterm}), note that:
	\begin{align}
	T_1 &= \left\Vert \frac{1}{n}\sum_{j=1}^n (\nabla f_j(\bx_j^{(t)}) - \nabla f_j(\Bar{\bx}^{(t)}) + \nabla f_j(\Bar{\bx}^{(t)}) - \nabla f_j(\bx^*) ) \right\Vert^2  \notag \\
	& \leq \frac{2}{n}\sum_{j=1}^n \Vert \nabla f_j(\bx_j^{(t)}) - \nabla f_j(\Bar{\bx}^{(t)}) \Vert^2 + 2 \left\Vert \frac{1}{n}\sum_{j=1}^n \nabla f_j(\Bar{\bx}^{(t)}) - \frac{1}{n}\sum_{j=1}^n \nabla f_j(\bx^*) \right\Vert^2  \notag \\
	& \leq \frac{2L^2}{n}\sum_{j=1}^n \Vert \bx_j^{(t)} - \Bar{\bx}^{(t)} \Vert^2 + 4L  (f(\Bar{\bx}^{(t)}) - f^*)  \label{suppl_cvx_t1_bound}
	\end{align}
	where in the last inequality, we've used $L-$Lipschitz gradient property of $f_j's$ to bound the first term and optimality of $\bx^*$ for $f$ (i.e $\nabla f(\bx^*) = 0$) and $L-$smoothness property (\ref{l_smooth_prop}) of $f$  to bound the second term as:
	 $ \left\Vert \frac{1}{n}\sum_{j=1}^n \nabla f_j(\Bar{\bx}^{(t)}) - \frac{1}{n}\sum_{j=1}^n \nabla f_j(\bx^*) \right\Vert^2 = \verts{ \nabla f (\Bar{\bx}^{(t)} ) - \nabla f (\bx^*)  }^2   \leq 2L \left( f(\Bar{\bx}^{(t)} ) -f^*  \right) $.\\
	To bound $T_2$ in (\ref{suppl_cvx_lemm_firstterm}), note that:
	\begin{align}
	-\frac{1}{\eta_t}T_2 & = -\frac{2}{n} \sum_{j=1}^n \left[ \left\langle \Bar{\bx}^{(t)} - \bx_j^{(t)}, \nabla f_j(\bx_j^{(t)}) \right\rangle + \left\langle \bx_j^{(t)} - \bx^*, \nabla f_j(\bx_j^{(t)}) \right\rangle \right] \notag  \\
	\intertext{Using expression for $\mu$-strong convexity (\ref{mu_strong_cvx}) and $L$-smoothness (\ref{l_smooth}) for $f_j , \, j \in [n] $  : }
	& \leq -\frac{2}{n} \sum_{j=1}^n \left[  f_j(\Bar{\bx}^{(t)}) - f_j(\bx_j^{(t)}) - \frac{L}{2}\Vert \Bar{\bx}^{(t)} - \bx_j^{(t)}  \Vert^2  + f_j(\bx_j^{(t)}) - f_j(\bx^*) + \frac{\mu}{2}\Vert \bx_j^{(t)} - \bx^*  \Vert^2 \right] \notag \\
	& = -2  (f(\Bar{\bx}^{(t)}) - f(\bx^*)) + \frac{L+ \mu}{n}\sum_{j=1}^n \Vert \Bar{\bx}^{(t)} - \bx_j^{(t)}  \Vert^2   - \frac{\mu}{n} \sum_{j=1}^n \left[\Vert \Bar{\bx}^{(t)} - \bx_j^{(t)}  \Vert^2  +  \Vert \bx_j^{(t)} - \bx^*  \Vert^2 \right] \notag \\
	& \leq  -2  (f(\Bar{\bx}^{(t)}) - f(\bx^*)) + \frac{L+ \mu}{n}\sum_{j=1}^n \Vert \Bar{\bx}^{(t)} - \bx_j^{(t)}  \Vert^2   - \frac{\mu}{2n} \sum_{j=1}^n \left[\Vert \Bar{\bx}^{(t)} - \bx^*  \Vert^2 \right] \notag \\
	& =  -2  (f(\Bar{\bx}^{(t)}) - f(\bx^*)) + \frac{L+ \mu}{n}\sum_{j=1}^n \Vert \Bar{\bx}^{(t)} - \bx_j^{(t)}  \Vert^2   - \frac{\mu}{2} \Vert \Bar{\bx}^{(t)} - \bx^*  \Vert^2 \label{suppl_cvx_t2_bound}
	\end{align}
	Substituting (\ref{suppl_cvx_t1_bound}),(\ref{suppl_cvx_t2_bound}) in (\ref{suppl_cvx_lemm_firstterm}) and using it in (\ref{suppl_cvx_lemm_total}), we get the desired result :
	\begin{align*}
	\mathbb{E}_{\boldsymbol{\xi}^{(t)}} \Vert \Bar{\bx}^{(t+1)} - \bx^* \Vert^2 &\leq \left( 1-\frac{\eta_t \mu}{2} \right) \Vert \Bar{\bx}^{(t)} - \bx^* \Vert^2 + \frac{\eta_t^2 \Bar{\sigma}^2}{n} - 2\eta_t (1-2L\eta_t) (f(\Bar{\bx}^{(t)}) - f^*) \\
	&\qquad + \eta_t \left(  \frac{2\eta_t L^2 + L + \mu}{n}  \right) \sum_{j=1}^n \Vert \Bar{\bx}^{(t)} - \bx_j^{(t)} \Vert^2
	\end{align*}
\end{proof}
We now proceed to the main proof for Theorem ~\ref{thm_cvx_li}.
\begin{proof}[Proof of Theorem ~\ref{thm_cvx_li}]
	From Lemma \ref{suppl_cvx_lemm}, we have that :
	\begin{align*}
	\mathbb{E}_{\boldsymbol{\xi}^{(t)}} \Vert \Bar{\bx}^{(t+1)} - \bx^* \Vert^2 &\leq \left( 1-\frac{\eta_t \mu}{2} \right) \Vert \Bar{\bx}^{(t)} - \bx^* \Vert^2 + \frac{\eta_t^2 \Bar{\sigma}^2}{n} - 2\eta_t (1-2L\eta_t) (f(\Bar{\bx}^{(t)}) - f^*) \\
	&\qquad+ \eta_t \left(  \frac{2\eta_t L^2 + L + \mu}{n}  \right) \sum_{j=1}^n \Vert \Bar{\bx}^{(t)} - \bx_j^{(t)} \Vert^2
	\end{align*}
	Taking expectation w.r.t the whole process gives us:
	\begin{align} \label{suppl_cvx_lemm_use_li}
	\mathbb{E}\Vert \Bar{\bx}^{(t+1)} - \bx^* \Vert^2 &\leq \left( 1-\frac{\eta_t \mu}{2} \right) \mathbb{E} \Vert \Bar{\bx}^{(t)} - \bx^* \Vert^2 + \frac{\eta_t^2 \Bar{\sigma}^2}{n} - 2\eta_t (1-2L\eta_t) (\mathbb{E}f(\Bar{\bx}^{(t)}) - f^*) \notag \\
	&\qquad + \eta_t \left(  \frac{2\eta_t L^2 + L + \mu}{n}  \right) \sum_{j=1}^n \mathbb{E}\Vert \Bar{\bx}^{(t)} - \bx_j^{(t)} \Vert^2
	\end{align}
Let $I_{(t+1)_0}$ denote the latest synchronization step before or equal to $(t+1)$. Then we have:
\begin{align*}
\bX^{(t+1)} & = \bX^{I_{(t+1)_0}} - \sum_{t' = I_{(t+1)_0}}^{t}\eta_{t'}\partial F(\bX^{(t')}, \boldsymbol{\xi}^{(t')} ) \\
\Bar{\bX}^{(t+1)} & = \bar{\bX}^{I_{(t+1)_0}} - \sum_{t' = I_{(t+1)_0}}^{t}\eta_{t'}\partial F(\bX^{(t')}, \boldsymbol{\xi}^{(t')} ) \frac{\mathbbm{1}\mathbbm{1}^T}{n} 
\end{align*} 
Thus the following holds:
\begin{align*} \label{suppl_cvx_li_gammat_temp}
\mathbb{E} & \Vert \bX^{(t+1)} -  \Bar{\bX}^{(t+1)} \Vert_F^2 2 = \mathbb{E} \left\Vert \bX^{I_{(t+1)_0}} - \Bar{\bX}^{I_{(t+1)_0}} -\sum_{t' = I_{(t+1)_0}}^{t} \eta_{t'}\partial F(\bX^{(t')}, \boldsymbol{\xi}^{(t')} ) \left( \mathbf{I} - \frac{1}{n} \mathbbm{1}\mathbbm{1}^T  \right) \right\Vert_F^2 \\
& \hspace{2cm}  \leq 2 \mathbb{E}\Vert \bX^{I_{(t+1)_0}} - \Bar{\bX}^{I_{(t+1)}}  \Vert_F^2 + 2\mathbb{E} \left\Vert \sum_{t' = I_{(t+1)_0}}^{t} \eta_{t'}\partial F(\bX^{(t')}, \boldsymbol{\xi}^{(t')} ) \left( \mathbf{I} - \frac{1}{n} \mathbbm{1}\mathbbm{1}^T  \right) \right\Vert_F^2 \notag 
\end{align*}
	Using (\ref{bound_frob_mult}) for the second term in above and noting that $\mathbb{E} \left\Vert \sum_{t' = I_{(t+1)_0}}^{{t}} \eta_{t'}\partial F(\bX^{(t')}, \boldsymbol{\xi}^{(t')} )\right\Vert_F^2  \leq \eta_{I_{(t+1)_0}} nH^2G^2 $ and $\Vert \frac{\mathbbm{1}\mathbbm{1}^T}{n} - \mathbf{I} \Vert_2^2 = 1 $ from (\ref{bound_gap_grad}) and (\ref{bound_W_mat}) (with $k=0$) respectively, we get: 
	\begin{align}
\mathbb{E} \Vert \bX^{(t+1)} -  \Bar{\bX}^{(t+1)} \Vert_F^2 & {\leq} 2 \mathbb{E}\Vert \bX^{I_{(t+1)_0}} - \Bar{\bX}^{I_{(t+1)_0}}  \Vert_F^2 + 2H^2n \eta_{I_{(t+1)_0}}^2G^2
\end{align}
 For $A_{I_{(t+1)_0}} = 2nG^2H^2 + \frac{p}{2}\left(\frac{8nG^2H^2}{\omega} + \frac{5\omega n c_{I_{(t+1)_0}}}{4}\right)$, the first term in (\ref{suppl_cvx_li_gammat_temp}) can be bounded by Lemma \ref{lem_dec_li_sgd} as:
\begin{align*}
\mathbb{E}\Vert \bX^{I_{(t+1)_0}} - \Bar{\bX}^{I_{(t+1)_0}}  \Vert_F^2  \leq \frac{20A_{I_{(t+1)_0}}}{p^2}\eta_{I_{(t+1)_0}}^2
\end{align*}
Substituting above bound in (\ref{suppl_cvx_li_gammat_temp}), we have:
\begin{align*}
\mathbb{E} \Vert \bX^{(t+1)} -  \Bar{\bX}^{(t+1)} \Vert_F^2 \leq \frac{40A_{I_{(t+1)_0}}}{p^2}\eta_{I_{(t+1)_0}}^2 + 2H^2n \eta_{I_{(t+1)_0}}^2G^2
\end{align*}
Using the above bound for the last term in (\ref{suppl_cvx_lemm_use_li}), we have $\mathbb{E} \Vert \bX^{(t)} -  \Bar{\bX}^{(t)} \Vert_F^2 = \mathbb{E} \left[ \sum_{j=1}^n \Vert \Bar{\bx}^{(t)} - \bx_j^{(t)} \Vert^2\right] \leq \left(\frac{40A_{I_{(t)_0}}}{p^2}\eta_{I_{(t)_0}}^2 + 2H^2n \eta_{I_{(t)_0}}^2G^2\right) $ where $I_{(t)_0}$ denotes the last synchronization step before or equal to $t$. This gives us:
\begin{align}
\mathbb{E} \Vert \Bar{\bx}^{(t+1)} - \bx^* \Vert^2 \leq & \left( 1-\frac{\eta_t \mu}{2} \right) \mathbb{E}\Vert \Bar{\bx}^{(t)} - \bx^* \Vert^2 + \frac{\eta_t^2 \Bar{\sigma}^2}{n} - 2\eta_t (1-2L\eta_t) ( \mathbb{E}   f(\Bar{\bx}^{(t)}) - f^*)  \notag \\
& + \eta_t \left(  \frac{2\eta_t L^2 + L + \mu}{n}  \right) \left( \frac{40A_{I_{(t)_0}}}{p^2} + 2nH^2 G^2 \right)\eta_{I_{(t)_0}}^2 \label{cvx_proof_interim1}
\end{align} 
To proceed with the proof of the theorem, we note that $A_{I_{(t)_0}} \leq A_t $ as $I_{(t)_0}$ denotes the last synchronization index before $t$ and $\{ A_t \}_{t=0}^{T-1}$ is an increasing sequence (as $\{ c_t \}_{t=0}^{T-1}$ is increasing sequence). We also note the following relation for the learning rate:
\begin{align*}
\frac{\eta_{I_{(t)_0}}}{\eta_{t}} = \frac{a+t}{a+ I_{(t)_0}} \leq \frac{a+I_{(t)_0}+H}{a+ I_{(t)_0}} = \left( 1 + \frac{H}{a+ I_{(t)_0}} \right) \leq \left( 1 + \frac{H}{a} \right) \stackrel{ (a \geq H) } {\leq} 2
\end{align*}
Using the above relation in  \eqref{cvx_proof_interim1}, we get:
\begin{align*}
\mathbb{E} \Vert \Bar{\bx}^{(t+1)} - \bx^* \Vert^2 \leq & \left( 1-\frac{\eta_t \mu}{2} \right) \mathbb{E}\Vert \Bar{\bx}^{(t)} - \bx^* \Vert^2 + \frac{\eta_t^2 \Bar{\sigma}^2}{n} - 2\eta_t (1-2L\eta_t) ( \mathbb{E}  f(\Bar{\bx}^{(t)}) - f^*) \\
& + 4 \eta_t \left(  \frac{2\eta_t L^2 + L + \mu}{n}  \right) \left( \frac{40A_t}{p^2} + 2nH^2 G^2 \right)\eta_t^2
\end{align*}
For $\eta_t = \frac{8}{\mu (a+t)}$ and $a \geq \text{max} \{ \frac{32L}{\mu}, \frac{5H}{p}\} $, we have $\eta_t \leq \frac{1}{4L}$. This implies :
$ 2L\eta_t - 1 \leq -\frac{1}{2}$ and  $(2\eta_t L^2 + L + \mu) \leq (2L + \mu) $. Using these in the above equation gives:
\begin{align*}
\mathbb{E} \Vert \Bar{\bx}^{(t+1)} - \bx^* \Vert^2 & \leq \left( 1-\frac{\eta_t \mu}{2} \right) \mathbb{E} \Vert \Bar{\bx}^{(t)} - \bx^* \Vert^2 - \eta_t (\mathbb{E} f(\Bar{\bx}^{(t)}) - f^*) + \frac{\eta_t^2 \Bar{\sigma}^2}{n} \\
& \qquad + 4 \eta_t^3 \left(  \frac{2L + \mu}{n}  \right) \left( \frac{40A_t}{p^2} + 2nH^2 G^2 \right)
\end{align*}
Substituting value of $A_t = 2nG^2H^2 + \frac{p}{2} \left(\frac{8nG^2H^2}{\omega} + \frac{5\omega n c_t}{4}\right) $ , we get:
\begin{align*}
\mathbb{E} \Vert \Bar{\bx}^{(t+1)} - \bx^* \Vert^2 & \leq \left( 1-\frac{\eta_t \mu}{2} \right) \mathbb{E} \Vert \Bar{\bx}^{(t)} - \bx^* \Vert^2 - \eta_t (\mathbb{E}f(\Bar{\bx}^{(t)}) - f^*) + \frac{\eta_t^2 \Bar{\sigma}^2}{n}  \\
&\qquad+ 8 \eta_t^3 \left(  \frac{2L + \mu}{n}  \right) \left( \frac{40}{p^2}+ \frac{80}{p\omega} + \frac{50\omega c_t}{4pG^2H^2} + 1 \right)nG^2H^2
\end{align*}
We use Lemma \ref{suppl_cvx_lemma_stitch} for the sequence relation above by defining:
\begin{align*}
a_t & = \mathbb{E} \Vert \Bar{\bx}^{(t)} - \bx^* \Vert^2 \\
e_t & = \mathbb{E}f(\Bar{\bx}^{(t)}) - f^* \\
P & = 1 \\ 
Q & = \frac{ \Bar{\sigma}^2}{n} \\
R & = 8 \left(  {2L + \mu}  \right) \left( \frac{40}{p^2}+ \frac{80}{p\omega} + 1 \right)G^2H^2 \\
U_t & =  100 \left(  \frac{2L + \mu}{p}  \right)\omega c_t
\end{align*}
For $w_t = (a+t)^2$, $a_0 =  \Vert \Bar{\bx}^{(0)} - \bx^* \Vert^2 $ and $e_t = \mathbb{E}f(\Bar{\bx}^{(t)}) - f^*$, this gives us the relation:
\begin{align*}
\frac{1}{S_T} \sum_{t=0}^{T-1}w_t e_t & \leq \frac{\mu a^3}{8S_T}a_0^2 + \frac{4T(T+2a)}{\mu S_T}\frac{ \Bar{\sigma}^2}{n} + \frac{512T}{\mu^2 S_T} \left(  {2L + \mu}  \right) \left( \frac{40}{p^2}+ \frac{80}{p\omega} + 1 \right)G^2H^2 \\
& \qquad + \frac{6400c_0 \omega T ^{(2-\epsilon)}}{\mu^2 (2-\epsilon)S_T} \left(  \frac{2L + \mu}{p}  \right)
\end{align*}
where $\epsilon \in (0,1)$. From the convexity of $f$, we finally have:
\begin{align*}
		\mathbb{E}f(\bx_{avg}^{(T)}) - f^* & \leq \frac{\mu a^3}{8S_T}a_0^2 + \frac{4T(T+2a)}{\mu S_T}\frac{ \Bar{\sigma}^2}{n} + \frac{512T}{\mu^2 S_T} \left(  {2L + \mu}  \right) \left( \frac{40}{p^2}+ \frac{80}{p\omega} + 1 \right)G^2H^2 \\
		& \qquad + \frac{6400c_0 \omega T ^{(2-\epsilon)}}{\mu^2 (2-\epsilon)S_T} \left(  \frac{2L + \mu}{p}  \right)
\end{align*}
where  $\Bar{\bx}^{(T)}_{avg} = \frac{1}{S_T} \sum_{t=0}^{T-1}w_t \bar{\bx}^{(t)}$.
We finally use the fact that $p \leq \omega$ (as $\delta \leq 1$ and $p:= \frac{\gamma^* \delta}{8}$ with $\gamma^* \leq \omega $). This implies the above expression as:
\begin{align*}
\mathbb{E}f(\bx_{avg}^{(T)}) - f^* & \leq \frac{\mu a^3}{8S_T}a_0^2 + \frac{4T(T+2a)}{\mu S_T}\frac{ \Bar{\sigma}^2}{n} + \frac{512T}{\mu^2 S_T} \left(  2L + \mu  \right) \left( \frac{160}{p^2}\right)G^2H^2 \\
& \qquad +\frac{6400 c_0 \omega T ^{(2-\epsilon)}}{\mu^2 (2-\epsilon)S_T} \left(  \frac{2L + \mu}{p}  \right)
\end{align*}
This completes proof of Theorem \ref{thm_cvx_li}.
\end{proof}
\begin{lemma} \label{suppl_cvx_lemma_stitch}
	(Variant of \cite[Lemma 3.3]{stich_sparsified_2018})
	Let $\{ a_t \}_{t \geq 0}, a_t \geq 0, e_t \}_{t \geq 0}, e_t \geq 0 $ be sequences satisfying :
	\begin{align*}
	a_{t+1} \leq \left(1- \frac{\mu \eta_t}{2} \right) a_{t} - \eta_t e_t P + \eta_t^2Q + \eta_t^3R + \eta_t^3U_t ,
	\end{align*}
	Let stepsize $\eta_t = \frac{8}{\mu (a+t)} $ and $U_t = U c_t $, constants $P>0,Q,R \geq 0,U \geq 0, \mu>0, a>1 $ and  $c_t  \geq 0$ for all $t$ with $c_t \sim o(t)$, specifically, assume that $c_t \leq c_0 t^{(1-\epsilon)}$ for some $c_0 \geq 0$ and $\epsilon \in (0,1)$. Then it holds that:
	\begin{align*}
	\frac{P}{S_T} \sum_{t=0}^{T-1}w_t e_t \leq \frac{\mu a^3}{8S_T}a_0 + \frac{4T(T+2a)}{\mu S_T}Q + \frac{64T}{\mu^2 S_T}R + \frac{64c_0T^{(2-\epsilon)}}{\mu^2 (2-\epsilon) S_T }U,
	\end{align*}
	where $w_t = (a+t)^2$ and $S_T := \sum_{t=0}^{T-1}w_t = \frac{T}{6} (2T^2+6aT-3T+6a^2-6a+1) \geq \frac{1}{3}T^3$
\end{lemma}
\begin{proof}
	The proof follows some steps similar to that of \cite[Lemma 3.3]{stich_sparsified_2018}. We first multiply both sides of the expression by $\frac{w_t}{\eta_t}$ which gives:
		\begin{align*}
	a_{t+1} \frac{w_t}{\eta_t} \leq \left(1- \frac{\mu \eta_t}{2} \right) \frac{w_t}{\eta_t} a_{t} - w_t e_t P + w_t \eta_tQ + w_t \eta_t^2R + w_t \eta_t^2U_t 
	\end{align*}
	Using the fact that $\left(1- \frac{\mu \eta_t}{2} \right) \frac{w_t}{\eta_t} \leq \frac{w_{t-1}}{\eta_{t-1}} $ (shown in \cite[Lemma 3.3]{stich_sparsified_2018} ) and then substituting the value of $\left(1- \frac{\mu \eta_t}{2} \right) \frac{w_{t-1}}{\eta_{t-1}} $ recursively, we get:
	\begin{align*}
	a_{T} \frac{w_{T-1}}{\eta_{T-1}} \leq \left(1- \frac{\mu \eta_0}{2} \right) \frac{w_0}{\eta_0} a_{0} -  \sum_{t=0}^{T-1} w_t e_t P + \sum_{t=0}^{T-1} w_t \eta_tQ + \sum_{t=0}^{T-1} w_t \eta_t^2R + \sum_{t=0}^{T-1} w_t \eta_t^2U_t 
	\end{align*}
	Rearranging the terms in above and noting that $\frac{w_0}{\eta_0} = \frac{\mu a^3}{8} $, we get:
		\begin{align} \label{cvx_stich_lemma_interim1}
	P \sum_{t=0}^{T-1} w_t e_t  \leq  \frac{\mu a^3}{8} + \sum_{t=0}^{T-1} w_t \eta_tQ + \sum_{t=0}^{T-1} w_t \eta_t^2R + \sum_{t=0}^{T-1} w_t \eta_t^2U_t 
	\end{align}
	We now bound the terms in the RHS of \eqref{cvx_stich_lemma_interim1}. The bounds for the second and third term are given in \cite[Lemma 3.3]{stich_sparsified_2018}, which are:
	\begin{align*}
		\sum_{t=0}^{T-1} w_t \eta_tQ  & \leq \frac{4QT(T+2a)}{\mu} \\
		\sum_{t=0}^{T-1} w_t \eta_t^2R & \leq \frac{64RT}{\mu^2}
	\end{align*}
	To bound the last term in RHS of \eqref{cvx_stich_lemma_interim1}, we note that $U_t = U c_t$ where $c_t \sim o(t)$ and $U \geq 0$ is a constant. Thus, we can assume that $c_t \leq c_0 t^{1-\epsilon}$ for some $\epsilon \in (0,1)$, and proceed to bound the terms as:
	\begin{align*}
	\sum_{t=0}^{T-1} w_t \eta_t^2U_t \leq U \sum_{t=0}^{T-1} w_t \eta_t^2 c_0t^{(1-\epsilon)} = \frac{64U}{\mu^2} \sum_{t=0}^{T-1}  t^{(1-\epsilon)} \leq \frac{64Uc_0}{\mu^2} \int_{0}^{T} t^{(1-\epsilon)} dt = \frac{64Uc_0 T^{(2-\epsilon)}}{\mu^2 (2-\epsilon)} 
	\end{align*}
	Substituting these bounds in \eqref{cvx_stich_lemma_interim1} yields:
		\begin{align*}
	P \sum_{t=0}^{T-1} w_t e_t  \leq  \frac{\mu a^3}{8} + \frac{4QT(T+2a)}{\mu} + \frac{64RT}{\mu^2} + \frac{64Uc_0 T^{(2-\epsilon)}}{\mu^2 (2-\epsilon)}  
	\end{align*}
	Dividing both sides in above by $S_T := \sum_{t=0}^{T-1}w_t = \frac{T}{6} (2T^2+6aT-3T+6a^2-6a+1) \geq \frac{1}{3}T^3$, we have:
			\begin{align*}
	\frac{P}{S_T} \sum_{t=0}^{T-1}w_t e_t \leq \frac{\mu a^3}{8S_T}a_0 + \frac{4T(T+2a)}{\mu S_T}Q + \frac{64T}{\mu^2 S_T}R + \frac{64c_0T^{(2-\epsilon)}}{\mu^2 (2-\epsilon) S_T }U 
	\end{align*}
\end{proof}
\subsection{Proof of Lemma \ref{lemm_dec_li_sgd_fix}} \label{proof_lemm_dec_li_sgd_fix}
\begin{lemma*} (Restating Lemma \ref{lemm_dec_li_sgd_fix} )
		Let $ \{ \bx_t ^{(i)}  \}_{t=0}^{T-1} $ be generated according to Algorithm \ref{alg_dec_sgd_li} under assumptions of Theorem \ref{thm_noncvx_fix_li} with constant stepsize $\eta$ and threshold function $c_t \leq  \frac{1}{\eta^{(1-\epsilon)}}  $ for all $t$, for some $\epsilon \in (0,1)$ and define $\bar{\bx}_t = \frac{1}{n} \sum_{i=1}^{n} \bx_t^{(i)} $.
	Consider the set of synchronization indices as $\mathcal{I}_T$ =  $\{ I_{(1)},I_{(2)}, \hdots, I_{(t)}, \hdots  \}$. Then for any $I_{(t)} \in \mathcal{I}_T$, we have:
	\begin{align*}
	\sum_{j=1}^n \mathbb{E}\Vert \bar{\bx}^{I_{(t)}} - \bx^{I_{(t)}}_{j} \Vert^2 = \mathbb{E} \Vert \bX^{I_{(t)}} - \Bar{\bX}^{I_{(t)}} \Vert_F^2  \leq \frac{4A\eta^2}{p^2}
	\end{align*}
	where  $p = \frac{\delta \gamma}{8} $, $\delta := 1 - | \lambda_2(W)|$, $\omega$ is compression parameter for operator $\C$ and $A=2nG^2H^2 + \frac{p }{2} \left(\frac{8nG^2H^2}{\omega} + \frac{5\omega n }{4 \eta^{1-\epsilon}} \right) $.
\end{lemma*}
\begin{proof}[Proof of Lemma \ref{lemm_dec_li_sgd_fix}]
	We use the same steps for the Proof of Lemma \ref{lem_dec_li_sgd} with $\eta_t = \eta$ and $c_t \leq \frac{1}{\eta^{1-\epsilon}} $ (from some $\epsilon \in (0,1)$) till (\ref{suppl_lemm_rec_rel_li}). This gives us:
	\begin{align*} 
	e_{I_{(t+1)}} \leq \left( 1- \frac{p}{2} \right) e_{I_{(t)}} + \frac{2A}{p} \eta^2
	\end{align*}
	where   $e_{I_{(t+1)}} := \mathbb{E} \Vert \bX^{I_{(t+1)}} - \Bar{\bX}^{I_{(t+1)}} \Vert_F^2 + \mathbb{E} \Vert \bX^{I_{(t+1)}} - \hat{\bX}^{I_{(t+2)}} \Vert_F^2$ and  $A = 2nG^2H^2 + \frac{p}{2}\left(\frac{8nG^2H^2}{\omega} + \frac{5\omega n}{4 \eta^{1-\epsilon} }\right) $. \\
	It can be seen that $e_{I_{(t)}}  \leq \frac{4A}{p^2}\eta^2$ satisfies the recursion above, similar to argument in \cite[Lemma A.1]{koloskova_decentralized_2019}. Observing that $\mathbb{E}[ \Vert \bX^{I_{(t)}} - \Bar{\bX}^{I_{(t)}} \Vert_F^2 ] \leq e_{I_{(t)}} $ completes the proof.
\end{proof}
\subsection{Proof for Theorem \ref{thm_noncvx_fix_li} (Non-convex objective with constant step size)} \label{proof_thm_noncvx_fix_li}
\begin{proof}[Proof of Theorem ~\ref{thm_noncvx_fix_li}]
	We start the proof with learning rate set to $\eta_t$. We do not use any implicit algebraic structure of the learning rate until (\ref{suppl_fix_var_eqn_li}), thus the analysis remains the same till then for both constant learning rate $\eta_t = \eta$ and for decaying $\eta_t$. We do this to reuse the analysis till (\ref{suppl_fix_var_eqn_li}) in the proof for non-convex objective with varying step size (Theorem \ref{thm_noncvx_var_li}) provided in Section \ref{proof_thm_noncvx_var_li}. We substitute $\eta_t=\eta$ after (\ref{suppl_fix_var_eqn_li}) in this section to proceed with proof for non-convex objective with fixed step size. \\ \\
	Initial part of the proof uses techniques from \cite[Theorem A.2]{koloskova_decentralized_2019}. Consider expectation taken over sampling at time instant $t$: ${\boldsymbol{\xi}^{(t)}} = \{ \xi_1^{(t)}, \xi_2^{(t)}, \hdots, \xi_n^{(t)} \}$ and using $\Bar{\bX}^{(t)}  = \Bar{\bX}^{(t+\frac{1}{2})} $ (from (\ref{mean_seq_iter})) which gives: $ \Bar{\bx}^{(t+1)} =  \frac{1}{n}\sum_{j=1}^n \nabla F_j(\bx_j^{(t)},\xi_j^{(t)}) $ ,  which gives us:
	\begin{align} \label{non_cvx_upd}
	\mathbb{E}_{\boldsymbol{\xi}^{(t)}} f(\bar{\bx}^{(t+1)}) & = \mathbb{E}_{\boldsymbol{\xi}^{(t)}} f \left(\bar{\bx}^{(t)} - \frac{\eta_t}{n}\sum_{j=1}^n \nabla F_j(\bx_j^{(t)},\boldsymbol{\xi}^{(t)}_{j} ) \right) \notag \\
	\intertext{Using the L-smoothness of $f$ as in (\ref{l_smooth}),we get: }
	\mathbb{E}_{\boldsymbol{\xi}^{(t)}} f(\bar{\bx}^{(t+1)})  & \leq f(\bar{\bx}^{(t)}) - \mathbb{E}_{\boldsymbol{\xi}^{(t)}} \left\langle \nabla f(\bar{\bx}^{(t)}) , \frac{\eta_t}{n}\sum_{j=1}^n \nabla F_j(\bx_j^{(t)},\boldsymbol{\xi}^{(t)}_{j} ) \right\rangle \notag \\
	& \hspace{1.5cm} + \mathbb{E}_{\boldsymbol{\xi}^{(t)}} \frac{L}{2}\eta_t^2 \left\Vert \frac{1}{n}\sum_{j=1}^n \nabla F_j(\bx_j^{(t)},\boldsymbol{\xi}^{(t)}_{j} ) \right\Vert_2^2
	\end{align}
	To estimate the second term in (\ref{non_cvx_upd}), we note that :
	\begin{align*} \label{non_cvx_upd_t1}
	-\eta_t & \mathbb{E}_{\boldsymbol{\xi}^{(t)}} \left\langle \nabla f(\bar{\bx}^{(t)}) , \frac{1}{n}\sum_{j=1}^n \nabla F_j(\bx_j^{(t)},\boldsymbol{\xi}^{(t)}_{j} ) \right\rangle \\
	& \hspace{2cm} = - \eta_t \left\langle \nabla f(\bar{\bx}^{(t)}) , \frac{1}{n}\sum_{j=1}^n \nabla f_j(\bx_j^{(t)} ) \right\rangle \notag \\
	& \hspace{2cm} \stackrel{(a)}{=} - \eta_t  \Vert \nabla f(\bar{\bx}^{(t)})  \Vert_2^2 + \eta_t \left\langle \nabla f(\bar{\bx}^{(t)}) , \nabla f(\bar{\bx}^{(t)}) - \frac{1}{n}\sum_{j=1}^n \nabla f_j(\bx_j^{(t)}) \right\rangle \notag \\
	& \hspace{2cm} = - \eta_t  \Vert \nabla f(\bar{\bx}^{(t)})  \Vert_2^2 + \eta_t  \left\langle \nabla f(\bar{\bx}^{(t)}) ,  \frac{1}{n}\sum_{j=1}^n ( \nabla f_j(\bar{\bx}^{(t)}) - \nabla f_j(\bx_j^{(t)}) ) \right\rangle \notag \\
	& \hspace{2cm} \stackrel{(b)}{\leq} - \frac{\eta_t}{2}  \Vert \nabla f(\bar{\bx}^{(t)})  \Vert_2^2 + \frac{\eta_t}{2n}  \sum_{j=1}^n  \Vert \nabla f_j(\bar{\bx}^{(t)}) - \nabla f_j(\bx_j^{(t)}) \Vert^2 \notag 
	\end{align*}
   where in $(a)$ we add and subtract $\nabla f(\bar{\bx}^{(t)})$ and $(b)$ follows by noting that $ \left\langle \mathbf{p},\mathbf{q}  \right\rangle \leq \frac{\verts{\mathbf{p}}^2 +\verts{\mathbf{q}}^2  }{2}$ for any $\mathbf{p},\mathbf{q} \in \mathbb{R}^d$. Using $L$-Lipschitz continuity of gradient of $f_j$ for $j \in [n]$,we have:
	\begin{align}
	-\eta_t \mathbb{E}_{\boldsymbol{\xi}^{(t)}} \left\langle \nabla f(\bar{\bx}^{(t)}) , \frac{1}{n}\sum_{j=1}^n \nabla F_j(\bx_j^{(t)},\boldsymbol{\xi}^{(t)}_{j} ) \right\rangle& \leq - \frac{\eta_t}{2}  \Vert \nabla f(\bar{\bx}^{(t)})  \Vert_2^2 + \frac{\eta_t L^2}{2n}  \sum_{j=1}^n  \Vert \bar{\bx}^{(t)} - \bx_j^{(t)} \Vert^2
	\end{align}
	To estimate the last term in (\ref{non_cvx_upd}), we add and subtract $\nabla f(\bar{\bx}^{(t)}) = \frac{1}{n}\sum_{j=1}^{n} \nabla f_i (\bar{\bx}_t ) $ and $\frac{1}{n}\sum_{j=1}^n \nabla f_j({\bx}_t^{(j)})$
	\begin{align} 
	&\frac{L}{2}\eta_t^2 \mathbb{E} _{\boldsymbol{\xi}^{(t)}} \left\Vert \frac{1}{n}\sum_{j=1}^n \nabla F_j(\bx_j^{(t)},\boldsymbol{\xi}^{(t)}_{j} ) \right\Vert_2^2 \notag \\
	&\hspace{1cm} = \mathbb{E} _{\boldsymbol{\xi}^{(t)}} \left[ \frac{L}{2}\eta_t^2 \left\Vert \frac{1}{n}\sum_{j=1}^n (\nabla F_j(\bx_j^{(t)},\boldsymbol{\xi}^{(t)}_{j} ) - \nabla f_j(\bx_j^{(t)} ) ) + \frac{1}{n}\sum_{j=1}^n (\nabla f_j(\bx_j^{(t)}) - \nabla f_j(\bar{\bx}^{(t)})) + \nabla f(\bar{\bx}^{(t)}) \right\Vert_2^2 \right] \notag \\
	&\hspace{1cm} \leq  L\eta_t^2 \mathbb{E} _{\boldsymbol{\xi}^{(t)}} \left\Vert \frac{1}{n}\sum_{j=1}^n (\nabla F_j(\bx_j^{(t)},\boldsymbol{\xi}^{(t)}_{j} - \nabla f_j(\bx_j^{(t)} ) )  \right\Vert_2^2 +  \frac{2L\eta_t^2}{n} \sum_{j=1}^n  \left\Vert  (\nabla f_j(\bx_j^{(t)}) - \nabla f_j(\bar{\bx}^{(t)}))\right\Vert_2^2 \notag \\
	& \hspace{2cm} + 2L\eta_t^2 \left\Vert \nabla f(\bar{\bx}^{(t)}) \right\Vert_2^2 \notag 
	\end{align}
	Using the variance bound (\ref{bound_var}) for the first term and $L-$Lipschitz continuity of gradients of $f_j$ for $j \in [n]$ for the second, we get:
	\begin{align}\label{non_cvx_upd_t2}
	\frac{L}{2}\eta_t^2 \mathbb{E} _{\boldsymbol{\xi}^{(t)}} \left\Vert \frac{1}{n}\sum_{j=1}^n \nabla F_j(\bx_j^{(t)},\boldsymbol{\xi}^{(t)}_{j} ) \right\Vert_2^2 
	& \leq \frac{L\eta_t^2 \bar{\sigma}^2}{n} +    \frac{2L^3\eta_t^2}{n} \sum_{j=1}^n \left\Vert  \bx_j^{(t)} - \bar{\bx}^{(t)} \right\Vert_2^2 + 2L\eta_t^2 \left\Vert \nabla f(\bar{\bx}^{(t)}) \right\Vert_2^2
	\end{align}
	Substituting (\ref{non_cvx_upd_t1}), (\ref{non_cvx_upd_t2}) to (\ref{non_cvx_upd}) and taking expectation w.r.t the entire process gives:
	\begin{align} \label{suppl_fix_var_eqn} 
	\mathbb{E} [f(\bar{\bx}^{(t+1)})] 
	& \leq  \mathbb{E}f(\bar{\bx}^{(t)}) - \eta_t \left( \frac{1}{2} - 2L\eta_t \right) \mathbb{E} \Vert \nabla f(\bar{\bx}^{(t)})  \Vert_2^2 + \frac{L\eta_t^2 \bar{\sigma}^2}{n} \notag \\
	& \qquad  + \left( \frac{\eta_tL^2}{2n} + \frac{2L^3\eta_t^2}{n} \right) \sum_{j=1}^n \mathbb{E} \Vert \bar{\bx}^{(t)} - \bx_j^{(t)} \Vert^2 
	\end{align}
	Let $I_{(t+1)_0}$ denote the latest synchronization step before or equal to $(t+1)$. Then we have:
	\begin{align*}
	\bX^{(t+1)}  & = \bX^{I_{(t+1)_0}} - \sum_{t' = I_{(t+1)_0}}^{t}\eta_{t'}\partial F(\bX^{(t')}, \boldsymbol{\xi}^{(t')} ) \\
	\Bar{\bX}^{(t+1)}  & = \bar{\bX}^{I_{(t+1)_0}} - \sum_{t' = I_{(t+1)_0}}^{t}\eta_{t'} \partial F(\bX^{(t')}, \boldsymbol{\xi}^{(t')} ) \frac{\mathbbm{1}\mathbbm{1}^T}{n} 
	\end{align*} 
	Thus the following holds:
	\begin{align*} 
	\mathbb{E} & \Vert \bX^{(t+1)} -  \Bar{\bX}^{(t+1)} \Vert_F^2  = \mathbb{E} \left\Vert \bX^{I_{(t+1)_0}} - \Bar{\bX}^{I_{(t+1)_0}} -\sum_{t' = I_{(t+1)_0}}^{t} \eta_{t'}\partial F(\bX^{(t')}, \boldsymbol{\xi}^{(t')} ) \left( \mathbf{I} - \frac{1}{n} \mathbbm{1}\mathbbm{1}^T  \right) \right\Vert_F^2  \\
	& \hspace{2cm} \leq 2 \mathbb{E}\Vert \bX^{I_{(t+1)_0}} - \Bar{\bX}^{I_{(t+1)_0}}  \Vert_F^2 + 2\mathbb{E} \left\Vert \sum_{t' = I_{(t+1)_0}}^{t} \eta_{t'}\partial F(\bX^{(t')}, \boldsymbol{\xi}^{(t')} ) \left( \mathbf{I} - \frac{1}{n} \mathbbm{1}\mathbbm{1}^T  \right) \right\Vert_F^2 
	\end{align*}
	Using (\ref{bound_frob_mult}) for the second term in above and noting that $\mathbb{E} \left\Vert \sum_{t' = I_{(t+1)_0}}^{{t}} \eta_{t'}\partial F(\bX^{(t')}, \boldsymbol{\xi}^{(t')} )\right\Vert_F^2  \leq \eta_{I_{(t+1)_0}} nH^2G^2 $ and $\Vert \frac{\mathbbm{1}\mathbbm{1}^T}{n} - \mathbf{I} \Vert_2^2 = 1 $ from (\ref{bound_gap_grad}) and (\ref{bound_W_mat}) (with $k=0$) respectively, we have:
	\begin{align}\label{suppl_noncvx_li_gammat_temp}
	\mathbb{E} \Vert \bX^{(t+1)} -  \Bar{\bX}^{(t+1)} \Vert_F^2  \leq 2 \mathbb{E}\Vert \bX^{I_{(t+1)_0}} - \Bar{\bX}^{I_{(t+1)_0}}  \Vert_F^2 + 2H^2n \eta_{I_{(t+1)_0}}^2G^2
	\end{align}
	By noting that $ \sum_{j=1}^{n} \mathbb{E} \Vert \bar{\bx}^{(t)} - \bx_j^{(t)} \Vert^2  = 	\mathbb{E} \Vert \bX^{(t)} -  \Bar{\bX}^{(t)} \Vert_F^2 $, 	we use (\ref{suppl_noncvx_li_gammat_temp}) to bound the last term in (\ref{suppl_fix_var_eqn}) which gives:
	\begin{align} \label{suppl_fix_var_eqn_li} 
	\mathbb{E} [f(\bar{\bx}^{(t+1)})] 
	& \leq  \mathbb{E}f(\bar{\bx}^{(t)}) - \eta_t \left( \frac{1}{2} - 2L\eta_t \right) \mathbb{E} \Vert \nabla f(\bar{\bx}^{(t)})  \Vert_2^2 + \frac{L\eta_t^2 \bar{\sigma}^2}{n} \notag \\
	& \hspace{1cm} + \left( \frac{\eta_tL^2}{2n} + \frac{2L^3\eta_t^2}{n} \right) \left[2 \mathbb{E}\Vert \bX^{I_{(t)_0}} - \Bar{\bX}^{I_{(t)_0}}  \Vert_F^2 + 2H^2n \eta_{I_{(t)_0}}^2G^2 \right]
	\end{align}
	We now replace $\eta_t$ with a fixed learning rate $\eta$ to proceed with the proof :
	\begin{align*}
	\mathbb{E} [f(\bar{\bx}^{(t+1)})] 
	& \leq  \mathbb{E}f(\bar{\bx}^{(t)}) - \eta \left( \frac{1}{2} - 2L\eta \right) \mathbb{E} \Vert \nabla f(\bar{\bx}^{(t)})  \Vert_2^2 + \frac{L\eta^2 \bar{\sigma}^2}{n} \\
	& \qquad  + \left( \frac{\eta L^2}{2n} + \frac{2L^3\eta^2}{n} \right) \left[2 \mathbb{E}\Vert \bX^{I_{(t)_0}} - \Bar{\bX}^{I_{(t)_0}}  \Vert_F^2 + 2H^2n \eta^2G^2 \right]
	\end{align*}
	Using Lemma \ref{lemm_dec_li_sgd_fix}, for $A=2nG^2H^2 + \frac{p }{2} \left(\frac{8nG^2H^2}{\omega} + \frac{5\omega n }{4 \eta^{(1-\epsilon)}} \right) $, we have $\mathbb{E} \Vert \bX^{I_{(t)_0}} - \Bar{\bX}^{I_{(t)_0}} \Vert_F^2  \leq \frac{4A\eta^2}{p^2}$. Substituting this in above relation gives us:
	\begin{align*}
	\mathbb{E} f(\bar{\bx}^{(t+1)}) & \leq \mathbb{E} f(\bar{\bx}^{(t)}) - \eta \left( \frac{1}{2} - 2L\eta \right) \mathbb{E} \Vert \nabla f(\bar{\bx}^{(t)})  \Vert_2^2+ \frac{L\bar{\sigma}^2\eta^2}{n} \\
	& \qquad + \left( \frac{\eta L^2}{2n} + \frac{2L^3\eta^2}{n} \right) \left(\frac{8A}{p^2} + 2nH^2G^2 \right) \eta^2 
	\end{align*}
	For the choice of $\eta = \sqrt{\frac{n}{T}}$ and $T \geq 64nL^2$, we have $\eta \leq \frac{1}{8L}$, giving:
	\begin{align*}
	\mathbb{E} f(\bar{\bx}^{(t+1)}) & \leq \mathbb{E} f(\bar{\bx}^{(t)}) - \frac{\eta}{4} \mathbb{E} \Vert \nabla f(\bar{\bx}^{(t)})  \Vert_2^2 +  \frac{L^2}{2n}\left(\frac{8A}{p^2} + 2nH^2G^2 \right)\eta^3 \\
	& \qquad + \frac{2L^3}{n}\left(\frac{8A}{p^2} + 2nH^2G^2 \right) \eta^4 + \frac{L\bar{\sigma}^2\eta^2}{n} 
	\end{align*}
	Rearranging the terms in above and summing from $0$ to $T-1$, we get:
	\begin{align*}
	\sum_{t=0}^{T-1} \eta  \mathbb{E} \Vert \nabla f(\bar{\bx}^{(t)}) \Vert_2^2 & \leq 4 \left( f(\bar{\bx}_{0}) - \mathbb{E} f(\bar{\bx}^{(t)}) \right) +  \frac{2L^2}{n}\left(\frac{8A}{p^2} + 2nH^2G^2 \right) { \sum_{t=0}^{T-1}\eta^3} \\
	& \hspace{1cm} + \frac{8L^3}{n}\left(\frac{8A}{p^2} + 2nH^2G^2 \right){ \sum_{t=0}^{T-1}\eta^4} + \frac{4L \bar{\sigma}^2}{n} {\sum_{t=0}^{T-1}\eta^2}
	\end{align*}
	Dividing both sides by $\eta T$ and by noting that $\mathbb{E} f(\bar{\bx}^{(t)}) \geq f^*$ , we have:
	\begin{align*}
	\frac{\sum_{t=0}^{T-1}  \mathbb{E} \Vert \nabla f(\bar{\bx}^{(t)}) \Vert_2^2}{T} & \leq \frac{4 \left( f(\bar{\bx}_{0}) - f^* \right) }{\eta T} +  \frac{2L^2}{n}\left(\frac{8A}{p^2} + 2nH^2G^2 \right) \eta^2 \\
	& \qquad + \frac{8L^3}{n}\left(\frac{8A}{p^2} + 2nH^2G^2 \right)\eta^3 + \frac{4L \bar{\sigma}^2}{n} \eta
	\end{align*}
	Noting that $\frac{8A}{p^2} \geq 2nH^2G^2 $, we get:
	\begin{align*}
		\frac{\sum_{t=0}^{T-1}  \mathbb{E} \Vert \nabla f(\bar{\bx}^{(t)}) \Vert_2^2}{T}  \leq \frac{4 \left( f(\bar{\bx}_{0}) - f^* \right) }{\eta T} +  \frac{32L^2A}{np^2} \eta^2 + \frac{128L^3A}{np^2}\eta^3 + \frac{4L \bar{\sigma}^2}{n} \eta
	\end{align*}
	Substituting the value of $A=2nG^2H^2 + \frac{p }{2} \left(\frac{8nG^2H^2}{\omega} + \frac{5\omega n}{4 \eta^{1-\epsilon}} \right) $, we have:
	\begin{align*}
\frac{\sum_{t=0}^{T-1}  \mathbb{E} \Vert \nabla f(\bar{\bx}^{(t)}) \Vert_2^2}{T}  & \leq \frac{4 \left( f(\bar{\bx}_{0}) - f^* \right) }{\eta T} +  \frac{32L^2}{np^2} \eta^2 (1+4L\eta) \left[ 2nG^2H^2 + \frac{p }{2} \left(\frac{8nG^2H^2}{\omega} \right)  \right]  \\ 
& \qquad  + \frac{32L^2}{np^2} \eta^2 (1+4L\eta) \left( \frac{5p\omega n}{8 \eta^{1-\epsilon}} \right)     + \frac{4L \bar{\sigma}^2}{n} \eta \\
& = \frac{4 \left( f(\bar{\bx}_{0}) - f^* \right) }{\eta T} +  \frac{64G^2H^2L^2}{p^2} \eta^2 (1+4L\eta) \left( 1+ \frac{2p}{\omega}  \right)  \\ 
& \qquad  + \frac{20L^2 \omega}{p} \eta^{(1+\epsilon)} (1+4L\eta)      + \frac{4L \bar{\sigma}^2}{n} \eta
\end{align*}	
	Substituting $\eta = \sqrt{\frac{n}{T}}$, we get the convergence rate as:
	\begin{align*}
	\frac{\sum_{t=0}^{T-1}  \mathbb{E} \Vert \nabla f(\bar{\bx}^{(t)}) \Vert_2^2}{T}  &
	 \leq \frac{4 \left( f(\bar{\bx}_{0}) - f^* + L\bar{\sigma}^2 \right) }{\sqrt{nT}} +  \frac{64G^2H^2L^2n}{T p^2} \left( 1+ \frac{2p}{\omega}  \right)  \\
	 & \qquad +  \frac{256G^2H^2L^3  n^{\nicefrac{3}{2}} }{T^{\nicefrac{3}{2}} p^2} \left( 1+ \frac{2p}{\omega}  \right) + \frac{20L^2 \omega  \sqrt{n^{(1+\epsilon)}}  } { p \sqrt{T^{(1+\epsilon)}} }  +     \frac{80L^3 \omega  \sqrt{n^{(2+\epsilon)}}  } { p \sqrt{T^{(2+\epsilon)}} }
	\end{align*}	
	for some $\epsilon \in (0,1)$. This completes the proof of Theorem \ref{thm_noncvx_fix_li}.
\end{proof}
\subsection{Non-convex objective with varying stepsize} \label{proof_thm_noncvx_var_li}
\begin{theorem}[Smooth, non-convex case with decaying learning rate]\label{thm_noncvx_var_li} 
	Suppose $f_i$, for all $i\in[n]$ be $L$-smooth. Let $\C$ be a compression operator with parameter equal to $\omega \in (0,1]$. Let $gap(\I_T)\leq H$. 
	If we run SPARQ-SGD with decaying learning rate $\eta_t := \frac{b}{a+t}$ (with $a \geq 8bL$, $b>0$), an increasing threshold function $c_t \sim o(t)$, specifically, $c_t \leq c_0 t^{(1-\epsilon)} $ for all $t$ where $\epsilon \in (0,1)$ and consensus step-size $\gamma = \frac{2\delta \omega}{64 \delta + \delta^2 + 16 \beta^2 + 8 \delta \beta^2 - 16\delta \omega}$, (where $\beta = \max_i \{ 1- \lambda_i(W) \}$), and let the algorithm generate $\{ \bx_i ^{(t)}  \}_{t=0}^{T-1}$ for $i\in[n]$. Then for $p=\frac{\gamma\delta}{8}$, the averaged iterates $\bar{\bx}^{(t)} := \frac{1}{n} \sum_{i=0}^n \bx_i^{(t)}$ satisfy:
	\begin{align*}
\frac{\sum_{t=0}^{T-1} \eta_t  \mathbb{E} \Vert \nabla f(\bar{\bx}^{(t)}) \Vert_2^2}{{\sum_{t=0}^{T-1}\eta_t}}  & \leq \frac{4 \left( f(\bar{\bx}_{0}) -f^* \right) }{b \log \left( \frac{T+a-1}{a} \right)} +   \frac{3840L^2G^2H^2}{p^2} \frac{ \left(\frac{b^3}{a^3} + \frac{b^3}{2a^2}\right)}{b \log \left( \frac{T+a-1}{a} \right)}  + \frac{400L^2 \omega}{p}\frac{\left(\frac{c_0 b^3}{a^3} + \frac{b^3}{(1+\epsilon)a^{(1+\epsilon)}}\right) }{b \log \left( \frac{T+a-1}{a} \right)}  \\
& \qquad+ \frac{15360L^3G^2H^2}{p^2} \frac{ \left(\frac{b^4}{a^4} + \frac{b^4}{3a^3}\right)}{b \log \left( \frac{T+a-1}{a} \right)}+ \frac{1600L^3 \omega}{p} \frac{ \left(\frac{c_0 b^4}{a^4} + \frac{b^4}{(2+\epsilon)a^{(2+\epsilon)}}\right) }{b \log \left( \frac{T+a-1}{a} \right)} \\
& \qquad + \frac{4L \bar{\sigma}^2}{n} \frac{\left(\frac{b^2}{a^2}+\frac{b^2}{a}\right)}{b \log \left( \frac{T+a-1}{a} \right)}
	\end{align*}
\end{theorem}
Thus, for decaying learning rate, we get a convergence rate of $\mathcal{O} \left( \frac{1}{\log T} \right)$. 
\begin{proof}
	We can use the proof of Theorem \ref{thm_noncvx_fix_li} exactly until (\ref{suppl_fix_var_eqn_li}) which gives us:
	\begin{align*} 
	\mathbb{E} [f(\bar{\bx}^{(t+1)})] & \leq  \mathbb{E}f(\bar{\bx}^{(t)}) - \eta_t \left( \frac{1}{2} - 2L\eta_t \right) \mathbb{E} \Vert \nabla f(\bar{\bx}^{(t)})  \Vert_2^2 + \frac{L\eta_t^2 \bar{\sigma}^2}{n} \\
	& \hspace{1cm} + \left( \frac{\eta_tL^2}{2n} + \frac{2L^3\eta_t^2}{n} \right) \left[2 \mathbb{E}\Vert \bX^{I_{(t)_0}} - \Bar{\bX}^{I_{(t)_0}}  \Vert_F^2 + 2H^2n \eta_{I_{(t)_0}}^2G^2 \right]
	\end{align*}
	By Lemma \ref{lem_dec_li_sgd}, for $A_{I_{(t)_0}} = 2nG^2H^2 + \frac{p}{2}\left(\frac{8nG^2H^2}{\omega} + \frac{5\omega n c_{I_{(t)_0}}}{4}\right)$ with $p = \frac{\gamma \delta}{8}$ ($\gamma$ is defined in statement of Theorem \ref{thm_cvx_li}), we have : $\mathbb{E}\Vert \bX^{I_{(t)_0}} - \Bar{\bX}^{I_{(t)_0}}  \Vert_F^2  \leq \frac{20A_{I_{(t)_0}}}{p^2}\eta_{I_{(t)_0}}^2$. Substituting this in above, we have:
	\begin{align} \label{suppl_noncvx_var_li_eqn}
	\mathbb{E} [f(\bar{\bx}^{(t+1)})] 
	& \leq  \mathbb{E}f(\bar{\bx}^{(t)}) - \eta_t \left( \frac{1}{2} - 2L\eta_t \right) \mathbb{E} \Vert \nabla f(\bar{\bx}^{(t)})  \Vert_2^2 + \frac{L\eta_t^2 \bar{\sigma}^2}{n}  \notag \\
	& \qquad + \left( \frac{\eta_tL^2}{2n} + \frac{2L^3\eta_t^2}{n} \right) \left[\frac{40A_{I_{(t)_0}}}{p^2}\eta_{I_{(t)_0}}^2 + 2H^2n \eta_{I_{(t)_0}}^2G^2 \right]
	\end{align}
	We also note that:
	$\frac{\eta_{I_{(t)_0}}}{\eta_{t}} = \frac{a+t}{a+ I_{(t)_0}} \leq \frac{a+I_{(t)_0}+H}{a+ I_{(t)_0}} = \left( 1 + \frac{H}{a+ I_{(t)_0}} \right) \leq \left( 1 + \frac{H}{a} \right) \stackrel{(a \geq H)}{\leq} 2$. As ${I_{(t)_0}}$ denotes the last synchronization index before $t$ and $c_t$ is increasing in $t$, we have $A_{I_{(t)_0}} \leq A_t$.
	\begin{align*}
	\mathbb{E} f(\bar{\bx}^{(t+1)}) & \leq \mathbb{E} f(\bar{\bx}^{(t)}) - \eta_t \left( \frac{1}{2} - 2L\eta_t \right) \mathbb{E} \Vert \nabla f(\bar{\bx}^{(t)})  \Vert_2^2 + \frac{L\eta_t^2 \bar{\sigma}^2}{n} \\
	& \qquad + \left( \frac{\eta_tL^2}{2n} + \frac{2L^3\eta_t^2}{n} \right) \left( \frac{160A_t}{p^2} + 8nH^2 G^2 \right) \eta_t^2 
	\end{align*}
	where $A_t = 2nG^2H^2 + \frac{p}{2}\left(\frac{8nG^2H^2}{\omega} + \frac{5\omega n c_t}{4}\right)$. For the choice of $\eta_t = \frac{b}{t+a}$ and $a \geq 8bL$, we have $\eta_t \leq \frac{1}{8L}$, giving:
	\begin{align*}
	\mathbb{E} f(\bar{\bx}^{(t+1)}) & \leq \mathbb{E} f(\bar{\bx}^{(t)}) - \frac{\eta_t}{4} \mathbb{E} \Vert \nabla f(\bar{\bx}^{(t)})  \Vert_2^2+ \frac{L\eta_t^2 \bar{\sigma}^2}{n} +  \frac{L^2}{2n} \left( \frac{160A_t}{p^2} + 8nH^2 G^2 \right) \eta_t^3\\
	& \qquad + \frac{2L^3}{n} \left( \frac{160A_t}{p^2} + 8nH^2 G^2 \right) \eta_t^4 
	\end{align*}
	Noting that $\frac{160A_t}{p^2} \geq 8nG^2H^2$, we can simplify the above expression as:
	\begin{align*}
\mathbb{E} f(\bar{\bx}^{(t+1)}) \leq \mathbb{E} f(\bar{\bx}^{(t)}) - \frac{\eta_t}{4} \mathbb{E} \Vert \nabla f(\bar{\bx}^{(t)})  \Vert_2^2 +  \frac{160L^2}{np^2} A_t \eta_t^3 +  \frac{640L^3}{np^2} A_t \eta_t^4 + \frac{L\eta_t^2 \bar{\sigma}^2}{n}
\end{align*}	
Substituting the value of $A_t = 2nG^2H^2 + \frac{p}{2}\left(\frac{8nG^2H^2}{\omega} + \frac{5\omega n c_t}{4}\right)$, we have:
\begin{align*}
	\mathbb{E} f(\bar{\bx}^{(t+1)}) & \leq \mathbb{E} f(\bar{\bx}^{(t)}) - \frac{\eta_t}{4} \mathbb{E} \Vert \nabla f(\bar{\bx}^{(t)})  \Vert_2^2 +  320L^2G^2H^2 \left( \frac{1}{p^2} + \frac{2}{p \omega}  \right)  \eta_t^3 +  \frac{100L^2 \omega}{p} c_t \eta_t^3 \\
	& \qquad +  1280L^3G^2H^2 \left( \frac{1}{p^2} + \frac{2}{p \omega}  \right)  \eta_t^4 +  \frac{400L^3 \omega}{p} c_t \eta_t^4  + \frac{L\eta_t^2 \bar{\sigma}^2}{n}
\end{align*}	
Using the fact that $p \leq \omega$ (as $\delta \leq 1$ and $p:= \frac{\gamma^* \delta}{8}$ with $\gamma^* \leq \omega $), the above can be further simplified as:
\begin{align*}
\mathbb{E} f(\bar{\bx}^{(t+1)}) & \leq \mathbb{E} f(\bar{\bx}^{(t)}) - \frac{\eta_t}{4} \mathbb{E} \Vert \nabla f(\bar{\bx}^{(t)})  \Vert_2^2 +  \frac{960L^2G^2H^2}{p^2}  \eta_t^3 +  \frac{100L^2 \omega}{p} c_t \eta_t^3 \\
& \qquad +  \frac{3840L^3G^2H^2}{p^2} \eta_t^4 +  \frac{400L^3 \omega}{p} c_t \eta_t^4  + \frac{L\eta_t^2 \bar{\sigma}^2}{n}
\end{align*}
	Rearranging the terms in above and summing from $0$ to $T-1$, we get:
	\begin{align*}
	\sum_{t=0}^{T-1} \eta_t  \mathbb{E} \Vert \nabla f(\bar{\bx}^{(t)}) \Vert_2^2  &\leq 4 \left( f(\bar{\bx}_{0}) - \mathbb{E} f(\bar{\bx}^{(t)}) \right) +  \frac{3840L^2G^2H^2}{p^2} { \sum_{t=0}^{T-1}\eta_t^3} + \frac{15360L^3G^2H^2}{p^2} { \sum_{t=0}^{T-1}\eta_t^4} \\
	& \qquad +  \frac{400L^2 \omega}{p} { \sum_{t=0}^{T-1} c_t \eta_t^3} + \frac{1600L^3 \omega}{p} { \sum_{t=0}^{T-1} c_t \eta_t^4} + \frac{4L \bar{\sigma}^2}{n} {\sum_{t=0}^{T-1}\eta_t^2}
	\end{align*}
	Dividing both sides by $\sum_{t=0}^{T-1}\eta_t$ , we get:
	\begin{align} \label{non_cvx_li_rec_rel}
	\frac{\sum_{t=0}^{T-1} \eta_t  \mathbb{E} \Vert \nabla f(\bar{\bx}^{(t)}) \Vert_2^2}{{\sum_{t=0}^{T-1}\eta_t}}  & \leq \frac{4 \left( f(\bar{\bx}_{0}) - \mathbb{E} f(\bar{\bx}_{T-1}) \right) }{\sum_{t=0}^{T-1}\eta_t} +   \frac{3840L^2G^2H^2}{p^2} \frac{ \sum_{t=0}^{T-1}\eta_t^3}{\sum_{t=0}^{T-1}\eta_t}  + \frac{400L^2 \omega}{p}\frac{\sum_{t=0}^{T-1}c_t\eta_t^3}{\sum_{t=0}^{T-1}\eta_t} \notag \\
	& \qquad+ \frac{15360L^3G^2H^2}{p^2} \frac{ \sum_{t=0}^{T-1}\eta_t^4}{\sum_{t=0}^{T-1}\eta_t}+ \frac{1600L^3 \omega}{p} \frac{ \sum_{t=0}^{T-1} c_t \eta_t^4}{\sum_{t=0}^{T-1}\eta_t}  + \frac{4L \bar{\sigma}^2}{n} \frac{\sum_{t=0}^{T-1}\eta_t^2}{\sum_{t=0}^{T-1}\eta_t}
	\end{align}
	We now note the following bounds on the sums involved in the RHS of \eqref{non_cvx_li_rec_rel}:
	\begin{align*}
	\sum_{t=0}^{T-1} \eta_t & \geq \int_{0}^{T-1} \frac{b}{t+a} dt = b \log \left( \frac{T+a-1}{a} \right) \\
	\sum_{t=0}^{T-1} \eta_t^2 & \leq \eta_0^2 + \int_{0}^{T-1} \frac{b^2}{(t+a)^2} dt  \leq \eta_0^2 + \int_{0}^{\inf} \frac{b^2}{(t+a)^2} dt = \eta_0^2 + \frac{b^2}{a} = \frac{b^2}{a^2} + \frac{b^2}{a}  \\
	\sum_{t=0}^{T-1} \eta_t^3 & \leq \eta_0^3 + \int_{0}^{T-1} \frac{b^3}{(t+a)^3} dt  \leq \eta_0^3 + \int_{0}^{\inf} \frac{b^3}{(t+a)^3} dt = \eta_0^3 + \frac{b^3}{2a^2} = \frac{b^3}{a^3} + \frac{b^3}{2a^2} \\
	\sum_{t=0}^{T-1} \eta_t^4 & \leq \eta_0^4 + \int_{0}^{T-1} \frac{b^4}{(t+a)^4} dt  \leq \eta_0^4 + \int_{0}^{\inf} \frac{b^4}{(t+a)^4} dt = \eta_0^3 + \frac{b^4}{3a^3} = \frac{b^4}{a^4} + \frac{b^4}{3a^3}
	\end{align*}
	\begin{align*}
	\sum_{t=0}^{T-1}c_t \eta_t^3 & \leq c_{(0)} \eta_0^3 + \int_{0}^{T-1} \frac{b^3 t^{(1-\epsilon)} }{(t+a)^3} dt  \leq c_{(0)}\eta_0^3 +\int_{0}^{T-1} \frac{b^3 (t+a)^{(1-\epsilon)} }{(t+a)^3} dt \leq c_{(0)}\eta_0^3 + \frac{b^3}{(1+\epsilon)a^{(1+\epsilon)}}  \\
	\sum_{t=0}^{T-1} c_t \eta_t^4 & \leq c_{(0)} \eta_0^4 + \int_{0}^{T-1} \frac{b^4t^{(1-\epsilon)}}{(t+a)^4} dt  \leq c_{(0)}\eta_0^4 + \int_{0}^{T-1} \frac{b^4(t+a)^{(1-\epsilon)} }{(t+a)^4} dt \leq c_{(0)}\eta_0^4 + \frac{b^4}{(2+\epsilon)a^{(2+\epsilon)}} 
	\end{align*}
	Substituting these bounds in (\ref{non_cvx_li_rec_rel}) and noting that $\mathbb{E} f(\bar{\bx}_{T-1}) \geq f(x^*) = f^*  $   we get:
	\begin{align*}
	\frac{\sum_{t=0}^{T-1} \eta_t  \mathbb{E} \Vert \nabla f(\bar{\bx}^{(t)}) \Vert_2^2}{{\sum_{t=0}^{T-1}\eta_t}}  & \leq \frac{4 \left( f(\bar{\bx}_{0}) -f^* \right) }{b \log \left( \frac{T+a-1}{a} \right)} +   \frac{3840L^2G^2H^2}{p^2} \frac{ \left(\frac{b^3}{a^3} + \frac{b^3}{2a^2}\right)}{b \log \left( \frac{T+a-1}{a} \right)}  + \frac{400L^2 \omega}{p}\frac{\left(\frac{c_{(0)} b^3}{a^3} + \frac{b^3}{(1+\epsilon)a^{(1+\epsilon)}}\right) }{b \log \left( \frac{T+a-1}{a} \right)}  \\
& \qquad+ \frac{15360L^3G^2H^2}{p^2} \frac{ \left(\frac{b^4}{a^4} + \frac{b^4}{3a^3}\right)}{b \log \left( \frac{T+a-1}{a} \right)}+ \frac{1600L^3 \omega}{p} \frac{ \left(\frac{c_{(0)} b^4}{a^4} + \frac{b^4}{(2+\epsilon)a^{(2+\epsilon)}}\right) }{b \log \left( \frac{T+a-1}{a} \right)} \\
& \qquad + \frac{4L \bar{\sigma}^2}{n} \frac{\left(\frac{b^2}{a^2}+\frac{b^2}{a}\right)}{b \log \left( \frac{T+a-1}{a} \right)}
	\end{align*}
This completes proof of Theorem \ref{thm_noncvx_var_li} .
\end{proof} 

\end{document}